\documentclass{article}
\usepackage{geometry}
\usepackage{hyperref}
\usepackage[T1]{fontenc}
\usepackage[round]{natbib}
\usepackage{amsmath}
\usepackage{amssymb}
\usepackage{amsfonts}
\usepackage{mathrsfs}
\usepackage{dsfont}
\usepackage{bm}
\usepackage{graphicx}
\usepackage{booktabs}
\def\dif{\mathop{}\hphantom{\mskip-\thinmuskip}\mathrm{d}}
\let\daccent\d
\let\d\relax
\newcommand\d{\ifmmode\dif\else\expandafter\daccent\fi}
\usepackage{ntheorem}
\theoremseparator{.}

\newtheorem{theorem}{Theorem}[section]
\newtheorem{corollary}{Corollary}[section]

\newtheorem{proposition}{Proposition}[section]
\newtheorem{assumption}{Assumption}[section]

\title{PAC-Bayesian Generalization Bounds for Graph Convolutional Networks on Inductive Node Classification}
\author{
    Huayi Tang \& Yong Liu \\
    Gaoling School of Artificial Intelligence\\
    Renmin University of China\\
    \texttt{\{huayitang,liuyonggsai\}@ruc.edu.cn}
}

\date{}

\hypersetup{colorlinks, 
            linkcolor=red,
            citecolor=blue,
            urlcolor=magenta,
            linktocpage,
            plainpages=false}

\begin{document}

\maketitle

\begin{abstract}
    Graph neural networks (GNNs) have achieved remarkable success in processing graph-structured data across various applications. A critical aspect of real-world graphs is their dynamic nature, where new nodes are continually added and existing connections may change over time. Previous theoretical studies, largely based on the transductive learning framework, fail to adequately model such temporal evolution and structural dynamics. In this paper, we presents a PAC-Bayesian theoretical analysis of graph convolutional networks (GCNs) for inductive node classification, treating nodes as dependent and non-identically distributed data points. We derive novel generalization bounds for one-layer GCNs that explicitly incorporate the effects of data dependency and non-stationarity, and establish sufficient conditions under which the generalization gap converges to zero as the number of nodes increases. Furthermore, we extend our analysis to two-layer GCNs, and reveal that it requires stronger assumptions on graph topology to guarantee convergence. This work establishes a theoretical foundation for understanding and improving GNN generalization in dynamic graph environments.
\end{abstract}

\section{Introduction}

Graph neural networks (GNNs) \citep{Gori2005, Scarselli2009} represent a family of fundamental models designed for learning representations from graph-structured data, which consists of nodes with features and edges representing connections. Given an input graph, GNNs iteratively aggregate node features according to the graph's topological structure. These aggregated features are then transformed through non-linear activation functions to generate hidden representations. Through multiple such aggregation-and-transformation iterations, the final hidden features produced by GNNs serve as comprehensive node-level embeddings, which can be subsequently utilized for downstream tasks including node classification \citep{Scarselli2009}, link prediction \citep{Kipf2016}, and community detection \citep{Cavallari2017}. Furthermore, graph-level representations can be derived by applying a readout function to these node embeddings, enabling applications such as graph classification \citep{Li2016}. The field of GNNs has experienced rapid architectural evolution in recent years, progressing from traditional spatial \citep{Justin2017, Kipf2017, Hamilton2017, Petar2018, Keyulu2018, Chen2020} and spectral \citep{Defferrard2016, Gasteiger2018, Chien2021, Zhu2021, He2021, Bianchi2022} GNNs to more advanced graph transformers \citep{Yuan2025} and, most recently, graph foundation models \citep{Mao2024, Liu2025}. These advanced architectures have found widespread application across diverse domains, including recommendation systems \citep{Gao2023}, financial applications \citep{Wang2022}, molecular property prediction \citep{Wieder20201}, drug discovery \citep{Fang2025}, and intelligent transportation systems \citep{Rahmani2023}.

The remarkable success of graph neural networks (GNNs) in practical applications has motivated researchers to conduct deeper theoretical analyses, in order to further improve their performance or design novel network architectures. Similar to other neural networks \citep{He2025}, theoretical analyses of GNNs typically fall into three categories:  expressive power \citep{Zhang2025, Morris2024}, optimization \citep{Morris2024}, and generalization \citep{Morris2024, Vasileiou2025}. While expressive power is important, we focus more critically on generalization---the ability of GNN models to perform well on unseen data---as this aspect most directly reflects real-world application requirements. We also remarked that optimization would also be included when studying generalization, since the optimization algorithm used for training GNNs would also be considered. Previous studies analyzing the generalization performance of GNNs have primarily focused on two tasks: graph classification and node classification. Graph classification is often considered theoretically simpler, as each graph is treated as an independent data point sampled from a common distribution. In contrast, node classification presents significantly greater analytical challenges due to the inherent structural dependencies among nodes within a graph.
Existing studies \citep{Oono2020, Esser2021, Cong2021, Tang2023} address this issue by adopting the transductive learning framework (referred to as transductive setting~$1$ or the problem of estimating the values of a function at given points in \citealp{Vapnik1982, Vapnik1998}), where test nodes are sampled without replacement from the complete node set. Under this setting, the model has access to features from both training and test nodes during training, while only the training node labels are observed, with the objective of predicting labels for the test nodes. This approach offers significant practical and theoretical advantages. From a practical standpoint, it aligns with the standard practice of training GNNs in real-world application scenarios. Theoretically, this framework simplifies analysis by treating training and test nodes partition as the sole source of randomness, thereby circumventing the need to explicitly handle complex inter-node dependencies. However, this setting presents several limitations. First, it requires the graph structure to remain static—--an assumption rarely satisfied in real-world applications. For instance, in social networks, new nodes (users) are continuously added, while existing edges (friendships) may be deleted over time \citep{Feng2024}. Second, the transductive framework only evaluates model performance on observed nodes, providing no theoretical guarantees for new nodes whose features and connections are unavailable during training. These limitations motivate us to develop learning guarantees for GNNs under the inductive learning framework, where the population risk is defined over previously unseen nodes rather than the original dataset set.

In this work, we present a PAC-Bayesian theoretical analysis of graph convolutional networks (GCNs) \citep{Kipf2017} for inductive node classification. Under this setting, GCNs operate on graphs with inherently dependent node features, with the objective of predicting labels for newly introduced nodes whose features and connections to the existing graph are unavailable during training. Each node is treated as a data point composed of features and an unique label. Unlike transductive learning frameworks, our approach treats node features, labels, and edge connections as random variables to properly account for the stochastic nature of real-world graph evolution. First, we establish general PAC-Bayesian generalization bounds for learning algorithms on dependent and non-identically distributed data. Building on this foundation, we derive specific generalization bounds for one-layer GCNs. Our proof technique involves a key decomposition of the generalization gap into two components, where the first term is unbiased yet the second one is biased. For the second one, we introduce an auxiliary distribution for aggregation coefficients and further decompose it into several other terms. The resulting bounds explicitly capture the effects of data points dependency and non-stationarity. We then provide more fine-grained formulations of the derived bounds under the assumption that data points are generated by a geometrically ergodic Markov process. Furthermore, we establish non-trivial sufficient conditions guaranteeing that the generalization gap converges to zero as the number of nodes approaches infinity. Finally, we extend our analysis to two-layer GCN architectures. While these results are similar with that of one-layer GCNs, they require stronger assumptions about the graph's topological structure to guarantee a meaningful convergence tendency. 

The rest of this paper is organized as follows. We introduce the mathematical notations and assumptions used throughout this paper along with the inductive node classification setting in Section~\ref{preliminaries}. The main theoretical results are presented in Section~\ref{main_sec} and the experimental setting and results are detailed in Section. Finally, we give a brief discussion and then conclude the paper in Section~\ref{dis_con}. Complete proofs of all theoretical results in the main text are provided in the appendix.

\section{Preliminaries}\label{preliminaries}

\subsection{Notations of Probability Theory and Information Theory}

We first introduce some commonly used mathematical notations. The sequence and the set are represented by $(\cdot)$ and $\{\cdot\}$, respectively. The notations $\mathbb{R}$ and $\mathbb{N}$ represents the set of all real number and natural numbers, respectively. Particularly, $\mathbb{R}_{+}$ and $\mathbb{N}_{+}$ are the set of all non-negative real number and positive natural numbers, respectively. We use $[n]$ as a shorthand of the set $\{1, 2, \ldots, n\}$. Given a sequence $(S_1,S_2,\ldots,S_n)$, the notation $S_{[i,j]}$ represent the contiguous subsequence $(S_i,\ldots,S_j)$ from index $i$ to $j$, where $i,j \in [n]$. Particularly, if $i<j$, we use $S_{(i,j]}$ to denote contiguous subsequence $(S_{i+1},\ldots,S_j)$ from index $(i+1)$ to $j$. We stipulate that $S_{[i,j]}=S_i$ if $i = j$, and $S_{i:j}=\varnothing$ if $i>j$. Clearly, if $i=1$ and $j=n$, we have $S_{[1,n]} = (S_1,S_2,\ldots,S_n)$, and we would shorten it as $S_{[n]}$. 

Next, we present some notations in probability theory and information theory that used in this work. For any given non-empty $\Omega$, we use $(\Omega, \Sigma_{\Omega})$ to denote a measurable space, where $\Sigma_{\Omega}$ is a sigma-algebra on $\Omega$. The sets of all probability measures on $(\Omega, \Sigma_{\Omega})$ is denoted by $\mathcal{P}(\Omega, \Sigma_{\Omega})$, which would be shortened as $\mathcal{P}(\Omega)$ when the sigma-algebra associated with the measure is clear from the context. Each probability measure $P \in \mathcal{P}(\Omega, \Sigma_{\Omega})$ can be combined with $(\Omega, \Sigma_{\Omega})$, which give raise to a probability space $(\Omega, \Sigma_{\Omega}, P)$. We adopt the convention that using uppercase letters to denote random variables, while their corresponding lowercase letters represent their realizations. Concretely, for a given probability space $(\Omega, \Sigma_{\Omega}, P)$, a random variable $X: \Omega \to \mathbb{R}$ is a $\Sigma_{\Omega}$-measurable function that satisfies $X^{-1}(B) = \{ \omega \in \Omega: X(\omega) \in B\} \in \Sigma_{\Omega}$ for all Borel subset $B$. Accordingly, its distribution measure $P_X$ is defined as a pushforward measure of $P$ given by $P_X = P(\{ \omega \in \Omega: X(\omega) \in B\})$. For two random variables $X$ and $Y$ defined on the same probability space, their joint distribution and the conditional distribution of $Y$ given $X$ are denoted by $P_{X,Y}$ and $P_{Y|X}$, respectively. We use $P \ll Q$ to represent that $P$ is absolutely continuous with respect to (w.r.t.) $Q$, that is, $Q(X)=0 \Rightarrow P(X)=0$ for all $X \in \Sigma_{\Omega}$. By the Radon-Nikodym theorem, $P \ll Q$ guarantees the existence of the Radon-Nikodym derivative $\dif P/\dif Q$ \citep{Rudin1987, Hellstrom2025}. With this notation, we introduce two information measures that would be used in this paper: the total varation and the R\'{e}nyi divergence ($\alpha$ divergence). Specifically, for any two probability measures $P, Q \in \mathcal{P}(\Omega)$, the total variation between them is defined by ${\rm D}_{\rm TV}(P,Q) = \sup_{\mathcal{A} \in \Sigma_{\Omega}} \vert P(\mathcal{A}) - Q(\mathcal{A}) \vert$. Let $\alpha \in (0,1) \cup (1, \infty)$, the R\'{e}nyi divergence between $P$ and $Q$ is denoted by ${\rm D}_{\alpha}(P||Q) = \log ( (\mathbb{E}_Q[(\dif P/\dif Q)^\alpha])^{1/(\alpha-1)} )$. Unlike the total variation, the R\'{e}nyi divergence ${\rm D}_{\alpha}(P||Q)$ typically differs from ${\rm D}_{\alpha}(Q||P)$ unless $\alpha=1/2$. Besides, the R\'{e}nyi divergence ${\rm D}_{\alpha}(P||Q)$ tends to the Kullback–Leibler (KL) divergence ${\rm D}_{\rm KL}(P||Q)$ and the worst-case regret ${\rm D}_{\infty}(P||Q) = \log (\mathrm{ess} \sup_P \{\dif P /\dif Q\})$ as $\alpha \to 1$ and $\alpha \to \infty$ \citep{Erven2014}, respectively. 

Lastly, we briefly give some notations and definition of Markov chain. For two given measurable spaces $(\mathcal{X}, \Sigma_{\mathcal{X}})$ and $(\mathcal{Y}, \Sigma_{\mathcal{Y}})$, a Markov transition kernel from $\mathcal{X}$ to $\mathcal{Y}$ is defined as a mapping $P: \mathcal{X} \times \Sigma_{\mathcal{Y}} \to [0,1]$ such that: (1) for each $x \in \mathcal{X}$, $P(x, \cdot)$ is a probability measure over $(\mathcal{Y}, \Sigma_{\mathcal{Y}})$, and (2) for each $B \in \Sigma$, $P(\cdot, B)$ is a measurable function on $(\mathcal{X}, \Sigma_{\mathcal{X}})$. Intuitively speaking, $P(x, B)$ is the probability of transitioning from state $x \in \mathcal{X}$ to a set of states $B \subseteq \mathcal{Y}$ in one step. To simplify the analysis, we stipulate that that $\mathcal{Y}=\mathcal{X}$ in this paper, thereby $P$ describing how the system evolves from the current state $x \in \mathcal{X}$ to the next state $y \in \mathcal{X}$. For any two Markov kernels $P_1$ and $P_2$, the product kernel is the defined as $(P_1 \circ P_2) (x, B) = \int_{\mathcal{X}} P_2(u, B) P_1(x, \dif u)$ for each $x \in \mathcal{X}$ and $B \in \Sigma_{\mathcal{X}}$. Then, for any fixed integer $n \in \mathbb{N}_+$, the product of $n$ Markov kernel $P$ is defined recursively by $P^{n+1} \triangleq P^n \circ P$. Similarly, $P^n(x, B)$ is the probability of transitioning from state $x$ to a set of states $B \subseteq \mathcal{X}$ in $n$-step. A Markov chain $\{X_1,\ldots,X_n\}$ with transition kernel $P$ and stationary distribution $\pi$ is geometrically ergodic \citep{Meyn2009} if there exists a constant $\rho \in (0,1)$ and a function $M: \mathcal{X} \to \mathbb{R}_+$ such that ${\rm D}_{\rm TV}(P^n(x,\cdot), \pi) \leq M(x) \rho^n$ for each $x \in \mathcal{X} $ and $n \in \mathbb{N}_{+}$. We refer interested readers for the book of \cite{Meyn2009} for more details on this topic.

\subsection{Notations of Graph Convolutional Networks and Problem Setting}

Let $\mathcal{G}=(\mathcal{V}, \mathcal{E})$ be an undirected graph without self-loops or isolated vertices, where $\mathcal{V}$ and $\mathcal{E}$ are the edge sets and node sets, respectively. Denote by $n = \vert \mathcal{V} \vert$ the number of nodes in $\mathcal{G}$. The node set of $\mathcal{G}$ is represented as $\mathcal{V} = \{v_i\}_{i=1}^n$, where each node $v_i \in \mathcal{V}$ contains features $X_i \in \mathbb{R}^d$ and a label $Y_i \in \mathcal{Y}$ for $i \in [n]$. We treat the feature-label pair $S_i = (X_i, Y_i) \in (\mathcal{X} \times \mathcal{Y})$ as a data point.
The structure of $\mathcal{G}$ is characterized by its adjacent matrix $(A_{i,j})_{n \times n} \in \{0,1\}^{n \times n}$, where $A_{i,j} = 1$ if and only if $(v_i,v_j) \in \mathcal{E}$. Since $\mathcal{G}$ does not have self-connections, we have $A_{i,i} = 0$ for each $i \in [n]$. For any $i \in [n]$, denote by $D_i = \sum_{j=1}^n A_{i,j}$ the degree of node $v_i$, the normalized adjacent matrix $(\tilde{A}_{i,j})_{n \times n} \in \mathbb{R}^{n \times n}$ with self-connections is defined as
\begin{equation}\label{tilde_a}
    \tilde{A}_{i,j} = \begin{cases}
        \frac{A_{i,j}}{\sqrt{D_i+1}\sqrt{D_j+1}}, & j \ne i \\
        \frac{1}{D_i+1}, & j = i
    \end{cases}.
\end{equation}

In this paper, we focus on the learning task that a new node $v_{n+1}$ with features $X_{n+1}$ and label $Y_{n+1}$ is added to the current graph $\mathcal{G}$ in future, and our goal is to train a GCN model based on $\mathcal{G}$ and use it to predict the label of the new coming node. Clearly, this learning task is inductive since the model is unaware of both the feature of the new coming nodes and their potential connections to the existing nodes during training. To depict the structure of the graph after adding the new coming node, we introduce the sequence of aggregation coefficients $A_{n+1,[n+1]}=(A_{n+1,1},A_{n+1,2},\ldots,A_{n+1,n+1})$, where $A_{n+1,i} \neq 0$ if and only if there is an edge between node $v_i$ and the new coming node $v_{n+1}$ for each $i \in [n]$. To maintain the symmetry of the adjacent matrix, we also introduce another sequence $(A_{1,n+1},A_{2,n+1},\ldots,A_{n,n+1})$, where $A_{i,n+1} = A_{n+1,i}$ for each $i \in [n]$. Similarly, we would assume that the new coming node $v_{n+1}$ does not have self-connection, thereby we have $A_{n+1,n+1}=0$. We remark that both the data point $S_{n+1}=(X_{n+1},Y_{n+1})$ and the connections $A_{n+1,[n+1]}$ of the new coming node $v_{n+1}$ are random varaibles drawn from the conditional law $P(\cdot|S_{[n]}, (A_{i,j})_{n \times n})$.

Now we present the formulation of the training objective. Since the bias vector can be absorbed into the weight matrix, we would neglect the bias vector in the forward process for simplicity. Let us first discuss the case that the model is a single layer GCN with learnable weight matrix $W \in \mathbb{R}^{d \times \vert \mathcal{Y} \vert}$. For any $j \in [n]$, let $\mathcal{N}(j)$ denote the neighbors' indices of node $v_j$ in $\mathcal{V}$, and $\mathcal{N}^+(j) = \mathcal{N}(j) \cup \{j\}$ denote the inclusive neighbors' indices. Recall that both the features of the new coming node and its connection with current nodes are not available during training. For any $j \in [n]$, the hidden feature of the node $v_j$ after aggregation is
\begin{equation}
\begin{aligned}
    f_{W,j} (X_{[n]}, (\tilde{A}_{i,j})_{n \times n}) = \phi \Bigg( \sum_{k=1}^n \tilde{A}_{j,k} X_k W_1 \Bigg),
\end{aligned}
\end{equation}
where $\phi$ is the activation function. For ease of analysis, we default to the ReLU activation function unless otherwise specified throughout this paper. Here we use $W = [{\rm vec}[W_1]]$ to denote the collection of all learnable parameters, where $\text{vec}[\cdot]$ is the vectorization operator that reshapes a matrix into column vector by stacking its column vectors sequentially. The parameter space is defined accordingly by $\mathcal{W} = \mathbb{R}^{d \times \vert \mathcal{Y} \vert}$. Denote by $\widehat{\mathcal{Y}} \in \mathbb{R}^{\vert \mathcal{Y} \vert}$ and $\ell: \widehat{\mathcal{Y}} \times \mathcal{Y} \to \mathbb{R}_{\geq 0}$ the space of hidden feature and the loss function, respectively. The empirical risk of this GNN model is defined as
\begin{equation}
    \widehat{R}(W,S_{[n]},(A_{i,j})_{n \times n}) = \frac{1}{n} \sum_{j=1}^n \ell(f_{W,j}(X_{[n]}, (\tilde{A}_{i,j})_{n \times n}), Y_j).
\end{equation}
After training, the new coming node is now added to the graph, and both its feature and connections with previous nodes are revealed to the model. Then, the hidden feature of the node $v_{n+1}$ is defined as
\begin{equation}
\begin{aligned}
    f_W(X_{[n+1]}, A_{n+1,[n+1]}) = \sum_{k=1}^{n+1} A_{n+1,k} X_k W,
\end{aligned}
\end{equation}
which is computed by aggregating the hidden features based on the connections $A_{n+1,[n+1]}$. The expected risk of the model is accordingly defined as
\begin{equation}
    R(W,S_{[n]},(A_{i,j})_{n \times n}) = \mathbb{E}_{S_{n+1}, A_{n+1,[n+1]}} \big[ \ell(f_W(X_{[n+1]}, A_{n+1,[n+1]}),Y_{n+1}) \big\vert S_{[n]},(A_{i,j})_{n \times n} \big].
\end{equation}
The inductive generalization gap is then defined as
\begin{equation}
    \mathcal{E}(W,S_{[n]},(A_{i,j})_{n \times n}) = R(W,S_{[n]},(A_{i,j})_{n \times n}) - \widehat{R}(W,S_{[n]},(A_{i,j})_{n \times n}).    
\end{equation}
Next, we discuss the case that the model is a two layer GCN. Denote by $(\hat{A}_{i,j})_{n \times n} \in \mathbb{R}^{n \times n}$ the normalized adjacent matrix with
\begin{equation}\label{hat_a}
    \hat{A}_{i,j} = \begin{cases}
        \frac{A_{i,j}}{\sqrt{D_i}\sqrt{D_j}}, & j \ne i \\
        0, & j = i
    \end{cases}.
\end{equation}
For any $\imath \in [n]$, the hidden feature of the node $v_{\imath}$ after aggregation is 
\begin{equation}
    f_{W,\imath}(X_{[n]}, (\hat{A}_{i,j})_{n \times n}) = \phi \Bigg( \sum_{j=1}^n \hat{A}_{\imath, j} \phi \Bigg( \sum_{k=1}^n \hat{A}_{j,k} X_k W_1 \Bigg) W_2 \Bigg),
\end{equation}
where $W_1 \in \mathbb{R}^{d \times h}, W_2 \in \mathbb{R}^{h \times \vert \mathcal{Y} \vert}$ are learnable weight matrices. Similarly, we use $W = [{\rm vec}[W_1], {\rm vec}[W_2]]$ to denote the collection of all learnable parameters. Similarly, for the node $v_{n+1}$, its hidden feature is
\begin{equation}
\begin{aligned}
    & f_W(X_{[n+1]}, (\hat{A}_{i,j})_{n \times n}, A_{n+1,[n+1]}) \\
    = & \phi \Bigg( \sum_{j=1}^n A_{n+1,j} \phi \Bigg( \sum_{k=1}^n \hat{A}_{j,k} X_k W_1 + A_{j,n+1} X_{n+1} W_1 \Bigg) W_2 + A_{n+1,n+1} \phi \Bigg( \sum_{k=1}^{n+1} A_{n+1,k} X_k W_1 \Bigg) W_2 \Bigg).
\end{aligned}
\end{equation}
The expected risk are accordingly defined as
\begin{equation}
    R(W,S_{[n]},(A_{i,j})_{n \times n}) = \mathbb{E}_{S_{n+1}, A_{n+1,[n+1]}} \big[ \ell(f_W(X_{[n+1]}, (A_{i,j})_{n \times n}, A_{n+1,[n+1]}),Y_{n+1}) \big\vert S_{[n]}, (A_{i,j})_{n \times n} \big].
\end{equation}
Let $\mathcal{A}$ be the space that includes all values of $(A_{i,j})_{n \times n}$. Under the perspective of PAC-Bayes theory, the learning algorithm is regarded as a stochastic mapping from the training data $(S_{[n]},(A_{i,j})_{n \times n})$ to the parameter $W$, and its randomness is depicted by the Markov transition kernel $P$ from $(\mathcal{X} \times \mathcal{Y})^n \times \mathcal{A}$ to $\mathcal{W}$. For any fixed realization $(s_{[n]},(a_{i,j})_{n \times n})$ of the training data, this Markov kernel yields a probability measure $P((s_{[n]},(a_{i,j})_{n \times n}), \cdot)$ on $(\mathcal{W}, \Sigma_{\mathcal{W}})$, which is generally referred as the posterior distribution $Q(s_{[n]},(a_{i,j})_{n \times n})$. We will shorten this notation to $Q$ when the training data is clear from the context. Also, for any random variable $W: \mathcal{W} \to \mathbb{R}$, the pushforward measure of $Q$ is given by $Q_W$, which would be shorted as $Q$ following the convention in PAC-Bayesian literature. The main goal in this paper is to establish an upper bound for the expectation $\mathbb{E}_{W \sim Q}[\mathcal{E}(W,S_{[n]},(A_{i,j})_{n \times n})]$, where $Q$ should be understood as $Q_W$ from a rigorous perspective.

\subsection{Assumptions}

In this sections, we present some assumptions used throughout this paper. The first assumptions is that the Euclidean norm of the node feature vector is bounded.
\begin{assumption}\label{assump1}
    Suppose that there exists a constant $c_x > 0$ such that $\Vert x \Vert_2 \leq c_x$ holds for all $x \in \mathcal{X}$.
\end{assumption}
This assumption is commonly used in the theoretical analysis of GNNs \citep{Garg2020, Esser2021, Cong2021, Tang2023}, and it can be easily satisfied through applying normalization on the node features. The second assumption is that the Frobenius norm of the normalized adjacent matrix with and without self-loop is bounded.
\begin{assumption}\label{assump2}
    Suppose that there exists a constant $c_a > 0$ such that $\Vert (\tilde{a}_{i,j})_{n \times n} \Vert_{\rm F}, \Vert (\hat{a}_{i,j})_{n \times n} \Vert_{\rm F} \leq c_a$ holds for all $(a_{i,j})_{n \times n} \in \mathcal{A}$, where $(\tilde{a}_{i,j})_{n \times n}$ and $(\hat{a}_{i,j})_{n \times n}$ are defined in Eq.~(\ref{tilde_a}) and Eq.~(\ref{hat_a}), respectively.
\end{assumption}
The Frobenius norm of the normalized adjacent matrix explicitly appears in the generalization bounds of GNNs \citep{Liao2021, Esser2021, Cong2021, Tang2023}, and it can further upper bounded by functions regarding the maximum and minimum node degree. Thus, this assumption can always be satisfied. The third assumption is that the spectral norms of every learnable weight matrix are bounded.
\begin{assumption}\label{assump3}
    Denote by $\mathcal{W} = \mathbb{R}^{d \times \vert \mathcal{Y} \vert}$ the parameter space. Suppose that there exist a constant $c_w > 0$ such that $\Vert w \Vert_2 \leq c_w$ holds for all $w \in \mathcal{W}$.
\end{assumption}
\begin{assumption}\label{assump4}
    Denote by $\mathcal{W} = \mathbb{R}^{d \times h} \times \mathbb{R}^{h \times \vert \mathcal{Y} \vert}$ the parameter space. Suppose that there exist a constant $c_w > 0$ such that $\Vert w_1 \Vert, \Vert w_2 \Vert \leq c_w$ hold for all $(w_1, w_2) \in \mathcal{W}$.
\end{assumption}
This assumption is also commonly used in the theoretical analysis of GNNs \citep{Garg2020, Esser2021, Cong2021, Tang2023}. In practice, we can adopt the layer normalization technique to guarantee that this assumption holds with small value of $c_w$. The fourth assumption is that the loss function $\ell(\cdot, \cdot)$ is $L$-Lipschitz with respect to the first argument.
\begin{assumption}\label{assump5}
    Suppose that there exists a constant $L > 0$ such that $\vert \ell(\hat{y},y) - \ell(\hat{y}',y) \vert \leq L \Vert \hat{y}' - \hat{y} \Vert_2$ holds for any $\hat{y}, \hat{y}' \in \widehat{\mathcal{Y}}$ and $y \in \mathcal{Y}$.
\end{assumption}
Generally, this assumption holds for many commonly used loss function. For example, considering the log-softmax function
\begin{equation}
    \ell(\hat{y}, y) = - \sum_{k=1}^{\vert \mathcal{Y} \vert} y_k \log \frac{e^{\hat{y}_k}}{\sum_{k'=1}^{\vert \mathcal{Y} \vert} e^{\hat{y}_{k'}}},
\end{equation}
which is a standard loss function used in classification task. \cite{Tang2023} show that it satisfies the Lipschitz continuity assumption with $L = \sqrt{2}$. The fifth assumption is that the loss function is bounded.
\begin{assumption}\label{assump6}
    Suppose that there exists a constant $M > 0$ such that $\ell(\hat{y}, y) \leq M$ holds for any $\hat{y} \in \widehat{Y}$ and $y \in \mathcal{Y}$.
\end{assumption}
Although some loss functions could be unbounded, such as the log-softmax function mention above, its numerical value is generally finite in practice, owned to the well-developed neural network weight initialization techniques and matured optimization algorithms. The last assumption is that the activation function is $L_{\phi}$-Lipschitz.
\begin{assumption}\label{assump7}
    Suppose that there exists a constant $L_{\phi} > 0$ such that $\vert \phi(x) - \phi(x') \vert \leq L_{\phi} \vert x - x' \vert $ for any $x, x' \in \mathbb{R}$.
\end{assumption}
This assumption holds for many widely used activation functions. For example, let $\phi(x) = \max(x,0)$ be the ReLU function, one can be easily verified that this assumption holds with $L_{\phi}=1$. Furthermore, for any vector $x, y \in \mathbb{R}^d$, this assumption implies that $\Vert \phi(x) - \phi(y) \Vert \leq L_{\phi} \Vert x - y \Vert$.

\section{Main Results}\label{main_sec}

\subsection{PAC-Bayesian Bounds for Dependent and Non-Identically Distributed Data}\label{sec1}
As previously mentioned, we treat the graph as a collections of random variables with inherent dependencies, which bring us the first challenge when establishing the bounds. The second challenge is that, since the GCN parameters are learned from the data $S$, the empirical risk $\widehat{R}(W,S_{[n]},(A_{i,j})_{n \times n})$ becomes a biased estimator of the expected risk $R(W,S_{[n]},(A_{i,j})_{n \times n})$. Let us turn to the first challenge in this section. Recall that we need to provide an upper bound for the expectation $\mathbb{E}_{W \sim Q}[\mathcal{E}(W,S_{[n]},(A_{i,j})_{n \times n})]$ in the context of PAC-Bayesian theory, as mentioned in Section~\ref{preliminaries}. Here, $Q$ is a data-dependent probability measure. By applying the change of measure technique, the problem reduces to analyze the expectation of a function involving $\mathcal{E}(W,S_{[n]},(A_{i,j})_{n \times n})$ under a data-independent probability measure $P$, which is termed as the prior in PAC-Bayesian theory. The penalty of replacing the posterior $Q$ with the prior $P$ is depicted by certain information measure \citep{Hellstrom2025}. The first challenge arises when analyzing the expectation of a function involving $\mathcal{E}(W,S_{[n]},(A_{i,j})_{n \times n})$ under the prior $P$, where we need to show that $\mathcal{E}(W,S_{[n]},(A_{i,j})_{n \times n})$ concentrates around its expectation with high probability. To this end, we  employ the concentration inequality established by \cite{Kontorovich2008, Kontorovich2017}, which is derived through the martingale method with data dependencies characterized by a Wasserstein matrix. We remark that alternative approaches exist for characterizing the dependency between data points, which consequently lead to concentration inequalities with different formulations. A more detailed discussion of this will be presented in Section~\ref{dis_con}. Integrating the above steps, we obtain the following general PAC-Bayesian bounds for models learning on dependent data.
\begin{theorem}\label{main_pro}
    Let $\Psi: \mathcal{W} \times (\mathcal{X} \times \mathcal{Y})^n \to \mathbb{R}_{\geq 0}$ be a function of $W$ and $S_{[n]}$. For any fixed realization $W=w$ and $i \in [n]$, suppose that $\Psi(w,s_{[n]})$ is $c_i$-Lipschitz w.r.t. $s_i$ under Hamming metric. Let $P \in \mathcal{P}(\mathcal{W})$ be a prior distribution independent of $S$. For any $0<\delta<1$, $\lambda>0$, $\alpha \in (0,1) \cup (1,\infty)$, and any $Q \in \mathcal{P}(\mathcal{W})$ such that $Q \ll P$, with probability at least $1-\delta$ over the randomness of $S_{[n]}$, we have
    \begin{equation}\label{main_pro_bound}
        \mathbb{E}_{W \sim Q} \left[ \Psi(W,S_{[n]}) \right] \leq \frac{\log \mathbb{E}_{W \sim P} \left[ \exp \left\{ \lambda \alpha \mathbb{E}_{S'_{[n]}}[\Psi(W,S'_{[n]})] \right\} \right] + {\rm D}_\alpha (Q||P)+\log(1/\delta)}{\lambda} + \frac{\lambda \left\Vert \bm{\Gamma} \bm{c} \right\Vert^2}{8},
    \end{equation}
    where $\bm{c} = (c_1,\ldots,c_n)^\top \in \mathbb{R}^n$ and $\bm{\Gamma} \in \mathbb{R}^{n \times n}$ whose $(i,j)$-th entry is
    \begin{equation}\label{gamma_matrix}
    \begin{aligned}
    \Gamma_{i,j} = \begin{cases}
        0, & i > j \\
        1, & i = j \\
        \mathop{\rm sup}_{x, z \in \Omega, x^{[n]\setminus \{i\}} = z^{[n]\setminus \{i\}}} {\rm D}_{\rm TV} \left( P_{[j,n]} \big( \cdot {\big |} x_{[i]} \big) - P_{[j,n]} \big(\cdot {\big |} z_{[i]} \big) \right), & i<j
    \end{cases}.
    \end{aligned}
    \end{equation}
\end{theorem}

Let us briefly check that the above bound by considering a special case that data points $S_{[n]} = (S_1,\ldots,S_n)$ are independent and identically distributed. Under this case, we have $\Gamma_{i,j}=0$ for any $i,j \in [n]$ such that $i<j$. If we further assume that $\Psi$ is bounded by a constant $C$, we have $c_i = C/n$ for all $i \in [n]$, which implies that $\Vert \bm{\Gamma} \bm{c} \Vert^2 = C^2/n$. Recall that the R\'{e}nyi divergence degenerate to the KL divergence when $\alpha \to 1.$ Therefore, using a extra assumption that $\mathbb{E}_S[\Psi(w,S)]=0$ holds for any realization $W=w$, we can recover the basic PAC-Bayesian bounds (Theorem~2.1 in \citealp{Alquier2021}) from Theorem~\ref{main_pro} by letting $\alpha \to 1$. Theorem~\ref{main_pro} is also closely related to Theorem~9 of \cite{Begin2016}, since both of them are derived from the same change of measure inequality of R\'{e}nyi divergence (\citealp[Theorem~2.1]{Atar2015}, \citealp[Theorem~8]{Begin2016}). Yet, there are still some significant differences between our result and theirs. First, Eq.~(\ref{main_pro_bound}) is a Catoni-typed PAC-Bayesian bound, since it contains an extra parameter $\lambda$. The bound in Theorem~9 is a Langford-Seeger-Maurer-typed PAC-Bayesian bound that does not include such parameter. Second, the bound in \citep{Begin2016} contains a convex function $\Delta: [0,1]^2 \to \mathbb{R}$, and various bounds can be derived by specifying the choice of $\Delta$. Our result can be regarded as a special case where this convex function is defined as $\Delta(q,p) = q - p$ for any $q,p \in [0,1]$. The advantage of our result is that the assumptions for data points are much weaker than that of \cite{Begin2016}: Theorem~\ref{main_pro} neither requires the data points to be independent nor identically distributed, yet Theorem~9 of \cite{Begin2016} requires both these assumptions. Besides, we also note that \cite{Chugg2023} have established time-uniform PAC-Bayesian bound that also included R\'{e}nyi divergence (see Theorem~6.2 in their paper). Compared with their result, our result provided in Theorem~\ref{main_pro} has a more explicit form and the dependence between data points are depicted by the term $\Vert \bm{\Gamma} \bm{c} \Vert^2$.

Before closing this section, we establish Langford-Seeger-Maurer-typed PAC-Bayesian bounds. The first result is derived by setting the parameter $\lambda = 1 - 1/(2n)$ and using the property of sub-Gaussian variables, inspired by the work of \cite{Hellstrom2020}. The second result is a single-draw generalization bound derived by the technique from \cite{Esposito2021}.
\begin{theorem}\label{main_pro2}
    Under the assumptions of Theorem~\ref{main_pro}, for any $0<\delta<1$, $\alpha \in (0,1) \cup (1,\infty)$, and any $Q \in \mathcal{P}(\mathcal{W})$ such that $Q \ll P$ with probability at least $1-\delta$ over the randomness of $S_{[n]}$, we have
    \begin{equation}\label{main_pro_bound1}
        \forall Q \in \mathcal{P}(\mathcal{W}): \mathbb{E}_{W \sim Q} \left[ \Psi(W,S) \right] \leq \mathbb{E}_{W \sim Q} \mathbb{E}_{S'} [ \Psi(W,S') ] + \sqrt{\frac{n \left( {\rm D}_{\alpha}(Q||P)+\log(\sqrt{2n}/\delta) \right) \Vert \bm{\Gamma} \bm{c} \Vert^2}{(2n-1)}}.
    \end{equation}
    Moreover, with probability at least $1-\delta$ over the randomness of $S_{[n]}$ and $W$, we have
    \begin{equation}\label{main_pro_bound2}
        \Psi(W,S) \leq \mathbb{E}_{S'} [\Psi(W,S')] + \sqrt{\frac{\Vert \bm{\Gamma} \bm{c} \Vert^2}{2}\left( I_{\alpha}(S,W) + \log 2 + \frac{\alpha}{\alpha-1} \log (1/\delta) \right)}.
    \end{equation}
\end{theorem}
For one hand, the bound in Eq.~(\ref{main_pro_bound2}) has more benign dependence on the number of data points, since the bound in Eq.~(\ref{main_pro_bound1}) has a extra term $\log (\sqrt{2n})$. For the other hand, the bound in Eq.~(\ref{main_pro_bound2}) contains a term $\frac{\alpha}{\alpha-1}$, and it goes to infinity as $\alpha$ approaches $1$. Thus, Eq.~(\ref{main_pro_bound2}) fails to give a reasonable learning guarantee when $\alpha \to 1$. However, Eq.~(\ref{main_pro_bound1}) is not affected from this issue as it does not contain such a term. Thus, each result in Theorem~\ref{main_pro2} has distinct strengths, making it applicable to different scenarios. In the next section, we would demonstrate how to apply the above results to establish generalization bounds for GCN model.

\subsection{Generalization Bounds for One-Layer GCN}\label{one_layer_sec}

In this section, we analyze the generalization behavior of a one-layer GCN model using the PAC-Bayesian for dependent and non-identically distributed data we have established. The first step is decomposing the generalization gap $\mathcal{E}(W,S_{[n]},(A_{i,j})_{n \times n})$ into the following two terms
\begin{equation}
\begin{aligned}
    \mathcal{E}_1(W,S_{[n]},(A_{i,j})_{n \times n}) = & \frac{1}{n} \sum_{j=1}^n \mathbb{E}_{S_{[j,n]}} \big[ \ell(f_{W,j}(X_{[n]}, (\tilde{A}_{i,j})_{n \times n}), Y_j) \big\vert S_{[j-1]} \big] - \frac{1}{n} \sum_{j=1}^n \ell(f_{W,j}(X_{[n]}, (\tilde{A}_{i,j})_{n \times n}), Y_j), \\
    \mathcal{E}_2(W,S_{[n]},(A_{i,j})_{n \times n}) = & \mathbb{E}_{S_{n+1}, A_{n+1,[n+1]}} \big[ \ell(f_W(X_{[n+1]}, A_{n+1,[n+1]}),Y_{n+1}) \big\vert S_{[n]},(A_{i,j})_{n \times n} \big] \\
    & - \frac{1}{n} \sum_{j=1}^n \mathbb{E}_{S_{[j,n]}} \big[ \ell(f_{W,j}(X_{[n]}, (\tilde{A}_{i,j})_{n \times n}), Y_j) \big\vert S_{[j-1]} \big].
\end{aligned}
\end{equation}
Let us firstly analyze the term $\mathcal{E}_1(W,S_{[n]},(A_{i,j})_{n \times n})$. It can be verified that
\begin{equation}
    \mathbb{E}_{W \sim Q} \mathbb{E}_{S'_{[n]}}[\mathcal{E}_1(W,S'_{[n]},(a_{i,j})_{n \times n})] = 0    
\end{equation}
for any fixed realization $(A_{i,j})_{n \times n}=(a_{i,j})_{n \times n}$. Thus, in order to apply results in Section~\ref{sec1}, we only need to derive an upper bound for the term $\vert \mathcal{E}_1(w,s_{[n]},(a_{i,j})_{n \times n}) - \mathcal{E}_1(w,s^{(i)}_{[n]},(a_{i,j})_{n \times n}) \vert$ for any fixed realization $W=w$, where $s^{(i)}_{[n]}=(s_1,\ldots,s_{i-1},s'_i,s_{i+1},\ldots,s_n)$ is the sequence of data points that differs from $s_{[n]}=(s_1,\ldots,s_n)$ only in the $i$-th entry. The result is as follows.
\begin{proposition}\label{pro1}
    Suppose that assumptions~\ref{assump1}, \ref{assump2}, \ref{assump3}, \ref{assump5}, and \ref{assump6} hold. For any $\delta \in (0,1)$, $\alpha \in (0,1) \cup (1,\infty)$, and any $Q \in \mathcal{P}(\mathcal{W})$ such that $Q \ll P$, with probability at least $1-\delta/2$ over the randomness of $S_{[n]}$ and $(A_{i,j})_{n \times n}$, we have
    \begin{equation}\label{bound1}
    \mathbb{E}_{W \sim Q} \left[ \mathcal{E}_1(W,S_{[n]},(A_{i,j})_{n \times n}) \right] \leq 2\sqrt{\frac{(2 c_x c_w c_a L + M \max (1, \Vert \widetilde{\bm{\Gamma}} \Vert_{\infty}))^2({\rm D}_{\alpha}(Q||P)+\log(2\sqrt{2n}/\delta) ) \Vert \bm{\Gamma} \Vert^2}{(2n-1)}},
    \end{equation}
    where $\widetilde{\bm{\Gamma}} \in \mathbb{R}^{n \times n}$ is a matrix whose $(i,j)$-th entry is
    \begin{equation}
        \widetilde{\Gamma}_{i,j} = \sup_{s_i, s'_i} {\rm D}_{\rm TV} \left( P_{S_{[j,n]} | S_{[i-1]}=s_{[i-1]}, S_{i}=s_i, S_{(i,j-1]}=s_{(i,j-1]}}, P_{S_{[j,n]} | S_{[i-1]}=s_{[i-1]}, S_{i}=s'_i, S_{(i,j-1]}=s_{(i,j-1]}} \right)
    \end{equation}
    if $j > i$ and $i \in [n-1]$, otherwise $\widetilde{\Gamma}_{i,j} = 0$.
\end{proposition}
For a given sequence $s = (s_1, \ldots, s_n)$, the quantity $\widetilde{\Gamma}_{i,j}$ depicts the effect of modifying only the $i$-th data point $s_i$ among the first $(j-1)$ data points $(s_{1},\ldots,s_{j-1})$ in this sequence on the distributional variation of subsequent data points $(s_j,\ldots,s_n)$. If the dependence between sample points in this sequence is weak, for example, if the $i$-th data point only depends on the previous one or the preceding $k$ data points, where $k$ is a fixed constant, then $\Vert \widetilde{\bm{\Gamma}} \Vert^2_{\infty}$ in Eq.~(\ref{bound1}) is of order $\mathcal{O}(1)$. We will demonstrate this in more detail later by assuming that $s$ is a Markov chain, where the $i$-th data point only depends on the previous one. Besides, the assumption that current data points almost only depends on the preceding $k$ data points is also used by \cite{Abeles2025} to analyze the generalization bound for learning algorithms on stationary non-i.i.d. process. Thus, if the R\'{e}nyi divergence term ${\rm D}_{\alpha}(Q||P)$ can be bounded properly, the bound provided in Eq.~(\ref{bound1}) can be regarded as of order $\mathcal{O}(\sqrt{\log(\sqrt{n})/n})$ in common cases. 

Now, we turn to the analysis of second term $\mathcal{E}_2(W,S_{[n]},(A_{i,j})_{n \times n})$, which is much more challenging than that of the first term since the expected risk $\mathbb{E}_{S_{n+1}, A_{n+1,[n+1]}|S} [ \ell(f_{w,j}(X_{[n+1]}, A_{n+1,[n+1]}),Y_{n+1}) ]$ is \emph{not} an unbiased estimator of the empirical risk $\sum_{i=1}^n \ell(f_{w,j}(X_{[n]}, \tilde{a}_{i,[n]}), Y_i)/n$. The reasons are two folds. First, the data points are not identically distributed, implying that $S_{n+1}$ and $S_i$ may follow different distributions for any $i \in [n]$. Second, even if they follow the same distribution, their aggregated features differ after the neighborhood aggregation due to the different neighbor sets, thereby the hidden feature $f_w(X_{[n+1]}, A_{n+1,[n+1]})$ could also differ from $f_w(X_{[n]}, \tilde{a}_{i,[n]})$ even when $S_{[n]}$ are i.i.d. data points. To address this issue, we firstly introduce an auxiliary distribution $Q_{A_{n+1,[n+1]}|(A_{i,j})_{n \times n}}$ for the aggregation coefficients $A_{n+1,[n+1]} = (A_{n+1,1},\ldots,A_{n+1,n+1})$ of the $(n+1)$-th node conditioned on the connections of existing nodes $(A_{i,j})_{n \times n}$:
\begin{equation}
    \mathbb{P} \left( \cap_{k \in [n] \setminus \{i\}} \{ A_{n+1,k} = \tilde{A}_{i,k} \} \cap \{ A_{n+1,i} = 0 \} \cap \{ A_{n+1,n+1} = \tilde{A}_{i,i} \} \right) = \frac{1}{n}, \ \forall i \in [n].
\end{equation}
Secondly, we need the following extra assumption on the dependency of connections and data points.
\begin{assumption}\label{assump8}
    Suppose that $A_{n+1,[n+1]}$ is independent of $S_{[n]}$ and $S_{n+1}$ is independent of both $A_{n+1,[n+1]}$ and $(A_{i,j})_{n \times n}$. That is, we have
    \begin{equation}
    \begin{aligned}
        P_{S_{n+1}, A_{n+1,[n+1]}|S_{[n]},(A_{i,j})_{n \times n}} = & P_{A_{n+1,[n+1]}|S_{[n]},(A_{i,j})_{n \times n}} P_{S_{n+1}|A_{n+1,[n+1]}, S_{[n]},(A_{i,j})_{n \times n}} \\
        = &  P_{A_{n+1,[n+1]}|(A_{i,j})_{n \times n}} P_{S_{n+1}|S_{[n]}}.
    \end{aligned}
    \end{equation}
\end{assumption}
This assumption plays an important role in decoupling the dependency. Denote by $P_{A_{n+1,[n+1]} | (A_{i,j})_{n \times n}}$ the original distribution of $A_{n+1,[n+1]}$ conditioned on $(A_{i,j})_{n \times n}$, we can decompose $\mathcal{E}_2(W,S_{[n]},(A_{i,j})_{n \times n})$ into the following two terms:
\begin{equation}\label{decom_key2}
\begin{aligned}
    & \mathbb{E}_{S_{n+1}, A_{n+1,[n+1]}|S_{[n]},(A_{i,j})_{n \times n}} \big[ \ell(f_W(X_{[n+1]}, A_{n+1,[n+1]}),Y_{n+1}) \big] \\
    & - \frac{1}{n} \sum_{j=1}^n \mathbb{E}_{S_{[j,n]}|S_{[j-1]}} \big[ \ell(f_{W,j}(X_{[n]}, (\tilde{A}_{i,j})_{n \times n}), Y_j) \big] \\
    = & \mathbb{E}_{A_{n+1,[n+1]}|(A_{i,j})_{n \times n}} \mathbb{E}_{S_{n+1}|S_{[n]}} \big[ \ell(f_W(X_{[n+1]}, A_{n+1,[n+1]}),Y_{n+1}) \big] \\
    & - \frac{1}{n} \sum_{j=1}^n \mathbb{E}_{S_{[j,n]}|S_{[j-1]}} \big[ \ell(f_{W,j}(X_{[n]}, (\tilde{A}_{i,j})_{n \times n}), Y_j) \big] \\
    = & \mathbb{E}_{P_{A_{n+1,[n+1]} | (A_{i,j})_{n \times n}}} \mathbb{E}_{S_{n+1}|S_{[n]}} \big[ \ell(f_W(X_{[n+1]}, A_{n+1,[n+1]}),Y_{n+1}) \big] \\
    & - \mathbb{E}_{Q_{A_{n+1,[n+1]} | (A_{i,j})_{n \times n}}} \mathbb{E}_{S_{n+1}|S_{[n]}} \big[ \ell(f_W(X_{[n+1]}, A_{n+1,[n+1]}),Y_{n+1}) \big] \\
    & + \mathbb{E}_{Q_{A_{n+1,[n+1]} | (A_{i,j})_{n \times n}}} \mathbb{E}_{S_{n+1}|S_{[n]}} \big[ \ell(f_W(X_{[n+1]}, A_{n+1,[n+1]}),Y_{n+1}) \big] \\
    & - \frac{1}{n} \sum_{j=1}^n \mathbb{E}_{S_{[j,n]}|S_{[j-1]}} \big[ \ell(f_{W,j}(X_{[n]}, (\tilde{A}_{i,j})_{n \times n}), Y_j) \big].
\end{aligned}
\end{equation}
By the definition of $Q_{A_{n+1,[n+1]}||(A_{i,j})_{n \times n}}$, the second term in Eq.~(\ref{decom_key2}) can be rewritten as
\begin{equation}
\begin{aligned}
    & \mathbb{E}_{Q_{A_{n+1,[n+1]}||(A_{i,j})_{n \times n}}} \mathbb{E}_{S_{n+1}|S_{[n]}} \big[ \ell(f_W(X_{[n+1]}, A_{n+1,[n+1]}),Y_{n+1}) \big\vert S_{[n]},(A_{i,j})_{n \times n} \big] \\
    & - \frac{1}{n} \sum_{j=1}^n \mathbb{E}_{S_{[j,n]}} \big[ \ell(f_{W,j}(X_{[n]}, (\tilde{A}_{i,j})_{n \times n}), Y_j) \big\vert S_{[j-1]} \big] \\
    = & \frac{1}{n} \sum_{j=1}^n \Big(\mathbb{E}_{S_{n+1}|S_{[n]}} \big[ \ell(f_W(X_{[n+1]}, (\tilde{A}_{j,[j-1]},0,\tilde{A}_{j,(j,n]},\tilde{A}_{j,j})),Y_{n+1}) \big] \\
    & - \mathbb{E}_{ S_{[j,n]} | S_{[j-1]}} \big[ \ell(f_{W,j}((X_{[j-1]}, X_{[j,n]}), (\tilde{A}_{i,j})_{n \times n}), Y_j) \big] \Big),
\end{aligned}
\end{equation}
which can be further bounded by several terms. Based on the above technique, we obtain the following result.
\begin{proposition}\label{pro2}
Suppose that assumptions~\ref{assump1}, \ref{assump2}, \ref{assump3}, \ref{assump5}, \ref{assump6}, and \ref{assump8} hold. For any $ \delta \in (0,1)$, $\alpha \in (0,1) \cup (1,\infty)$, and any $Q \in \mathcal{P}(\mathcal{W})$ such that $Q \ll P$, with probability at least $1-\delta/2$ over the randomness of $S_{[n]}$ and $(A_{i,j})_{n \times n}$, we have
\begin{equation}\label{bound2}
\begin{aligned}
    & \mathbb{E}_{W \sim Q} \left[ \mathcal{E}_2(W,S_{[n]},(A_{i,j})_{n \times n}) \right] \\
    \leq & \frac{M}{n} \sum_{i=1}^n {\rm D}_{\rm TV} \left( P_{S_{n+1}|S_{[n]}}, P_{S_i |S_{[i-1]}} \right) + \frac{M}{n} \sum_{i=1}^n \mathbb{E}_{S_{[i-1]}} \left[ {\rm D}_{\rm TV} \left( P_{S_{(i,n]}|S_{[i-1]}} \otimes P_{S_i |S_{[i-1]}}, P_{S_{[i,n]} | S_{[i-1]}} \right) \right] \\
    & + M {\rm D}_{\rm TV} \left( P_{A_{n+1,[n+1]} | (A_{i,j})_{n \times n}}, Q_{A_{n+1,[n+1]} | (A_{i,j})_{n \times n}} \right) \\
    & + \sqrt{\frac{(2 c_x c_w c_a L + M \Vert \widetilde{\bm{\Gamma}} \Vert_{\infty})^2({\rm D}_{\alpha}(Q||P)+\log(2\sqrt{2n}/\delta) ) \Vert \bm{\Gamma} \Vert^2}{(2n-1)}}.
\end{aligned}
\end{equation}
\end{proposition}
Compared with Eq.~(\ref{bound1}), Eq.~(\ref{bound2}) has some extra terms. Specifically, the first term 
\begin{equation}\label{term1}
    \frac{1}{n}\sum_{i=1}^n {\rm D}_{\rm TV} \left( P_{S_{n+1}|S_{[n]}}, P_{S_i |S_{[i-1]}} \right)
\end{equation}
is the mean total variation distance between the distribution of each data point in the full sequence and that of the $(n+1)$-th data point, which is also referred as the discrepancy between target distribution and the distribution of the sample \citep{Kuznetsov2015, Kuznetsov2020}. If we treat the data sequence $S$ as a stochastic process, this quantity is a natural measure of its non-stationarity \citep{Kuznetsov2015, Kuznetsov2020}. If the sequence $S$ is asymptotically stationary, this term can be further upper bounded by some coefficients, which will be detailed later. The second term
\begin{equation}\label{term2}
    \frac{1}{n} \sum_{i=1}^n \mathbb{E}_{S_{[i-1]}} \left[ {\rm D}_{\rm TV} \left( P_{S_{(i,n]}|S_{[i-1]}} \otimes P_{S_i |S_{[i-1]}}, P_{S_{[i,n]} | S_{[i-1]}} \right) \right]
\end{equation}
is the expectation of the mean total variation distance between the joint distribution of the current data point and the future ones and the product of their marginal distributions conditioned on the previous ones. By Pinsker's inequality and Jensen's inequality, this term can be further bounded by the conditional mutual information between the current data point and the future ones:
\begin{equation}
\begin{aligned}
    & \mathbb{E}_{S_{[i-1]}} \left[ \frac{1}{n}\sum_{i=1}^n {\rm D}_{\rm TV} \left( P_{S_{(i,n]}|S_{[i-1]}} \otimes P_{S_i |S_{[i-1]}}, P_{S_{[i,n]} | S_{[i-1]}} \right) \right] \\
    \leq & \mathbb{E}_{S_{[i-1]}} \left[ \frac{1}{\sqrt{2} n}\sum_{i=1}^n \sqrt{{\rm D}_{\rm KL} \left( P_{S_{(i,n]}|S_{[i-1]}} \otimes P_{S_i |S_{[i-1]}}, P_{S_{[i,n]} | S_{[i-1]}} \right)} \right] \\
    \leq & \frac{1}{\sqrt{2}n} \sum_{i=1}^n \sqrt{ \mathbb{E} \left[ {\rm D}_{\rm KL} \left( P_{S_{(i,n]}|S_{[i-1]}} \otimes P_{S_i |S_{[i-1]}}, P_{S_{[i,n]} | S_{[i-1]}} \right) \right]} = \frac{1}{\sqrt{2}n} \sum_{i=1}^n \sqrt{ I \left( S_i; S_{(i,n]} | S_{[i-1]} \right)}.
\end{aligned}
\end{equation}
Essentially, this term depicts the dependency between the current data point and the future ones conditioned on the previous ones. It can be further bounded under some extra assumptions on the distribution of data points $S$, which will also be detailed later. The third term ${\rm D}_{\rm TV} ( P_{A_{n+1,[n+1]} | (A_{i,j})_{n \times n}}, Q_{A_{n+1,[n+1]} | (A_{i,j})_{n \times n}} )$ is the total variation distance between the vanilla distribution of the aggregation coefficients and the auxiliary distribution conditioned on existing connections. Clearly, if these two distributions are identical, that is, $P_e = Q_e$, this term equals to zero. 
Besides, we remark that there also exists some distributions $P_{A_{n+1,[n+1]} | (A_{i,j})_{n \times n}}$ such that this quantity converges to zero as the increase of $n$. Let $k \in \mathbb{N}_+$ be a fixed integer. Denote by
\begin{equation}
    \widetilde{\mathcal{B}} = \{ (\tilde{b}_{i,1}, \ldots, \tilde{b}_{i,n+1}) \}_{i \in [k]}
\end{equation}
a given set, where each entry is a sequence of length $(n+1)$. Before moving on, we need to make a weak assumptions on the set $\widetilde{\mathcal{B}}$: there exists a constant $B > 0$ such that $0 \leq \tilde{b}_{i,j} \leq B$ holds for any $i \in [k]$ and $j \in [n]$. That is, every entries in the sequence from $\widetilde{\mathcal{B}}$ is non-negative and has a finite upper bound. Let $P_{A_{n+1,[n+1]} | (A_{i,j})_{n \times n}}$ be a distribution such that the following holds
\begin{equation}\label{pe}
\begin{aligned}
    & \mathbb{P} \left( \cap_{t \in [n] \setminus \{i\}} \{ A_{n+1,t} = \tilde{A}_{i,t} \} \cap \{ A_{n+1,i} = 0 \} \cap \{ A_{n+1,n+1} = \tilde{A}_{i,i} \} \right) = \frac{1}{n} - \frac{1}{n^{\frac{3}{2}}}, \ \forall i \in [n], \\
    & \mathbb{P} \left( \cap_{t \in [n+1]} \{ A_{n+1,t} = \tilde{b}_{j,t} \} \right) = \frac{1}{k\sqrt{n}}, \ \forall j \in [k].
\end{aligned}
\end{equation}
It can be verified easily that Eq.~(\ref{pe}) gives a valid probability measure for any fixed realization $(A_{i,j})_{n \times n} = (a_{i,j})_{n \times n}$.
By the definition of total variation distance, we have
\begin{equation}\label{term3}
    {\rm D}_{\rm TV} \left( P_{A_{n+1,[n+1]} | (A_{i,j})_{n \times n}}, Q_{A_{n+1,[n+1]} | (A_{i,j})_{n \times n}} \right) = \frac{1}{2} \sum_{i=1}^n \left\vert \left( \frac{1}{n} - \frac{1}{n^{\frac{3}{2}}} \right) - \frac{1}{n} \right\vert + \frac{1}{2}\sum_{j=1}^k \left\vert \frac{1}{k\sqrt{n}} - 0 \right\vert = \frac{1}{\sqrt{n}}.
\end{equation}
Therefore, for distributions $P_{A_{n+1,[n+1]} | (A_{i,j})_{n \times n}}$ and $Q_{A_{n+1,[n+1]} | (A_{i,j})_{n \times n}}$ that are close in a metric sense, the total variation distance between them decays at a rate $\mathcal{O}(1/\sqrt{n})$. Now we are in a place to establish the PAC-bayesian generalization bound for the single-layer GCN model by combining Proposition~\ref{pro1} with Proposition~\ref{pro2}. The result is as follows.
\begin{theorem}\label{one_layer_bound}
    Under the assumptions of Proposition~\ref{pro2}, for any $\delta \in (0,1)$, $\alpha \in (0,1) \cup (1,\infty)$, and any $Q \in \mathcal{P}(\mathcal{W})$ such that $Q \ll P$ with probability at least $1-\delta$ over the randomness of $S_{[n]}$ and $(A_{i,j})_{n \times n}$, we have
    \begin{equation}\label{one_layer_bound1}
    \begin{aligned}
        & \mathbb{E}_{W \sim Q} \big[ R(W,S_{[n]},(A_{i,j})_{n \times n}) \big] \\
        \leq & \mathbb{E}_{W \sim Q} \big[ \widehat{R}(W,S_{[n]},(A_{i,j})_{n \times n}) \big] + 3\sqrt{\frac{(2 c_x c_w c_a L + M \max (1, \Vert \widetilde{\bm{\Gamma}} \Vert_{\infty}))^2({\rm D}_{\alpha}(Q||P)+\log(2\sqrt{2n}/\delta) ) \Vert \bm{\Gamma} \Vert^2}{(2n-1)}} \\
        & + \frac{M}{n} \sum_{i=1}^n \left[ {\rm D}_{\rm TV} \left( P_{S_{n+1}|S_{[n]}}, P_{S_i |S_{[i-1]}} \right) + \mathbb{E}_{S_{[i-1]}} \left[ {\rm D}_{\rm TV} \left( P_{S_{(i,n]}|S_{[i-1]}} \otimes P_{S_i |S_{[i-1]}}, P_{S_{[i,n]} | S_{[i-1]}} \right) \right]  \right]\\
        & + M {\rm D}_{\rm TV} \left( P_{A_{n+1,[n+1]} | (A_{i,j})_{n \times n}}, Q_{A_{n+1,[n+1]} | (A_{i,j})_{n \times n}} \right).
    \end{aligned}
    \end{equation}
\end{theorem}
Compared with the upper bound derived for non i.i.d. data without graph structure \citep{Kuznetsov2015}, our result contain two extra terms that are displayed in Eq.~(\ref{term2}) and Eq.~(\ref{term3}), respectively. Thus, achieve better generalization performance of inductive task on graph structural data may be essentially more difficulty than that on ordinary data.

Now let us provide more fine-grained analysis on the terms in Eq.~(\ref{one_layer_bound1}) under some more concrete assumptions on the distributions of data points $S_{[n]}$. Following the work of \cite{Alquier2018}, we assume that $S_{[n]}=(S_1,\ldots,S_n)$ is a geometrically ergodic Markov process. Now let us start the elucidation by analyzing the terms $\Vert \bm{\Gamma} \Vert$ and $\Vert \widetilde{\bm{\Gamma}} \Vert_{\infty}$.
\begin{proposition}\label{pro3}
    Suppose that $(S_1,\ldots,S_n)$ is a geometrically ergodic Markov chain, whose transition kernel and stationary distribution is $P$ and $\pi$, respectively. Then we have
    \begin{equation}\label{gamma_bound3}
        \Vert \bm{\Gamma} \Vert \leq 1 + \frac{2 \sup_{x} M(x)}{1 - \rho}, \ \Vert \widetilde{\bm{\Gamma}} \Vert_{\infty} \leq 2 \rho \sup_{x} M(x).
    \end{equation}
\end{proposition}

If we further assume that the Markov chain is uniformly ergodic \citep{Meyn2009}, then there exists a constant $M_0 > 0$ such that $\sup_{x} M(x) \leq M_0$. Then, the upper bounds we derived in Eq.~(\ref{gamma_bound3}) can be regarded as constants, which implies that both $\Vert \bm{\Gamma} \Vert$ and $\Vert \widetilde{\bm{\Gamma}} \Vert_{\infty}$ are of order $\mathcal{O}(1)$ under the Markov assumption by Proposition~\ref{pro3}. Now, we turn to the analysis of the terms in Eq.~(\ref{term1}) and Eq.~(\ref{term2}).
\begin{proposition}\label{pro4}
    Under the assumption of Proposition~\ref{pro3}, we have
    \begin{equation}
    \begin{aligned}
        & \frac{1}{n} \sum_{i=1}^n \mathbb{E}_{S_{[i-1]}} \left[ {\rm D}_{\rm TV} \left( P_{S_{(i,n]}|S_{[i-1]}} \otimes P_{S_i |S_{[i-1]}}, P_{S_{[i,n]} | S_{[i-1]}} \right) \right] \\
        \leq & \frac{1}{n}\left( \frac{2\rho (1-\rho^{n-2}) \mathbb{E} \left[ M(S_1) \right]}{1-\rho} + 4{\rm D}_{\rm TV} \left( P_{S_1}, \pi \right) \right) + 2 \mathbb{E}_{S \sim \pi} \left[{\rm D}_{\rm TV} ( P(S, \cdot), \pi )\right].
    \end{aligned}
    \end{equation}
    and
    \begin{equation}
    \begin{aligned}
         \frac{1}{n}\sum_{i=1}^n {\rm D}_{\rm TV} \left( P_{S_{n+1}|S_{[n]}}, P_{S_i |S_{[i-1]}} \right) \leq {\rm D}_{\rm TV} \left( P_{S_{n+1}|S_n}, \pi \right) + \frac{1}{n}\sum_{i=1}^n {\rm D}_{\rm TV} \left( P_{S_i |S_{i-1}}, \pi \right).
    \end{aligned}
    \end{equation}
\end{proposition}
Proposition~\ref{pro4} shows that both these two terms converge to the same quantity $\mathbb{E}_{S \sim \pi} [{\rm D}_{\rm TV} ( P(S, \cdot), \pi )]$ as $n$ tends to infinity, and this quantity depicts the expected  between the transition kernel and the stationary distribution when sampling from the stationary distribution. If for each state $s$, the discrepancy between the transition kernel $P(s,)$ and the stationary distribution $\pi$ is small, then this expectation is small and thus the value of terms in Eq.~(\ref{term1}) and Eq.~(\ref{term2}) would be small when $n$ is sufficiently large. Particularly, if the transition kernel is fixed to $P(s,\cdot) = \pi(s)$ for any state, then the expectation $\mathbb{E}_{S \sim \pi} [{\rm D}_{\rm TV} ( P(S, \cdot), \pi )]$ equals zero and we exactly recover the case of i.i.d. data points. For more general cases, we will provide a non-trivial example to demonstrate that the expectation can be small later. Combining the results in Proposition~\ref{pro3} with that in Proposition~\ref{pro4}, we obtain the following generalization bound for one-layer GCN model under the Markov assumption.
\begin{corollary}\label{corollary}
    Suppose that $P_{A_{n+1,[n+1]} | (A_{i,j})_{n \times n}}$ follows the distribution defined in Eq.~(\ref{pe}). Under the assumptions of Proposition~\ref{pro1} and Proposition~\ref{pro3}, for any $0<\delta<1$, $\alpha \in (0,1) \cup (1,\infty)$, and any $Q \in \mathcal{P}(\mathcal{W})$ such that $Q \ll P$ with probability at least $1-\delta$ over the randomness of $S$, we have
    \begin{equation}\label{concrete_bound}
    \begin{aligned}
        & \mathbb{E}_{W \sim Q} \big[ R(W,S_{[n]},(A_{i,j})_{n \times n}) \big] \\
        \leq & \mathbb{E}_{W \sim Q} \big[ \widehat{R}(W,S_{[n]},(A_{i,j})_{n \times n}) \big] + 3\sqrt{\frac{(2 c_x c_w c_a L + 2 \rho M M_0)^2(1-\rho+2M_0)^2({\rm D}_{\alpha}(Q||P)+\log(2\sqrt{n}/\delta))}{2(1-\rho)^2(n-1)}} \\
        & + \frac{M}{\sqrt{n}} + \frac{2M}{n}\left( \frac{\rho(1-\rho^{n-2})\mathbb{E} \left[ M(S_1) \right]}{1-\rho} +  2{\rm D}_{\rm TV} \left(\pi, P_{S_1} \right) \right) + 2M \mathbb{E}_{S \sim \pi} \left[ {\rm D}_{\rm TV} \left(P(S,\cdot), \pi \right) \right] \\
        & + M{\rm D}_{\rm TV} \left( P_{S_{n+1}|S_n}, \pi \right) + \frac{M}{n}\sum_{i=1}^n {\rm D}_{\rm TV} \left( P_{S_i |S_{i-1}}, \pi \right).
    \end{aligned}
    \end{equation}
\end{corollary}
Let us consider an example where the state space of the Markov chain is finite. Without loss of generality, we assume that the state space is $\{1,\ldots,N\}$. Let $\bm{p} = (p_1,\ldots,p_N)$ be any valid probability distribution, that is, $p_{i} \in (0, 1)$ for each $i \in [n]$ and $\sum_{i=1}^N p_i = 1$. Denote by $\alpha_1, \ldots, \alpha_{N} \in (0,1)$ some fixed constants, we define the transition matrix as $\bm{P} = (P_{i,j})_{N \times N}$, where $P_{i,j} = (1-\alpha_j)p_j + \alpha_i \mathds{1}\{i = j\}$. We demonstrate that for a Markov chain $(S_1,\ldots,S_n)$ whose transition kernel is $\bm{P}$, the expectation $\mathbb{E}_{S \sim \pi} [{\rm D}_{\rm TV} ( P(S, \cdot), \pi )]$ can be upper bounded by the maximum value $\max_{j \in [N]} \alpha_j$. First, we check that $\bm{p}$ is the stationary distribution of this Markov chain. For any $j \in [N]$, notice that 
\begin{equation}
    \sum_{i=1}^N p_i P_{i,j} = \sum_{i=1}^N p_i \left(\alpha_i \mathds{1}\{i = j\} + (1-\alpha_j)p_j \right) = \alpha_j p_j + (1-\alpha_j)p_j \sum_{i=1}^N p_i = p_j.
\end{equation}
Thus, we have $\bm{p}^\top \bm{P} = \bm{p}$ and thus $\bm{p}$ is the stationary distribution. Next, for any $i \in [N]$, the total variation distance between transition probability $(P_{i,j})_{j \in [N]}$ and the stationary distribution $\bm{p}$ is
\begin{equation}
\begin{aligned}
    {\rm D}_{\rm TV} ( P(S=i, \cdot), \pi ) = & \frac{1}{2} \sum_{j=1}^N \left\vert (1-\alpha_j)p_j + \alpha_i \mathds{1}\{i = j\} - p_j \right\vert = \frac{1}{2} \bigg( \alpha_i(1-p_i) + \sum_{j \ne i} \alpha_j p_j \bigg) \\
    \leq & \frac{\max_{j \in [N]} \alpha_i}{2} \bigg( 1-p_i + \sum_{j \ne i} p_j \bigg) = (1-p_i) \max_{j \in [N]} \alpha_j.
\end{aligned}
\end{equation}
Thus, we have
\begin{equation}
    \mathbb{E}_{S \sim \pi} [{\rm D}_{\rm TV} ( P(S, \cdot), \pi )] = \sum_{i=1}^n p_i {\rm D}_{\rm TV} ( P(S=i, \cdot), \pi ) \leq \sum_{i=1}^n p_i(1-p_i) \max_{j \in [N]} \alpha_j \leq \frac{(N-1) \max_{j \in [N]} \alpha_j}{N^2} ,
\end{equation}
and
\begin{equation}\label{disq_term}
\begin{aligned}
    & {\rm D}_{\rm TV} \left( P_{S_{n+1}|S_n}, \pi \right) + \frac{1}{n}\sum_{i=1}^n {\rm D}_{\rm TV} \left( P_{S_i |S_{i-1}}, \pi \right) \\
    = & \frac{{\rm D}_{\rm TV} \left( P_{S_1}, \pi \right)}{n} + \left( 1 - \frac{1}{n} \right) {\rm D}_{\rm TV} \left( P_{S_{n+1}|S_n}, \pi \right) + \frac{1}{n} \sum_{i=2}^{n+1} {\rm D}_{\rm TV} \left( P_{S_i |S_{i-1}}, \pi \right) \\
    \leq & \frac{{\rm D}_{\rm TV} \left( P_{S_1}, \pi \right)}{n} + \frac{(n-1)(1-p_n)\max_{j \in [N]} \alpha_j}{n} + \frac{1}{n} \sum_{i=2}^{n+1} (1-p_{i-1}) \max_{j \in [N]} \alpha_j.
\end{aligned}
\end{equation}
Then, for any fixed constant $\varepsilon > 0$, we can always guarantee that both the expectation $\mathbb{E}_{S \sim \pi} [{\rm D}_{\rm TV} ( P(S, \cdot), \pi )]$ and the last two terms in Eq.~(\ref{disq_term}) do not exceed $\varepsilon$ by choosing $\alpha_1, \ldots, \alpha_{N} \in (0,1)$ such that $\max_{j \in [N]} \alpha_j \leq \varepsilon/2$. Under this example, Corollary~\ref{corollary} shows that if the number of data points is sufficient large, the expectation of the difference between the expected risk and the empirical risk of a one-layer GCN model under the posterior distribution does not exceed a small constant with high probability. We also point out that there may exist various approaches to analyze and derive upper bounds for the non-stationary related terms in Eq.~(\ref{term1}) and Eq.~(\ref{term2}), and we only display one of them in this paper. The upper bound we provided in Proposition~\ref{pro3} and Proposition~\ref{pro4} may not always be tight, yet they are sufficient for us to elucidate the effect of the non-stationary of the stochastic process $(S_1,\ldots,S_n)$ on the generalization gap of model.

\subsection{Generalization Bounds for Two-Layer GCN}

In Section~\ref{one_layer_sec}, we have established generalization bounds for one-layer GCN model and provided a comprehensive discussion of their implications. In this section, we further extend these theoretical findings to the case of two-layer GCN, following the same idea we demonstrate in Section~\ref{one_layer_sec}. The result is as follows.
\begin{theorem}\label{two_layer_bound}
    Suppose that assumptions~\ref{assump1}, \ref{assump2}, \ref{assump4}, \ref{assump5}, \ref{assump6}, \ref{assump7} and \ref{assump8} hold. For any $0<\delta<1$, $\alpha \in (0,1) \cup (1,\infty)$, and any $Q \in \mathcal{P}(\mathcal{W})$ such that $Q \ll P$ with probability at least $1-\delta$ over the randomness of $S_{[n]}$ and $(A_{i,j})_{n \times n}$, we have
    \begin{equation}\label{bound3}
    \begin{aligned}
        & \mathbb{E}_{W \sim Q} \big[ R(W,S_{[n]},(A_{i,j})_{n \times n}) \big] \\
        \leq & \mathbb{E}_{W \sim Q} \big[ \widehat{R}(W,S_{[n]},(A_{i,j})_{n \times n}) \big] + 3\sqrt{\frac{(2 c_x c^2_w c^2_a L L^2_{\phi} + M \Vert \widetilde{\bm{\Gamma}} \Vert_{\infty})^2({\rm D}_{\alpha}(Q||P)+\log(2\sqrt{2n}/\delta) ) \Vert \bm{\Gamma} \Vert^2}{(2n-1)}}\\
        & + M {\rm D}_{\rm TV} \left( P_{A_{n+1,[n+1]} | (A_{i,j})_{n \times n}}, Q_{A_{n+1,[n+1]} | (A_{i,j})_{n \times n}} \right) + \frac{M}{n} \sum_{i=1}^n {\rm D}_{\rm TV} \left( P_{S_{n+1}|S_{[n]}}, P_{S_i |S_{[i-1]}} \right) \\
        & + \frac{M}{n} \sum_{i=1}^n \mathbb{E}_S \left[ {\rm D}_{\rm TV} \Big( P_{S_{[i+1,n]}|S_{[i-1]}} \otimes P_{S_i |S_{[i-1]}}, P_{S_{[i,n]} | S_{[i-1]}} \Big) \right] + \frac{c_x c^2_w L L^2_{\phi} \Vert (\hat{A}_{ij})_{n \times n} \Vert^2_F}{n}.
    \end{aligned}
    \end{equation}
\end{theorem}

Generally, the result presented in Theorem~\ref{one_layer_bound} is similar to that for one-layer GCN model, yet there still exists some important differences. First, the bound in Theorem~\ref{two_layer_bound} requires more strict assumptions to guarantee a meaningful convergence tendency. Specifically, in order to guarantee the slack term Eq.~(\ref{bound3}) converge to zero as $n$ goes to infinity, the quantity $\Vert (\hat{A}_{ij})_{n \times n} \Vert^2_F$ should be of order $o(n)$. Although this condition may not be hold for all graphs, yet we demonstrate that it can be satisfied for a specific kind of social networks. In this kind of graphs, there are only two types of nodes: hub nodes $\{v^h_i \}_{i=1}^k$ and leaf nodes $\{v^l_i \}_{i=k+1}^n$. The number of hub nodes and the number of neighbors of leaf nodes are constants relative to $n$, that is, $k = \mathcal{O}(1)$. The hub nodes are connected to the vast majority of nodes in this graph. According to the type of nodes they connected, the edges in this graph can be divided into three categories: hub-hub edges, hub-leaf edges, and leaf-leaf edges. Their numbers are denoted by $m_{hh}$, $m_{hl}$, and $m_{ll}$, respectively. Let $d^{\rm max}_h = \max_{i \in \{1,\ldots,k\}} d_i $ and $d^{\rm min}_h = \max_{i \in \{1,\ldots,k\}} d_i $ and $d_l = \min_{i \in \{k+1,\ldots,n\}} d_i $ be the maximum and minimum degree of hub nodes, respectively. Let $d^{\rm min}_l = \max_{i \in \{k+1,\ldots,n\}} d_i $ be the minimum degree of leaf nodes. Then we have
\begin{equation}
\begin{aligned}
    \Vert (\hat{A}_{ij})_{n \times n} \Vert^2_F = & \sum_{i, j \in \{1,\ldots,k\}} \frac{2}{d_i d_j} + \sum_{i \in \{1,\ldots,k\}, j \in \{k+1,\ldots,n\}} \frac{2}{d_i d_j} + \sum_{i, j \in \{k+1,\ldots,n\}} \frac{2}{d_i d_j} \\
    \leq & \frac{2m_{hh}}{(d^{\rm min}_h)^2} + \frac{2m_{hl}}{d^{\rm min}_h d^{\rm min}_l} + \frac{2m_{ll}}{(d^{\rm min}_l)^2} \leq \frac{2k^2}{(d^{\rm min}_h)^2} + \frac{2 k d^{\rm max}_h} {d^{\rm min}_h d^{\rm min}_l} + \frac{2m_{ll}}{(d^{\rm min}_l)^2}.
\end{aligned}
\end{equation}
Using the assumptions that $d^{\rm max}_h = d^{\rm min}_h = \mathcal{O}(n)$ and $d^{\rm min}_l =  \mathcal{O}(1)$, we have $\Vert (\hat{a}_{ij})_{n \times n} \Vert^2_F = \mathcal{O}(m_{ll})$. Thus, if $m_{ll} = o(n)$, for example, $m_{ll} = \Theta(\sqrt{n})$ or $m_{ll} = \Theta(\log n)$, the term $\Vert (\hat{A}_{ij})_{n \times n} \Vert^2_F/n$ can converge to zero as $n$ tends to infinity. 

\section{Discussion and Conclusion}\label{dis_con}

In this paper, we introduce the mathematical framework for inductive node classification, and then analyze the generalization gap of one-layer and two-layer GCN models via PAC-Bayesian theory. Despite being preliminary, our results provide initial insights into the key factors that impact the generalization behavior of GCNs on inductive node classification. We outline several promising directions for future investigation:
\begin{itemize}
    \item \textbf{Problem setting and assumptions.} In our problem setting, the aggregation coefficients of existing nodes are fixed and they are treated as constants during the theoretical analysis. Besides, we assume that the aggregation coefficients of the new coming node is independent of the node features. Although these configuration and assumptions facilitates analytical simplification, its applicability might be limited for particular graph neural network architectures where the aggregation coefficients depend on node features, e.g., graph attention networks or graph transformers. Future works could explore whether it is possible to extend our results to a more general setting where all aggregation coefficients are random variables depend on node features.
    \item \textbf{Characterization of data dependencies}. How to model and characterize data dependency is a crucial step in analyzing the generalization performance of models. The approach we adopt in this work is inspired by \cite{Kontorovich2008, Kontorovich2017}, where data dependencies are characterized by a Wasserstein matrix. In addition, there are other methods for characterizing data dependencies, such as fractional chromatic number \citep{Janson2004} or forest complexity \citep{Zhang2019} for graph-dependent data. Studying how to extend our results to graph-dependent data using these tools is worth exploring.
    \item \textbf{Proof technique.} The core of our proof technique lies in decomposing the generalization gap into multiple components and systematically aligning them with corresponding terms in the generalization analysis of non i.i.d. data without graph structure \citep{Kuznetsov2015,Kuznetsov2020}. A natural question is whether we can develop more streamlined approaches to derive an upper bound the generalization gap that circumvent problematic terms like the adjacency matrix norm, which could make results apply to more general graph-structured data.
\end{itemize}

Overall, we believe this work represents a new starting point in unraveling the generalization behavior of graph neural networks. We hope the discussions presented above, along with the ideas and methods provided in this paper, will offer valuable insights for future research.

\newpage
\appendix

\section{Proofs}
\subsection{Proof of Theorem~\ref{main_pro}}
For any fixed realization $W=w$ and $i \in [n]$, since $\Psi(w,s)$ is $c_i$-Lipschitz in the $i$-th coordinate of its second argument under Hamming metric, for any $x, z \in \Omega$ such that $x^{[n]\setminus \{ i \}} = z^{[n]\setminus \{ i \}}$, we have
\begin{equation}
    \left\vert \Psi(w,x) - \Psi(w,z) \right\vert \leq c_i, \ i \in [n].
\end{equation}
Denote by $\bm{c} = (c_1, \ldots, c_n)^\top \in \mathbb{R}^n$ the vector and $\bm{\Gamma} \in \mathbb{R}^{n \times n}$ the matrix whose $(i,j)$-th entry is
\begin{equation}
\begin{aligned}
\Gamma_{i,j} = \begin{cases}
    0, & i > j \\
    1, & i = j \\
    \mathop{\rm sup}_{x, z \in \Omega, x^{[n]\setminus \{i\}} = z^{[n]\setminus \{i\}}} {\rm D}_{\rm TV} \left( P_{[j,n]} \big( \cdot {\big |} x_{[i]} \big) - P_{[j,n]} \big(\cdot {\big |} z_{[i]} \big) \right), & i<j
\end{cases}.
\end{aligned}
\end{equation}
Let $S'$ be the independent copy of $S$, that is, $S'$ is independent of $S$ and has the same distribution as $S$. By Theorem~4.1 and Theorem~4.2 of \cite{Kontorovich2017}, for any $\lambda \in \mathbb{R}$, $\alpha \in (0,1) \cup (1,\infty)$, and fixed realization $W=w$:
\begin{equation}\label{final_1}
    \mathbb{E}_S \left[ e^{\lambda \alpha \left(\Psi(w, S) - \mathbb{E}_{S'} [ \Psi(w, S') ] \right)} \right] \leq \exp \left\{ \frac{\lambda^2 \alpha^2 \Vert \bm{\Gamma} \bm{c} \Vert^2}{8} \right\}.
\end{equation}
Denote by $Q$ and $P$ the data-dependent posterior distribution and data-independent prior distribution on $\mathcal{W}$, respectively. 
Then we have
\begin{equation}
\begin{aligned}
    & \mathbb{E}_{W\sim P} \mathbb{E}_S \left[ e^{\lambda \alpha \left(\Psi(W,S) - \mathbb{E}_{S'} [ \Psi(W, S') ] \right)} \right] \\
    = & \int_w \mathbb{E}_S \left[ e^{\lambda \alpha \left( \Psi(w,S) - \mathbb{E}_{S'} [ \Psi(w, S') ] \right)} \right] \dif P(w) \\
    \leq & \int_w e^{\frac{\lambda^2 \alpha^2 \left\Vert \bm{\Gamma} \bm{c} \right\Vert^2}{8}} \dif P(w) = e^{\frac{\lambda^2 \alpha^2 \left\Vert \bm{\Gamma} \bm{c} \right\Vert^2}{8}}.
\end{aligned}
\end{equation}
By the variational representations of R\'{e}nyi divergence and Markov's inequality, for any $\lambda > 0$ and $\alpha \in (0,1) \cup (1,\infty)$:
\begin{equation}
\begin{aligned}
    & \mathbb{P}_S \left( \mathop{\rm sup}_{Q \in \mathcal{P}(\mathcal{W})} \left\{ \frac{\alpha}{\alpha-1} \log \mathbb{E}_{W \sim Q} \left[ e^{\lambda (\alpha - 1) \left(\Psi(W,S) - \mathbb{E}_{S'} [ \Psi(W, S') ] \right)} \right] - {\rm D}_{\alpha}(Q||P) \right\} - \frac{\lambda^2 \alpha^2 \left\Vert \bm{\Gamma} \bm{c} \right\Vert^2}{8} \geq \log (1/\delta) \right) \\
    = & \mathbb{P}_S \left( \exp \left\{ \mathop{\rm sup}_{Q \in \mathcal{P}(\mathcal{W})} \left\{ \frac{\alpha}{\alpha-1} \log \mathbb{E}_{W \sim Q} \left[ e^{\lambda (\alpha - 1) \left(\Psi(W,S) - \mathbb{E}_{S'} [ \Psi(W, S') ] \right)} \right] - {\rm D}_{\alpha}(Q||P) \right\} - \frac{\lambda^2 \alpha^2 \left\Vert \bm{\Gamma} \bm{c} \right\Vert^2 }{8} \right\} \geq 1/\delta \right) \\
    \leq & \delta \mathbb{E}_S \left[ \exp \left\{ \mathop{\rm sup}_{Q \in \mathcal{P}(\mathcal{W})} \left\{ \frac{\alpha}{\alpha-1} \log \mathbb{E}_{W \sim Q} \left[ e^{\lambda (\alpha - 1) \left(\Psi(W,S) - \mathbb{E}_{S'} [ \Psi(W, S') ] \right)} \right] - {\rm D}_{\alpha}(Q||P) \right\} - \frac{\lambda^2 \alpha^2 \left\Vert \bm{\Gamma} \bm{c} \right\Vert^2}{8} \right\} \right] \\
    = & \delta e^{ - \frac{\lambda^2 \alpha^2 \left\Vert \bm{\Gamma} \bm{c} \right\Vert^2}{8}} \mathbb{E}_S \left[ \exp \left\{ \mathop{\rm sup}_{Q \in \mathcal{P}(\mathcal{W})} \left\{ \frac{\alpha}{\alpha-1} \log \mathbb{E}_{W \sim Q} \left[ e^{\lambda (\alpha - 1) \left(\Psi(W,S) - \mathbb{E}_{S'} [ \Psi(W, S') ] \right)} \right] - {\rm D}_{\alpha}(Q||P) \right\} \right\} \right] \\
    = & \delta e^{ - \frac{\lambda^2 \alpha^2 \left\Vert \bm{\Gamma} \bm{c} \right\Vert^2}{8}} \mathbb{E}_S \left[ \mathbb{E}_{W \sim P} \left[ e^{\lambda \alpha \left(\Psi(W,S) - \mathbb{E}_{S'} [ \Psi(w, S') ] \right)} \right] \right] \\
    = & \delta e^{ - \frac{\lambda^2 \alpha^2 \left\Vert \bm{\Gamma} \bm{c} \right\Vert^2}{8}} \mathbb{E}_{W \sim P} \left[ \mathbb{E}_S \left[ e^{\lambda \alpha \left(\Psi(W,S) - \mathbb{E}_{S'} [ \Psi(W, S') ] \right)} \right] \right] \leq \delta.
\end{aligned}
\end{equation}
By Jensen's inequality,
\begin{equation}
    \log \mathbb{E}_{W \sim Q} \left[ e^{\lambda (\alpha - 1) \left(\Psi(W,S) - \mathbb{E}_{S'} [ \Psi(W, S') ] \right)} \right] \geq \lambda (\alpha-1) \mathbb{E}_{W \sim Q} \left[ \Psi(W,S) - \mathbb{E}_{S'} [ \Psi(w, S') ] \right].
\end{equation}
Therefore, for any $\lambda > 0$ and $\alpha \in (0,1) \cup (1,\infty)$, with probability at least $1-\delta$ over the randomness of $S$:
\begin{equation}
    \forall Q \in \mathcal{P}(\mathcal{W}): \mathbb{E}_{W \sim Q} \left[ \Psi(W,S) \right] \leq \mathbb{E}_{W \sim Q} \mathbb{E}_{S'} [ \Psi(W, S') ] + \frac{{\rm D}_{\alpha}(Q||P)+\log(1/\delta)}{\lambda \alpha} + \frac{\lambda \alpha \left\Vert \bm{\Gamma} \bm{c} \right\Vert^2}{8}.
\end{equation}
This finishes the proof. 

\subsection{Proof of Theorem~\ref{main_pro2}}
We start from Eq.~(\ref{final_1}). Using Theorem~2.6 (\uppercase\expandafter{\romannumeral4}) of \cite{Wainwright2019} and let $\lambda = 1- \frac{1}{2n} \in \left[ \frac{1}{2},1 \right)$, for any $w \in \mathcal{W}$ we have
\begin{equation}
    \mathbb{E}_S \left[ \exp \left\{ \frac{(2n-1)( \Psi(w,S) - \mathbb{E}_{S'} [ \Psi(w,S') ])^2}{n \Vert \bm{\Gamma} \bm{c} \Vert^2} \right\} \right] \leq \sqrt{2n}.
\end{equation}
Let $P$ be the data-independent prior distribution, we have
\begin{equation}
\begin{aligned}
    & \mathbb{E}_{W\sim P} \mathbb{E}_S \Bigg[ \exp \Bigg\{ \frac{(2n-1)(\Psi(W,S) - \mathbb{E}_{S'}[ \Psi(W,S') ])^2}{n \Vert \bm{\Gamma} \bm{c} \Vert^2} \Bigg\} \Bigg] \\
    = & \int_w \mathbb{E}_S \left[ \exp \left\{ \frac{(2n-1)( \Psi(w,S) - \mathbb{E}_{S'} [ \Psi(w,S')])^2}{n \Vert \bm{\Gamma} \bm{c} \Vert^2} \right\} \right] \dif P(w) \\
    \leq & \int_w \sqrt{2n} \dif P(w) = \sqrt{2n}.
\end{aligned}
\end{equation}
By the variational representations of R\'{e}nyi divergence and Markov's inequality, for any $\alpha \in (0,1) \cup (1,\infty)$:
\begin{equation}
\resizebox{0.99\hsize}{!}{$
\begin{aligned}
    & \mathbb{P}_S \left( \mathop{\rm sup}_{Q \in \mathcal{P}(\mathcal{W})} \left\{ \frac{\alpha}{\alpha-1} \log \mathbb{E}_{W \sim Q} \left[ \exp \left\{ \frac{ (\alpha - 1)(2n-1) (\Psi(W,S) - \mathbb{E}_{S'} [ \Psi(W,S') ] )^2}{\alpha n \Vert \bm{\Gamma} \bm{c} \Vert^2} \right\} \right] - {\rm D}_{\alpha}(Q||P) \right\} \geq \log \bigg( \frac{\sqrt{2n}}{\delta} \bigg) \right) \\
    = & \mathbb{P}_S \left( \exp \left\{ \mathop{\rm sup}_{Q \in \mathcal{P}(\mathcal{W})} \left\{ \frac{\alpha}{\alpha-1} \log \mathbb{E}_{W \sim Q} \left[ \exp \left\{ \frac{(\alpha - 1)(2n-1) (\Psi(W,S) - \mathbb{E}_{S'} [ \Psi(W,S') ] )^2}{\alpha n \Vert \bm{\Gamma} \bm{c} \Vert^2} \right\} \right] - {\rm D}_{\alpha}(Q||P) \right\} \right\} \geq \frac{\sqrt{2n}}{\delta} \right) \\
    \leq & \frac{\delta}{\sqrt{2n}} \mathbb{E}_S \left[ \exp \left\{ \mathop{\rm sup}_{Q \in \mathcal{P}(\mathcal{W})} \left\{ \frac{\alpha}{\alpha-1} \log \mathbb{E}_{W \sim Q} \left[ \exp \left\{ \frac{(\alpha - 1)(2n-1) (\Psi(W,S) - \mathbb{E}_{S'} [ \Psi(W,S') ] )^2}{\alpha n \Vert \bm{\Gamma} \bm{c} \Vert^2} \right\} \right] - {\rm D}_{\alpha}(Q||P) \right\} \right\} \right] \\
    = & \frac{\delta}{\sqrt{2n}} \mathbb{E}_S \left[ \mathbb{E}_{W \sim P} \left[ \exp \left\{\frac{(2n-1) (\Psi(W,S) - \mathbb{E}_{S'} [ \Psi(W,S') ] )^2}{ n \Vert \bm{\Gamma} \bm{c} \Vert^2} \right\} \right] \right] \\
    = & \frac{\delta}{\sqrt{2n}} \mathbb{E}_{W \sim P} \left[ \mathbb{E}_S \left[ \exp \left\{\frac{(2n-1) (\Psi(W,S) - \mathbb{E}_{S'} [ \Psi(W,S') ] )^2}{n \Vert \bm{\Gamma} \bm{c} \Vert^2} \right\} \right] \right] \leq \delta.
\end{aligned}$}
\end{equation}
By Jensen's inequality,
\begin{equation}
\begin{aligned}
    & \frac{\alpha}{\alpha-1} \log \mathbb{E}_{W \sim Q} \left[ \exp \left\{\frac{ (\alpha - 1)(2n-1) (\Psi(W,S) - \mathbb{E}_{S'} [ \Psi(W,S') ] )^2}{\alpha n \Vert \bm{\Gamma} \bm{c} \Vert^2} \right\} \right] \\
    \geq & \mathbb{E}_{W \sim Q} \left[ \frac{(2n-1) (\Psi(W,S) - \mathbb{E}_{S'} [ \Psi(W,S') ] )^2}{n \Vert \bm{\Gamma} \bm{c} \Vert^2} \right] \\
    \geq & \frac{(2n-1) ( \mathbb{E}_{W \sim Q} \left[ \Psi(W,S) \right] - \mathbb{E}_{W \sim Q} \mathbb{E}_{S'} [ \Psi(W,S') ] )^2}{n \Vert \bm{\Gamma} \bm{c} \Vert^2}.
\end{aligned}
\end{equation}
Thus, with probability at least $1-\delta$ over the randomness of $S$:
\begin{equation}
    \forall Q \in \mathcal{P}(\mathcal{W}): \mathbb{E}_{W \sim Q} \left[ \Psi(W,S) \right] \leq \mathbb{E}_{W \sim Q} \mathbb{E}_{S'} [ \Psi(W,S') ] + \sqrt{\frac{n ( {\rm D}_{\alpha}(Q||P)+\log(\sqrt{2n}/\delta) ) \Vert \bm{\Gamma} \bm{c} \Vert^2}{(2n-1)}}.
\end{equation}
This finishes the proof of Eq~(\ref{main_pro_bound1}). Now we turn to the proof of Eq.~(\ref{main_pro_bound2}). By Boole’s inequality, for any $t > 0$, $\lambda > 0$ and fixed realization $W=w$:
\begin{equation}\label{final_5}
\begin{aligned}
    & \mathbb{P}_S \left( \left\vert \Psi(w,S) - \mathbb{E}_{S'} [ \Psi(W,S') ]\right\vert \geq t \right) \\
    \leq & \mathbb{P}_S \left( \Psi(w,S) - \mathbb{E}_{S'} [ \Psi(W,S') ] \geq t \right) + \mathbb{P}_S \left( \mathbb{E}_{S'} [ \Psi(W,S') ] - \Psi(w,S) \geq t \right) \\
    = & \mathbb{P}_S \left( \exp\{ \lambda(\Psi(w,S) - \mathbb{E}_{S'} [ \Psi(W,S') ]) \} \geq e^{\lambda t} \right) + \mathbb{P}_S \left( \exp\{ \lambda(\mathbb{E}_{S'} [ \Psi(W,S') ] - \Psi(w,S)) \} \geq e^{\lambda t} \right) \\
    \leq & \mathbb{E}_S \left[ \exp \{ \lambda( \Psi(w,S) - \mathbb{E}_{S'} [ \Psi(W,S') ]) - \lambda t \} \right] + \mathbb{E}_S \left[ \exp \{ \lambda (\mathbb{E}_{S'} [ \Psi(W,S') ] - \Psi(w,S)) - \lambda t \} \right] \\
    \leq & 2\exp \left\{ \frac{\lambda^2 \Vert \bm{\Gamma} \bm{c} \Vert^2}{8} - \lambda t \right\},
\end{aligned}
\end{equation}
where the last inequality is obtained by Eq.~(\ref{final_1}). Since the above inequality holds for any $\lambda > 0$, by minimizing the right hand side with respect to $\lambda$, for any $t>0$, $\alpha \in (0,1) \cup (1,\infty)$ and any fixed realization:
\begin{equation}
    \mathbb{P}_S \left( \left\vert \Psi(w,S) - \mathbb{E}_{S'} [ \Psi(W,S') ]\right\vert \geq t \right) \leq 2 \exp \left\{ - \frac{2t^2}{\left\Vert \bm{\Gamma} \bm{c} \right\Vert^2} \right\}.
\end{equation}
By Theorem~4 of \cite{Esposito2021}, for any constants $\alpha, \alpha', \gamma, \gamma'$ such that $\frac{1}{\alpha}+\frac{1}{\gamma}=\frac{1}{\alpha'}+\frac{1}{\gamma'}=1$ and any $t>0$, we have
\begin{equation}
\begin{aligned}
    & \mathbb{P}_{W,S} \left( \left\vert \Psi(W,S) - \mathbb{E}_{S'} [ \Psi(W,S') ] \right\vert > t \right) \\
    \leq & \mathbb{E}^{\frac{1}{\gamma'}}_{W} \left[ \left( \mathbb{P}_S \left( \left\vert \Psi(W,S) - \mathbb{E}_{S'} [ \Psi(W,S') ] \right\vert > t \right) \right)^{\frac{\gamma'}{\gamma}} \right] \mathbb{E}^{\frac{1}{\alpha'}}_W \left[ \mathbb{E}^{\frac{\alpha'}{\alpha}}_S \left[ \left( \frac{\dif P_{S,W}}{\dif P_S \dif P_W} \right)^\alpha \right] \right].
\end{aligned}
\end{equation}
Plugging Eq.~(\ref{final_5}) into the above inequality we get
\begin{equation}
\begin{aligned}
    & \mathbb{P}_{W,S} \left( \left\vert \Psi(W,S) - \mathbb{E}_{S'} [ \Psi(W,S') ] \right\vert > t \right) \\
    \leq & \mathbb{E}^{\frac{1}{\gamma'}}_{W} \left[ \left( \mathbb{P}_S \left( \left\vert \Psi(W,S) - \mathbb{E}_{S'} [ \Psi(W,S') ] \right\vert > t \right) \right)^{\frac{\gamma'}{\gamma}} \right] \mathbb{E}^{\frac{1}{\alpha'}}_W \left[ \mathbb{E}^{\frac{\alpha'}{\alpha}}_S \left[ \left( \frac{\dif P_{S,W}}{\dif P_S \dif P_W} \right)^\alpha \right] \right] \\
    = & \left( \int_w  \left( \mathbb{P}_S \left( \left\vert \Psi(w,S) - \mathbb{E}_{S'} [ \Psi(W,S') ] \right\vert > t \right) \right)^{\frac{\gamma'}{\gamma}} \dif P_W(w) \right)^{\frac{1}{\gamma'}} \mathbb{E}^{\frac{1}{\alpha'}}_W \left[ \mathbb{E}^{\frac{\alpha'}{\alpha}}_S \left[ \left( \frac{\dif P_{S,W}}{\dif P_S \dif P_W} \right)^\alpha \right] \right] \\
    \leq & \left( \int_w  \left( 2 \exp \left\{ - \frac{2t^2}{\left\Vert \bm{\Gamma} \bm{c} \right\Vert^2} \right\} \right)^{\frac{\gamma'}{\gamma}} \dif P_W(w) \right)^{\frac{1}{\gamma'}} \mathbb{E}^{\frac{1}{\alpha'}}_W \left[ \mathbb{E}^{\frac{\alpha'}{\alpha}}_S \left[ \left( \frac{\dif P_{S,W}}{\dif P_S \dif P_W} \right)^\alpha \right] \right] \\
    = & 2^{\frac{1}{\gamma}} \exp \left\{ - \frac{2t^2}{ \gamma \left\Vert \bm{\Gamma} \bm{c} \right\Vert^2} \right\} \mathbb{E}^{\frac{1}{\alpha'}}_W \left[ \mathbb{E}^{\frac{\alpha'}{\alpha}}_S \left[ \left( \frac{\dif P_{S,W}}{\dif P_S \dif P_W} \right)^\alpha \right] \right].
\end{aligned}
\end{equation}
Let $\alpha' \to 1$ we have
\begin{equation}
\begin{aligned}
    \mathbb{P}_{W,S} \left( \left\vert \Psi(W,S) - \mathbb{E}_{S'} \left[ \Psi(W,S') \right] \right\vert > t \right) \leq & 2^{\frac{1}{\gamma}} \exp \left\{ - \frac{2t^2}{ \gamma \left\Vert \bm{\Gamma} \bm{c} \right\Vert^2} \right\} \mathbb{E}_W \left[ \mathbb{E}^{\frac{1}{\alpha}}_S \left[ \left( \frac{\dif P_{S,W}}{\dif P_S \dif P_W} \right)^\alpha \right] \right] \\
    = & 2^{\frac{1}{\gamma}} \exp \left\{ - \frac{2t^2}{ \gamma \left\Vert \bm{\Gamma} \bm{c} \right\Vert^2} + \frac{\alpha-1}{\alpha} I_{\alpha}(S,W) \right\} \\
    = & 2^{\frac{\alpha-1}{\alpha}} \exp \left\{ \frac{\alpha-1}{\alpha} \left( I_{\alpha}(S,W) - \frac{2t^2}{ \left\Vert \bm{\Gamma} \bm{c} \right\Vert^2} \right) \right\}.
\end{aligned}
\end{equation}
Therefore, for any $0 < \delta < 1$, with probability at least $1-\delta$ over the randomness of $S$ and $W$:
\begin{equation}
\begin{aligned}
    \Psi(W,S) - \mathbb{E}_{S'} [ \Psi(W,S') ] \leq & \left\vert \Psi(W,S) - \mathbb{E}_{S'} [ \Psi(W,S') ] \right\vert \\
    \leq & \sqrt{\frac{ \left\Vert \bm{\Gamma} \bm{c} \right\Vert^2}{2}\left( I_{\alpha}(S,W) + \log 2 + \frac{\alpha}{\alpha-1} \log (1/\delta) \right)}.
\end{aligned}
\end{equation}
This finishes the proof of Eq.~(\ref{main_pro_bound2}).

\subsection{Proof of Proposition~\ref{pro1}}
Denote by $s^{(i)}_{[n]}=(s_1,\ldots,s_{i-1},s'_i,s_{i+1},\ldots,s_n)$ the sequence of data points that differs from $s=(s_1,\ldots,s_n)$ only in the $i$-th entry. For any fixed realization $(A_{i,j})_{n \times n} = (a_{i,j})_{n \times n}$, we have
\begin{equation}\label{bound_difference1}
\begin{aligned}
    & \left\vert \mathcal{E}_1(w, s_{[n]}, (a_{i,j})_{n \times n}) - \mathcal{E}_1(w, s^{(i)}_{[n]}, (a_{i,j})_{n \times n}) \right\vert \\
    \leq & \Bigg\vert \frac{1}{n} \sum_{j=1}^n \Big( \ell(f_{w,j}(x_{[n]}, (\tilde{a}_{i,j})_{n \times n}), y_j) - \ell(f_{w,j}(\widetilde{x}_{[n]}, (\tilde{a}_{i,j})_{n \times n}), \widetilde{y}_j) \Big) \Bigg\vert \\
    & + \Bigg \vert \frac{1}{n} \sum_{j=1}^n \Big( \mathbb{E}_{S_{[j,n]}} \big[ \ell(f_{w,j}((x_{[j-1]},X_{[j,n]}),(\tilde{a}_{i,j})_{n \times n}), Y_j) \big\vert S_{[j-1]}=s_{[j-1]} \big] \\
    & - \mathbb{E}_{S_{[j,n]}} \big[ \ell(f_{w,j}((\widetilde{x}_{[j-1]},X_{[j,n]}), (\tilde{a}_{i,j})_{n \times n}), Y_j) \big\vert S_{[j-1]}=\widetilde{s}_{[j-1]} \big] \Big) \Bigg\vert.
\end{aligned}
\end{equation}
We first analyze the term
\begin{equation}\label{bound_emp1}
    \Bigg\vert \frac{1}{n} \sum_{j=1}^n \Big( \ell(f_{w,j}(x_{[n]}, (\tilde{a}_{i,j})_{n \times n}), y_j) - \ell(f_{w,j}(\widetilde{x}_{[n]}, (\tilde{a}_{i,j})_{n \times n}), \widetilde{y}_j) \Big) \Bigg\vert.
\end{equation}
For any $j \in [n]$ such that $j \ne i$, we have $\widetilde{x}_j = x_j$ and $\widetilde{y}_j = y_j$. If $i \in \mathcal{N}(j)$, then
\begin{equation}
\begin{aligned}
    & \left\vert \ell(f_{w,j}(x_{[n]}, (\tilde{a}_{i,j})_{n \times n}), y_j) - \ell(f_{w,j}(\widetilde{x}_{[n]}, (\tilde{a}_{i,j})_{n \times n}), \widetilde{y}_j) \right\vert \\
    = & \left\vert \ell(f_{w,j}(x_{[n]}, (\tilde{a}_{i,j})_{n \times n}), y_j) - \ell(f_{w,j}(\widetilde{x}_{[n]}, (\tilde{a}_{i,j})_{n \times n}), y_j) \right\vert \\
    \leq & L \Bigg\Vert \sum_{k=1}^n \tilde{a}_{j, k} \left( x_k - \widetilde{x}_k \right) w \Bigg\Vert_2 = L \Bigg\Vert \sum_{k \in \mathcal{N}^+(j)} \tilde{a}_{j, k} \left( x_k - \widetilde{x}_k \right) w \Bigg\Vert_2 = L \left\Vert \tilde{a}_{j, i} \left( x_i - x'_i \right) w \right\Vert_2 \\
    \leq & c_w \tilde{a}_{j,i} L \left\Vert x_i - x'_i \right\Vert_2 \leq c_w \tilde{a}_{j,i} L \left( \left\Vert x_i \right\Vert_2 + \left\Vert x'_i \right\Vert_2 \right) \leq 2 c_x c_w \tilde{a}_{j,i} L.
\end{aligned}
\end{equation}
Otherwise, if $i \notin \mathcal{N}(j)$, we have $f_{w,j}(x_{[n]}, (\tilde{a}_{i,j})_{n \times n}) = f_{w,j}(x_{[n]}, (\tilde{a}_{i,j})_{n \times n})$ and thus 
\begin{equation}
\begin{aligned}
    & \left\vert \ell(f_{w,j}(x_{[n]}, (\tilde{a}_{i,j})_{n \times n}), y_j) - \ell(f_{w,j}(\widetilde{x}_{[n]}, (\tilde{a}_{i,j})_{n \times n}), \widetilde{y}_j) \right\vert \\
    = & \left\vert \ell(f_{w,j}(x_{[n]}, (\tilde{a}_{i,j})_{n \times n}), y_j) - \ell(f_{w,j}(\widetilde{x}_{[n]}, (\tilde{a}_{i,j})_{n \times n}), y_j) \right\vert = 0.
\end{aligned}
\end{equation}
Combining the above two cases and noticing that $\tilde{a}_{j,i} = 1$ if and only if $i \in \mathcal{N}(j)$, we obtain that for any $j \in [n]$ such that $j \ne i$:
\begin{equation}\label{bound_emp2}
    \left\vert \ell(f_{w,j}(x_{[n]}, (\tilde{a}_{i,j})_{n \times n}), y_j) - \ell(f_{w,j}(\widetilde{x}_{[n]}, (\tilde{a}_{i,j})_{n \times n}), \widetilde{y}_j) \right\vert \leq 2 c_x c_w \tilde{a}_{j,i} L.
\end{equation}
For the case that $j=i$, by the assumption that $\vert \ell(\cdot, \cdot) \vert \leq M$, we have
\begin{equation}\label{bound_emp3}
\begin{aligned}
    \left\vert \ell(f_{w,j}(x_{[n]}, \tilde{a}_{j,[n]}), y_i) - \ell(f_{w,j}(\widetilde{x}_{[n]}, \tilde{a}_{j,[n]}), \widetilde{y}_i) \right\vert \leq M.
\end{aligned}
\end{equation}
Plugging Eqs.~(\ref{bound_emp2}, \ref{bound_emp3}) into Eq.~(\ref{bound_emp1}) and applying triangle inequality, we have
\begin{equation}\label{bound_difference2}
\begin{aligned}
    & \Bigg\vert \frac{1}{n} \sum_{j=1}^n \Big( \ell(f_{w,j}(x_{[n]}, (\tilde{a}_{i,j})_{n \times n}), y_j) - \ell(f_{w,j}(\widetilde{x}_{[n]}, (\tilde{a}_{i,j})_{n \times n}), \widetilde{y}_j) \Big) \Bigg\vert \\
    \leq & \frac{1}{n} \sum_{j=1}^n \Big\vert \ell(f_{w,j}(x_{[n]}, (\tilde{a}_{i,j})_{n \times n}), y_j) - \ell(f_{w,j}(\widetilde{x}_{[n]}, (\tilde{a}_{i,j})_{n \times n}), \widetilde{y}_j) \Big\vert \\
    = & \frac{1}{n} \sum_{j \ne i} \Big\vert \ell(f_{w,j}(x_{[n]}, \tilde{a}_{j,[n]}), y_j) - \ell(f_{w,j}(\widetilde{x}_{[n]}, \tilde{a}_{j,[n]}), \widetilde{y}_j) \Big\vert + \frac{\left\vert \ell(f_{w,i}(x_{[n]}, (\tilde{a}_{i,[n]})), y_i) - \ell(f_{w,i}(\widetilde{x}_{[n]}, \tilde{a}_{j,[n]}), \widetilde{y}_i) \right\vert}{n} \\
    \leq & \frac{2 c_x c_w L}{n} \sum_{j \ne i} \tilde{a}_{j,i} + \frac{M}{n} \leq \frac{2 c_x c_w c_a L + M}{n},
\end{aligned}
\end{equation}
where the last inequality is obtained by
\begin{equation}
    \sum_{j \ne i} \tilde{a}_{j,i} \leq \sum_{j=1}^n \tilde{a}_{j,i} = \sum_{j=1}^n \tilde{a}_{i,j} \leq \left\Vert (\tilde{a}_{ij})_{n \times n} \right\Vert_{\infty} \leq c_a.
\end{equation}
Now we analyze the term
\begin{equation}
\begin{aligned}
    \Bigg \vert & \frac{1}{n} \sum_{j=1}^n \Big( \mathbb{E}_{S_{[j,n]}} \big[ \ell(f_{w,j}((x_{[j-1]},X_{[j,n]}),(\tilde{a}_{i,j})_{n \times n}), Y_j) \big\vert S_{[j-1]}=s_{[j-1]} \big] \\
    & - \mathbb{E}_{S_{[j,n]}} \big[ \ell(f_{w,j}((\widetilde{x}_{[j-1]},X_{[j,n]}), (\tilde{a}_{i,j})_{n \times n}), Y_j) \big\vert S_{[j-1]}=\widetilde{s}_{[j-1]} \big] \Big) \Bigg\vert.
\end{aligned}
\end{equation}
If $i = n$, then we have $\widetilde{s}_j = s_j$ for $j \in [n-1]$, which implies that $P_{S_{[j,n]} | S_{[j-1]}=s_{[j-1]}} = P_{S_{[j,n]} | S_{[j-1]}=\widetilde{s}_{[j-1]}}$ for each $j \in [n-1]$. Then we have
\begin{equation}\label{bound_difference3}
\begin{aligned}
    & \Bigg \vert \frac{1}{n} \sum_{j=1}^n \Big( \mathbb{E}_{S_{[j,n]}} \big[ \ell(f_{w,j}((x_{[j-1]},X_{[j,n]}),(\tilde{a}_{i,j})_{n \times n}), Y_j) \big\vert S_{[j-1]}=s_{[j-1]} \big] \\
    & - \mathbb{E}_{S_{[j,n]}} \big[ \ell(f_{w,j}((\widetilde{x}_{[j-1]},X_{[j,n]}), (\tilde{a}_{i,j})_{n \times n}), Y_j) \big\vert S_{[j-1]}=\widetilde{s}_{[j-1]} \big] \Big) \Bigg\vert \\
    = & \Bigg \vert \frac{1}{n} \sum_{j=1}^n \Big( \mathbb{E}_{S_{[j,n]}} \big[ \ell(f_{w,j}((x_{[j-1]},X_{[j,n]}),(\tilde{a}_{i,j})_{n \times n}), Y_j) \big\vert | S_{[j-1]}=s_{[j-1]} \big] \\
    & - \mathbb{E}_{S_{[j,n]}} \big[ \ell(f_{w,j}((x_{[j-1]},X_{[j,n]}), (\tilde{a}_{i,j})_{n \times n}), Y_j) \big\vert S_{[j-1]}=s_{[j-1]} \big] \Big) \Bigg\vert = 0.
\end{aligned}
\end{equation}
If $i \ne n$, then we have $\widetilde{s}_j = s_j$ for $j \in [i-1]$, which implies that $P_{S_{[j,n]} | S_{[j-1]}=s_{[j-1]}} = P_{S_{[j,n]} | S_{[j-1]}=\widetilde{s}_{[j-1]}}$ for each $j \in [i-1]$. Now we further decompose this term into the following two terms
\begin{equation}\label{bound_difference4}
\begin{aligned}
    & \Bigg\vert \frac{1}{n} \sum_{j=1}^n \Big( \mathbb{E}_{ S_{[j,n]} } \big[ \ell(f_{w,j}((x_{[j-1]},X_{[j,n]}), (\tilde{a}_{i,j})_{n \times n}), Y_j) \big\vert S_{[j-1]}=s_{[j-1]} \big] \\
    & - \mathbb{E}_{ S_{[j,n]}} \big[ \ell(f_{w,j}((\widetilde{x}_{[j-1]},X_{[j,n]}), (\tilde{a}_{i,j})_{n \times n}), Y_j) \big\vert S_{[j-1]}=\widetilde{s}_{[j-1]} \big] \Big) \Bigg\vert \\
    \leq & \Bigg\vert \frac{1}{n} \sum_{j=1}^i \Big( \mathbb{E}_{ S_{[j,n]} } \big[ \ell(f_{w,j}((x_{[j-1]},X_{[j,n]}), (\tilde{a}_{i,j})_{n \times n}), Y_j) \big\vert S_{[j-1]}=s_{[j-1]} \big] \\
    & - \mathbb{E}_{ S_{[j,n]}} \big[ \ell(f_{w,j}((\widetilde{x}_{[j-1]},X_{[j,n]}), (\tilde{a}_{i,j})_{n \times n}), Y_j) \big\vert S_{[j-1]}=\widetilde{s}_{[j-1]} \big] \Big) \Bigg\vert \\
    & + \Bigg\vert \frac{1}{n} \sum_{j=i+1}^n \Big( \mathbb{E}_{ S_{[j,n]} } \big[ \ell(f_{w,j}((x_{[j-1]},X_{[j,n]}), (\tilde{a}_{i,j})_{n \times n}), Y_j) \big\vert S_{[j-1]}=s_{[j-1]} \big] \\
    & - \mathbb{E}_{ S_{[j,n]}} \big[ \ell(f_{w,j}((\widetilde{x}_{[j-1]},X_{[j,n]}), (\tilde{a}_{i,j})_{n \times n}), Y_j) \big\vert S_{[j-1]}=\widetilde{s}_{[j-1]} \big] \Big) \Bigg\vert.
\end{aligned}
\end{equation}
It can be verified that the first term equals zero, thus it is sufficient to analyze the second term. For any $j \in [i+1, n]$, we have
\begin{equation}\label{bound_difference5}
\begin{aligned}
    & \Big\vert \mathbb{E}_{ S_{[j,n]}} \big[ \ell(f_{w,j}((x_{[j-1]},X_{[j,n]}), (\tilde{a}_{i,j})_{n \times n}), Y_{j}) \big\vert S_{[j-1]}=s_{[j-1]} \big] \\
    & - \mathbb{E}_{ S_{[j,n]}} \big[ \ell(f_{w,j}((\widetilde{x}_{[j-1]},X_{[j,n]}), (\tilde{a}_{i,j})_{n \times n}), Y_j) \big\vert S_{[j-1]}=\widetilde{s}_{[j-1]} \big] \Big\vert \\
    = & \Big\vert \mathbb{E}_{ S_{[j,n]}} \big[ \ell(f_{w,j}((x_{[i-1]},x_i,x_{(i,j-1]},X_{[j,n]}), (\tilde{a}_{i,j})_{n \times n}), Y_{j}) \big\vert S_{[i-1]}=s_{[i-1]}, S_{i}=s_i, S_{(i,j-1]}=s_{(i,j-1]} \big] \\
    & - \mathbb{E}_{ S_{[j,n]}} \big[ \ell(f_{w,j}((x_{[i-1]},x'_i,x_{(i,j-1]},X_{[j,n]}), (\tilde{a}_{i,j})_{n \times n}), Y_{j}) \big\vert S_{[i-1]}=s_{[i-1]}, S_{i}=s'_i, S_{(i,j-1]}=s_{(i,j-1]} \big] \Big\vert \\
    \leq & \Big\vert \mathbb{E}_{ S_{[j,n]}} \big[ \ell(f_{w,j}((x_{[i-1]},x_i,x_{(i,j-1]},X_{[j,n]}), (\tilde{a}_{i,j})_{n \times n}), Y_{j}) \big\vert S_{[i-1]}=s_{[i-1]}, S_{i}=s_i, S_{(i,j-1]}=s_{(i,j-1]} \big] \\
    & - \mathbb{E}_{ S_{[j,n]}} \big[ \ell(f_{w,j}((x_{[i-1]},x_i,x_{(i,j-1]},X_{[j,n]}), (\tilde{a}_{i,j})_{n \times n}), Y_{j}) \big\vert S_{[i-1]}=s_{[i-1]}, S_{i}=s'_i, S_{(i,j-1]}=s_{(i,j-1]} \big] \Big\vert \\
    & + \Big\vert \mathbb{E}_{ S_{[j,n]}} \big[ \ell(f_{w,j}((x_{[i-1]},x_i,x_{(i,j-1]},X_{[j,n]}), (\tilde{a}_{i,j})_{n \times n}), Y_{j}) \big\vert S_{[i-1]}=s_{[i-1]}, S_{i}=s'_i, S_{(i,j-1]}=s_{(i,j-1]} \big] \\
    & - \mathbb{E}_{ S_{[j,n]}} \big[ \ell(f_{w,j}((x_{[i-1]},x'_i,x_{(i,j-1]},X_{[j,n]}), (\tilde{a}_{i,j})_{n \times n}), Y_{j}) \big\vert S_{[i-1]}=s_{[i-1]}, S_{i}=s'_i, S_{(i,j-1]}=s_{(i,j-1]} \big] \Big\vert,
\end{aligned}
\end{equation}
where we have stipulated that $(i,j-1] \triangleq \varnothing$ if $j=i+1$. By the definition of total variation distance, for any $j \in (i,n]$ we have
\begin{equation}\label{bound_difference6}
\begin{aligned}
    & \Big\vert \mathbb{E}_{ S_{[j,n]}} \big[ \ell(f_{w,j}((x_{[i-1]},x_i,x_{(i,j-1]},X_{[j,n]}), (\tilde{a}_{i,j})_{n \times n}), Y_{j}) \big\vert S_{[i-1]}=s_{[i-1]}, S_{i}=s_i, S_{(i,j-1]}=s_{(i,j-1]} \big] \\
    & - \mathbb{E}_{ S_{[j,n]}} \big[ \ell(f_{w,j}((x_{[i-1]},x_i,x_{(i,j-1]},X_{[j,n]}), (\tilde{a}_{i,j})_{n \times n}), Y_{j}) \big\vert S_{[i-1]}=s_{[i-1]}, S_{i}=s'_i, S_{(i,j-1]}=s_{(i,j-1]} \big] \Big\vert \\
    = & \bigg\vert \int_{s_{[j,n]}} \ell(f_{w,j}((x_{[j-1]},x_{[j,n]}), (\tilde{a}_{i,j})_{n \times n}), y_{j}) \dif P_{S_{[j,n]} | S_{[i-1]}=s_{[i-1]}, S_{i}=s_i, S_{(i,j-1]}=s_{(i,j-1]}}(s_{[j,n]}) \\
    & - \int_{s_{[j,n]}} \ell(f_{w,j}((x_{[j-1]},x_{[j,n]}), (\tilde{a}_{i,j})_{n \times n}), y_{j}) \dif P_{S_{[j,n]} | S_{[i-1]}=s_{[i-1]}, S_{i}=s'_i, S_{(i,j-1]}=s_{(i,j-1]}}(s_{[j,n]}) \bigg\vert \\
    \leq & M {\rm D}_{\rm TV} \left( P_{S_{[j,n]} | S_{[i-1]}=s_{[i-1]}, S_{i}=s_i, S_{(i,j-1]}=s_{(i,j-1]}}, P_{S_{[j,n]} | S_{[i-1]}=s_{[i-1]}, S_{i}=s'_i, S_{(i,j-1]}=s_{(i,j-1]}} \right) \\
    \leq & M \sup_{s_i, s'_i} {\rm D}_{\rm TV} \left( P_{S_{[j,n]} | S_{[i-1]}=s_{[i-1]}, S_{i}=s_i, S_{(i,j-1]}=s_{(i,j-1]}}, P_{S_{[j,n]} | S_{[i-1]}=s_{[i-1]}, S_{i}=s'_i, S_{(i,j-1]}=s_{(i,j-1]}} \right).
\end{aligned}
\end{equation}
For any $j \in (i,n]$ and fixed realization $S_{[j,n]}=s_{[j,n]}$, following the procedure of deriving Eq.~(\ref{bound_emp2}), we have
\begin{equation}
\begin{aligned}
    & \left\vert \ell(f_{w,j}((x_{[i-1]},x_i,x_{(i,j-1]},x_{[j,n]}), (\tilde{a}_{i,j})_{n \times n}), y_j) - \ell(f_{w,j}((x_{[i-1]},x'_i,x_{(i,j-1]},x_{[j,n]}), (\tilde{a}_{i,j})_{n \times n}), y_j) \right\vert \\
    \leq & 2 c_x c_w \tilde{a}_{j,i} L,
\end{aligned}
\end{equation}
which implies that
\begin{equation}\label{bound_difference7}
\begin{aligned}
    & \Big\vert \mathbb{E}_{ S_{[j,n]} | S_{[i-1]}=s_{[i-1]}, S_{i}=s'_i, S_{(i,j-1]}=s_{(i,j-1]}} \left[ \ell(f_{w,j}((x_{[i-1]},x_i,x_{(i,j-1]},X_{[j,n]}), \tilde{a}_{i+1,[n]}), Y_{j}) \right] \\
    & - \mathbb{E}_{ S_{[j,n]} | S_{[i-1]}=s_{[i-1]}, S_{i}=s'_i, S_{(i,j-1]}=s_{(i,j-1]}} \left[ \ell(f_{w,j}((x_{[i-1]},x'_i,x_{(i,j-1]},X_{[j,n]}), \tilde{a}_{i+1,[n]}), Y_{j}) \right] \Big\vert \\
    \leq & \int_{s_{[j,n]}} \big\vert \ell(f_w((x_{[i-1]},x_i,x_{(i,j-1]},x_{[j,n]}), \tilde{a}_{i+1,[n]}), y_{j}) \\
    & - \ell(f_w((x_{[i-1]},x'_i,x_{(i,j-1]},x_{[j,n]}), \tilde{a}_{i+1,[n]}), y_{j}) \big\vert \dif P_{S_{[j,n]} | S_{[i-1]}=s_{[i-1]}, S_{i}=s'_i, S_{(i,j-1]}=s_{(i,j-1]}}(s_{[j,n]}) \\
    \leq & 2 c_x c_w \tilde{a}_{j,i} L.
\end{aligned}
\end{equation}
Denote by $\widetilde{\bm{\Gamma}} \in \mathbb{R}^{n \times n}$ a matrix such that
\begin{equation}
    \widetilde{\Gamma}_{i,j} = \sup_{s_i, s'_i} {\rm D}_{\rm TV} \left( P_{S_{[j,n]} | S_{[i-1]}=s_{[i-1]}, S_{i}=s_i, S_{(i,j-1]}=s_{(i,j-1]}}, P_{S_{[j,n]} | S_{[i-1]}=s_{[i-1]}, S_{i}=s'_i, S_{(i,j-1]}=s_{(i,j-1]}} \right)
\end{equation}
if $j > i$ and $i \in [n-1]$, otherwise $\widetilde{\Gamma}_{i,j} = 0$. Plugging Eq.~(\ref{bound_difference6}) and Eq.~(\ref{bound_difference7}) into Eq.~(\ref{bound_difference5}), for any $j \in (i,n]$ we have
\begin{equation}\label{bound_difference8}
\begin{aligned}
    & \Big\vert \mathbb{E}_{ S_{[j,n]}} \big[ \ell(f_{w,j}((x_{[j-1]},X_{[j,n]}), (\tilde{a}_{i,j})_{n \times n}), Y_{j}) \big\vert S_{[j-1]}=s_{[j-1]} \big] \\
    & - \mathbb{E}_{ S_{[j,n]}} \big[ \ell(f_{w,j}((\widetilde{x}_{[j-1]},X_{[j,n]}), (\tilde{a}_{i,j})_{n \times n}), Y_j) \big\vert S_{[j-1]}=\widetilde{s}_{[j-1]} \big] \Big\vert \leq 2 c_x c_w \tilde{a}_{j,i} L + M \widetilde{\Gamma}_{i,j}.
\end{aligned}
\end{equation}
Combining Eq.~(\ref{bound_difference3}), Eq.~(\ref{bound_difference4}) and Eq.~(\ref{bound_difference8}), we obtain
\begin{equation}\label{bound_difference9}
\begin{aligned}
    & \Bigg \vert \frac{1}{n} \sum_{j=1}^n \Big( \mathbb{E}_{S_{[j,n]}} \big[ \ell(f_{w,j}((x_{[j-1]},X_{[j,n]}),(\tilde{a}_{i,j})_{n \times n}), Y_j) \big\vert S_{[j-1]}=s_{[j-1]} \big] \\
    & - \mathbb{E}_{S_{[j,n]}} \big[ \ell(f_{w,j}((\widetilde{x}_{[j-1]},X_{[j,n]}), (\tilde{a}_{i,j})_{n \times n}), Y_j) \big\vert S_{[j-1]}=\widetilde{s}_{[j-1]} \big] \Big) \Bigg\vert \\
    \leq & \frac{\mathds{1}\{i \ne n \}}{n} \sum_{j=i+1}^n \left( 2c_x c_w \tilde{a}_{j,i} L + M \widetilde{\Gamma}_{i,j} \right) = \frac{\mathds{1}\{i \ne n \}}{n} \bigg( 2 c_x c_w L \sum_{j=i+1}^n \tilde{a}_{i,j} + M \sum_{j=1}^n \widetilde{\Gamma}_{i,j} \bigg) \\
    \leq & \frac{\mathds{1}\{i \ne n \}}{n} \bigg( 2 c_x c_w L \sum_{j=1}^n \tilde{a}_{i,j} + M \sum_{j=1}^n \widetilde{\Gamma}_{i,j} \bigg) \leq \frac{\mathds{1}\{i \ne n \}}{n} \bigg( 2 c_x c_w c_a L + M \Vert \widetilde{\bm{\Gamma}} \Vert_{\infty} \bigg).
\end{aligned}
\end{equation}
Plugging Eqs.~(\ref{bound_difference2}, \ref{bound_difference9}) into Eq.~(\ref{bound_difference1}), for any fixed realization $W=w$, $(A_{i,j})_{n \times n} = (a_{i,j})_{n \times n}$ and any $i \in [n]$, we have
\begin{equation}\label{bound_difference10}
\begin{aligned}
    \left\vert \mathcal{E}_1(w, s_{[n]}, (a_{i,j})_{n \times n}) - \mathcal{E}_1(w, s^{(i)}_{[n]}, (a_{i,j})_{n \times n}) \right\vert \leq & \frac{2 c_x c_w c_a L + M}{n} + \frac{\mathds{1}\{i \ne n \}}{n} \bigg( 2 c_x c_w c_a L + M \Vert \widetilde{\bm{\Gamma}} \Vert_{\infty} \bigg) \\
    \leq & \frac{2(2 c_x c_w c_a L + M \max (1, \Vert \widetilde{\bm{\Gamma}} \Vert_{\infty}))}{n}.
\end{aligned}
\end{equation}
Now we are in a place to derive the final result by applying Eq.~(\ref{final_1}). For each $i \in [n]$, Let $c_i$ be the right hand side of Eq.~(\ref{bound_difference10}), we have
\begin{equation}
\begin{aligned}
    \Vert \bm{c} \Vert^2_2 = & \frac{4}{n^2} \sum_{i=1}^n \left( 2 c_x c_w c_a L + M \max (1, \Vert \widetilde{\bm{\Gamma}} \Vert_{\infty}) \right)^2 = \frac{4(2 c_x c_w c_a L + M \max (1, \Vert \widetilde{\bm{\Gamma}} \Vert_{\infty}))^2}{n}.
\end{aligned}
\end{equation}
Now using the inequality that $\Vert \bm{\Gamma} \bm{c} \Vert_2 \leq \Vert \bm{\Gamma} \Vert \Vert \bm{c} \Vert_2$ and applying Eq.~(\ref{final_1}), for any fixed realization $W=w$ and $(A_{i,j})_{n \times n} = (a_{i,j})_{n \times n}$:
\begin{equation}
    \mathbb{E}_{S_{[n]}|(A_{i,j})_{n \times n} = (a_{i,j})_{n \times n}} \left[ e^{\lambda \alpha \big(\mathcal{E}_1(w, S_{[n]}, (a_{i,j})_{n \times n}) - \mathbb{E}_{S'_{[n]}} [ \mathcal{E}_1(w, s_{[n]}, (a_{i,j})_{n \times n}) ] \big)} \right] \leq \exp \left\{ \frac{\lambda^2 \alpha^2 \Vert \bm{\Gamma} \Vert^2 \Vert \bm{c} \Vert^2}{8} \right\}.
\end{equation}
Taking expectation over $(A_{i,j})_{n \times n}$ on both side yields
\begin{equation}
    \mathbb{E}_{S_{[n]}, (A_{i,j})_{n \times n}} \left[ e^{\lambda \alpha \big(\mathcal{E}_1(w, S_{[n]}, (a_{i,j})_{n \times n}) - \mathbb{E}_{S'_{[n]}} [ \mathcal{E}_1(w, s_{[n]}, (a_{i,j})_{n \times n}) ] \big)} \right] \leq \exp \left\{ \frac{\lambda^2 \alpha^2 \Vert \bm{\Gamma} \Vert^2 \Vert \bm{c} \Vert^2}{8} \right\}.
\end{equation}
Following the process of deriving Eq~(\ref{main_pro_bound1}), for any $\delta \in (0,1)$, with probability at least $1-\delta/2$ over the randomness of $S_{[n]}$ and $(A_{i,j})_{n \times n}$:
\begin{equation}
    \mathbb{E}_{W \sim Q} \left[ \mathcal{E}_1(W,S_{[n]},(A_{i,j})_{n \times n}) \right] \leq 2\sqrt{\frac{(2 c_x c_w c_a L + M \max (1, \Vert \widetilde{\bm{\Gamma}} \Vert_{\infty}))^2({\rm D}_{\alpha}(Q||P)+\log(2\sqrt{2n}/\delta) ) \Vert \bm{\Gamma} \Vert^2}{(2n-1)}} .
\end{equation}
This finishes the proof.

\subsection{Proof of Proposition~\ref{pro2}}
As shown in Section~\ref{sec1}, for any fixed realization $W=w$ and $(A_{i,j})_{n \times n} = (a_{i,j})_{n \times n}$, we have
\begin{equation}
\begin{aligned}
    & \mathbb{E}_{S_{n+1}, A_{n+1,[n+1]}|S_{[n]},(a_{i,j})_{n \times n}} \big[ \ell(f_w(X_{[n+1]}, A_{n+1,[n+1]}),Y_{n+1}) \big] \\
    & - \frac{1}{n} \sum_{j=1}^n \mathbb{E}_{S_{[j,n]}} \big[ \ell(f_{W,j}(X_{[n]}, (\tilde{a}_{i,j})_{n \times n}), Y_j) \big\vert S_{[j-1]} \big] \\
    = & \mathbb{E}_{P_{A_{n+1,[n+1]} | (A_{i,j})_{n \times n}} = (a_{i,j})_{n \times n}} \mathbb{E}_{S_{n+1}|S_{[n]}} \big[ \ell(f_w(X_{[n+1]}, A_{n+1,[n+1]}),Y_{n+1}) \big] \\
    & - \mathbb{E}_{Q_{A_{n+1,[n+1]} | (A_{i,j})_{n \times n}} = (a_{i,j})_{n \times n}} \mathbb{E}_{S_{n+1}|S_{[n]}} \big[ \ell(f_w(X_{[n+1]}, A_{n+1,[n+1]}),Y_{n+1}) \big] \\
    & + \frac{1}{n} \sum_{j=1}^n \Big(\mathbb{E}_{S_{n+1}|S_{[n]}} \big[ \ell(f_w(X_{[n+1]}, (\tilde{a}_{j,[j-1]},0,\tilde{a}_{j,(j,n]},\tilde{a}_{j,j})),Y_{n+1}) \big] \\
    & - \mathbb{E}_{ S_{[j,n]} | S_{[j-1]}} \big[ \ell(f_{w,j}((X_{[j-1]}, X_{[j,n]}), (\tilde{a}_{i,j})_{n \times n}), Y_j) \big] \Big).
\end{aligned}
\end{equation}
By Assumption~\ref{assump6}, the first term can be bounded by
\begin{equation}\label{exp_phi1}
\begin{aligned}
    & \mathbb{E}_{P_{A_{n+1,[n+1]} | (A_{i,j})_{n \times n}} = (a_{i,j})_{n \times n}} \mathbb{E}_{S_{n+1}|S_{[n]}} \big[ \ell(f_w(X_{[n+1]}, A_{n+1,[n+1]}),Y_{n+1}) \big] \\
    & - \mathbb{E}_{Q_{A_{n+1,[n+1]} | (A_{i,j})_{n \times n}} = (a_{i,j})_{n \times n}} \mathbb{E}_{S_{n+1}|S_{[n]}} \big[ \ell(f_w(X_{[n+1]}, A_{n+1,[n+1]}),Y_{n+1}) \big] \\
    \leq & M {\rm D}_{\rm TV} \left( P_{A_{n+1,[n+1]} | (A_{i,j})_{n \times n}} = (a_{i,j})_{n \times n}, Q_{A_{n+1,[n+1]} | (A_{i,j})_{n \times n}} = (a_{i,j})_{n \times n} \right).
\end{aligned}
\end{equation}
For the second term, notice that
\begin{equation}\label{exp_phi2}
\begin{aligned}
    & \frac{1}{n} \sum_{j=1}^n \mathbb{E}_{S_{n+1}|S_{[n]}} \left[ \ell(f_w(X_{[n+1]}, (\tilde{a}_{j,[j-1]},0,\tilde{a}_{j,(j,n]},\tilde{a}_{j,j})),Y_{n+1}) \right] \\
    & - \frac{1}{n} \sum_{j=1}^n \mathbb{E}_{ S_{[j,n]} | S_{[j-1]}} \left[ \ell(f_{w,j}((X_{[j-1]}, X_{[j,n]}), (\tilde{a}_{i,j})_{n \times n}), Y_j) \right] \\
    = & \frac{1}{n} \sum_{j=1}^n \mathbb{E}_{S_{n+1}|S_{[n]}} \left[ \ell(f_w(X_{[n+1]}, (\tilde{a}_{j,[j-1]},0,\tilde{a}_{j,(j,n]},\tilde{a}_{j,j})),Y_{n+1}) \right] \\
    & - \frac{1}{n} \sum_{j=1}^n \mathbb{E}_{S_j |S_{[j-1]}} \left[ \ell(f_{w,j}((X_{[j]},X_{(j,n]}), (\tilde{a}_{i,j})_{n \times n}), Y_j) \right] \\
    & + \frac{1}{n} \sum_{j=1}^n \mathbb{E}_{S_j |S_{[j-1]}} \left[ \ell(f_{w,j}((X_{[j]},X_{(j,n]}), (\tilde{a}_{i,j})_{n \times n}), Y_j) \right] \\
    & - \frac{1}{n} \sum_{j=1}^n \mathbb{E}_{ S_{[j,n]} | S_{[j-1]}} \left[ \ell(f_{w,j}((X_{[j-1]}, X_{[j,n]}), (\tilde{a}_{i,j})_{n \times n}), Y_j) \right].
\end{aligned}
\end{equation}
For any fixed $j \in [n]$, by definition
\begin{equation}
\begin{aligned}
    & f_w((X_{[n+1]}, (\tilde{a}_{j,[j-1]},0,\tilde{a}_{j,(j,n]},\tilde{a}_{j,j})) = \sum_{k \ne j} \tilde{a}_{j,k} X_k w + \tilde{a}_{j,j} X_{n+1} w, \\
    & f_{w,j}((X_{[j]},X_{(j,n]}), (\tilde{a}_{i,j})_{n \times n}) = \sum_{k \ne j} \tilde{a}_{j,k} X_k w + \tilde{a}_{j,j} X_j w.
\end{aligned}
\end{equation}
For any fixed realization $S_{[n]} = s_{[n]}$ and any unknown data point $S = (X,Y)$, we define
\begin{equation}
    \hat{y}(X) = \sum_{k \ne j} \tilde{a}_{j,k} x_k w + \tilde{a}_{j,j} X w, \ h(S) = \ell(\hat{y}(X),Y).
\end{equation}
Then we have
\begin{equation}
\begin{aligned}
    & h(S_{n+1}) = \ell(f_w((x_{[n]},X_{n+1}), (\tilde{a}_{j,[j-1]},0,\tilde{a}_{j,(j,n]},\tilde{a}_{j,j})),Y_{n+1}), \\
    & h(S_j) = \ell(f_{w,j}((x_{[j-1]},X_j,x_{(j,n]}), (\tilde{a}_{i,j})_{n \times n}), Y_j).
\end{aligned}
\end{equation}
Then, for any fixed realization $S = s = ((x_1, y_1), \ldots, (x_n,y_n))$ and any $j \in [n]$, we have
\begin{equation}
\begin{aligned}
    & \mathbb{E}_{S_{n+1}|S_{[n]}=s_{[n]}} \left[ \ell(f_w((x_{[n]}, X_{n+1}), (\tilde{a}_{j,[j-1]},0,\tilde{a}_{j,(j,n]},\tilde{a}_{j,j})),Y_{n+1}) \right] \\
    & - \mathbb{E}_{S_j |S_{[j-1]} = s_{[j-1]}} \left[ \ell(f_w((x_{[j-1]},X_j,x_{(j,n]}), (\tilde{a}_{i,j})_{n \times n}),Y_j) \right] \\
    = & \mathbb{E}_{S_{n+1}|S_{[n]}=s_{[n]}} \left[ h(S_{n+1}) \right] - \mathbb{E}_{S_j |S_{[j-1]} = s_{[j-1]}} \left[ h(S_j) \right] \\
    = & \mathbb{E}_{S_{n+1}|S_{[n]}=s_{[n]}} \left[ h(S_{n+1}) \right] - \mathbb{E}_{S_j |S_{[j-1]} = s_{[j-1]}} \left[ h(S_j) \right] \\
    = & \int_{s_{n+1}} h(s_{n+1}) \dif P_{S_{n+1}|S_{[n]}=s_{[n]}}(s_{n+1}) - \int_{s_j} h(s_j) \dif P_{S_j | S_{[j-1]} = s_{[j-1]}}(s_j) \\
    \leq & \left\vert \int_{s_{n+1}} h(s_{n+1}) \dif P_{S_{n+1}|S_{[n]}=s_{[n]}}(s_{n+1}) - \int_{s_j} h(s_j) \dif P_{S_j | S_{[j-1]} = s_{[j-1]}}(s_j) \right\vert \\
    \leq & M {\rm D}_{\rm TV} \left( P_{S_{n+1}|S_{[n]}=s_{[n]}}, P_{S_j |S_{[j-1]} = s_{[j-1]}} \right),
\end{aligned}
\end{equation}
which implies that
\begin{equation}
\begin{aligned}
    & \mathbb{E}_{S_{n+1}|S} \left[ \ell(f_w((x_{[n]}, X_{n+1}), (\tilde{a}_{j,[j-1]},0,\tilde{a}_{j,(j,n]},\tilde{a}_{j,j})),Y_{n+1}) \right] \\
    & - \mathbb{E}_{S_j |S_{[j-1]}} \left[ \ell(f_{w,j}((x_{[j-1]},X'_j,x_{(j,n]}), (\tilde{a}_{i,j})_{n \times n}), Y_j) \right] \\
    \leq & M {\rm D}_{\rm TV} \left( P_{S_{n+1}|S_{[n]}}, P_{S_j |S_{[j-1]}} \right).
\end{aligned}
\end{equation}
Therefore, we have
\begin{equation}\label{exp_phi3}
\begin{aligned}
    & \frac{1}{n} \sum_{j=1}^n \mathbb{E}_{S_{n+1}|S} \left[ \ell(f_w(X_{[n+1]}, (\tilde{a}_{j,[j-1]},0,\tilde{a}_{j,(j,n]},\tilde{a}_{j,j})),Y_{n+1}) \right] \\
    & - \frac{1}{n} \sum_{j=1}^n \mathbb{E}_{S_j |S_{[j-1]}} \left[ \ell(f_{w,j}((X_{[j]},X_{(j,n]}), (\tilde{a}_{i,j})_{n \times n}), Y_j) \right] \\
    \leq & \frac{M}{n} \sum_{i=1}^n {\rm D}_{\rm TV} \left( P_{S_{n+1}|S_{[n]}}, P_{S_i |S_{[i-1]}} \right).
\end{aligned}
\end{equation}
Define
\begin{equation}
\begin{aligned}
    \widetilde{\mathcal{E}}_2(W,S_{[n]},(A_{i,j})_{n \times n}) = & \frac{1}{n} \sum_{j=1}^n \mathbb{E}_{S_j |S_{[j-1]}} \left[ \ell(f_{W,j}((X_{[j]},X_{(j,n]}), (\tilde{A}_{i,j})_{n \times n}), Y_j) \right] \\
    & - \frac{1}{n} \sum_{j=1}^n \mathbb{E}_{ S_{[j,n]} | S_{[j-1]}} \left[ \ell(f_{W,j}((X_{[j-1]}, X_{[j,n]}), (\tilde{A}_{i,j})_{n \times n}), Y_j) \right],
\end{aligned}
\end{equation}
our next step is applying the result in Theorem~\ref{main_pro2} to show that 
\begin{equation}
    \mathbb{E}_{W \sim Q} \left[ \widetilde{\mathcal{E}}_2(W,S_{[n]},(A_{i,j})_{n \times n}) \right]
\end{equation}
can be bounded by its expectation in high probability. To this end, we need to provide an upper bound for the term $\vert \widetilde{\mathcal{E}}_2(w, s_{[n]}, (a_{i,j})_{n \times n}) - \widetilde{\Psi}_2(w, s^{(i)}_{[n]}, (a_{i,j})_{n \times n}) \vert$ for any fixed realization $(A_{i,j})_{n \times n} = (a_{i,j})_{n \times n}$, where $s^{(i)}_{[n]}=(s_1,\ldots,s_{i-1},s'_i,s_{i+1},\ldots,s_n)$ is the sequence of data points that differs from $s_{[n]}=(s_1,\ldots,s_n)$ only in the $i$-th entry. If $i = n$, then we have $\widetilde{s}_j = s_j$ for $j \in [n-1]$, which implies that $P_{S_{[j,n]} | S_{[j-1]}=s_{[j-1]}} = P_{S_{[j,n]} | S_{[j-1]}=\widetilde{s}_{[j-1]}}$. One can verify that $\vert \widetilde{\mathcal{E}}_2(w, s_{[n]}, (a_{i,j})_{n \times n}) - \widetilde{\mathcal{E}}_2(w, s^{(i)}_{[n]}, (a_{i,j})_{n \times n}) \vert = 0$ holds. If $i < n$, for any $j \in [i+1, n]$:
\begin{equation}\label{exp_phi4}
\begin{aligned}
    & \Big\vert \mathbb{E}_{ S_{j} | S_{[j-1]}=s_{[j-1]}} \left[ \ell(f_{w,j}((x_{[j-1]},X_j,x_{(j,n]}),(\tilde{a}_{i,j})_{n \times n}), Y_j) \right] \\
    & - \mathbb{E}_{ S_{j} | S_{[j-1]}=\widetilde{s}_{[j-1]}} \left[ \ell(f_{w,j}((\widetilde{s}_{[j-1]},X_j,x_{(j,n]}),(\tilde{a}_{i,j})_{n \times n}), Y_j) \right] \Big\vert \\
    = & \Big\vert \mathbb{E}_{ S_{j} | S_{[i-1]}=s_{[i-1]}, S_{i}=s_i, S_{(i,j-1]}=s_{(i,j-1]}} \left[ \ell(f_{w,j}((x_{[i-1]},x_i,x_{(i,j-1]},X_j,x_{(j,n]}),(\tilde{a}_{i,j})_{n \times n}), Y_j) \right] \\
    & - \mathbb{E}_{ S_{j} | S_{[i-1]}=s_{[i-1]}, S_{i}=s'_i, S_{(i,j-1]}=s_{(i,j-1]}} \left[ \ell(f_{w,j}((x_{[i-1]},x'_i,x_{(i,j-1]},X_j,x_{(j,n]}),(\tilde{a}_{i,j})_{n \times n}), Y_j) \right] \Big\vert \\
    \leq & \Big\vert \mathbb{E}_{ S_{j} | S_{[i-1]}=s_{[i-1]}, S_{i}=s_i, S_{(i,j-1]}=s_{(i,j-1]}} \left[ \ell(f_{w,j}((x_{[i-1]},x_i,x_{(i,j-1]},X_j,x_{(j,n]}),(\tilde{a}_{i,j})_{n \times n}), Y_j) \right] \\
    & - \mathbb{E}_{ S_{j} | S_{[i-1]}=s_{[i-1]}, S_{i}=s'_i, S_{(i,j-1]}=s_{(i,j-1]}} \left[ \ell(f_{w,j}((x_{[i-1]},x_i,x_{(i,j-1]},X_j,x_{(j,n]}),(\tilde{a}_{i,j})_{n \times n}), Y_j) \right] \Big\vert \\
    & + \Big\vert \mathbb{E}_{ S_{j} | S_{[i-1]}=s_{[i-1]}, S_{i}=s'_i, S_{(i,j-1]}=s_{(i,j-1]}} \left[ \ell(f_{w,j}((x_{[i-1]},x_i,x_{(i,j-1]},X_j,x_{(j,n]}),(\tilde{a}_{i,j})_{n \times n}), Y_j) \right] \\
    & - \mathbb{E}_{ S_{j} | S_{[i-1]}=s_{[i-1]}, S_{i}=s'_i, S_{(i,j-1]}=s_{(i,j-1]}} \left[ \ell(f_{w,j}((x_{[i-1]},x'_i,x_{(i,j-1]},X_j,x_{(j,n]}),(\tilde{a}_{i,j})_{n \times n}), Y_j) \right] \Big\vert,
\end{aligned}
\end{equation}
where we have stipulated that $(i,j-1] \triangleq \varnothing$ if $j=i+1$ and $[j+1,j] = \varnothing$ if $j=n$. Following the procedure of deriving Eqs.~(\ref{bound_difference6}, \ref{bound_difference7}), we have
\begin{equation}\label{exp_phi5}
\begin{aligned}
    & \Big\vert \mathbb{E}_{ S_{j} | S_{[i-1]}=s_{[i-1]}, S_{i}=s_i, S_{(i,j-1]}=s_{(i,j-1]}} \left[ \ell(f_w((x_{[i-1]},x_i,x_{(i,j-1]},X_j,x_{(j,n]}), \tilde{a}_{i+1,[n]}), Y_j) \right] \\
    & - \mathbb{E}_{ S_{j} | S_{[i-1]}=s_{[i-1]}, S_{i}=s'_i, S_{(i,j-1]}=s_{(i,j-1]}} \left[ \ell(f_w((x_{[i-1]},x_i,x_{(i,j-1]},X_j,x_{(j,n]}), \tilde{a}_{i+1,[n]}), Y_j) \right] \Big\vert \\
    \leq & M \sup_{s_i, s'_i} {\rm D}_{\rm TV} \left( P_{S_{j} | S_{[i-1]}=s_{[i-1]}, S_{i}=s_i, S_{(i,j-1]}=s_{(i,j-1]}}, P_{S_{j} | S_{[i-1]}=s_{[i-1]}, S_{i}=s'_i, S_{(i,j-1]}=s_{(i,j-1]}} \right) \\
    \leq & M \sup_{s_i, s'_i} {\rm D}_{\rm TV} \left( P_{S_{[j,n]} | S_{[i-1]}=s_{[i-1]}, S_{i}=s_i, S_{(i,j-1]}=s_{(i,j-1]}}, P_{S_{[j,n]} | S_{[i-1]}=s_{[i-1]}, S_{i}=s'_i, S_{(i,j-1]}=s_{(i,j-1]}} \right),
\end{aligned}
\end{equation}
and
\begin{equation}\label{exp_phi6}
\begin{aligned}
    & \Big\vert \mathbb{E}_{ S_{j} | S_{[i-1]}=s_{[i-1]}, S_{i}=s'_i, S_{(i,j-1]}=s_{(i,j-1]}} \left[ \ell(f_w((x_{[i-1]},x_i,x_{(i,j-1]},X_j,x_{(j,n]}), \tilde{a}_{i+1,[n]}), Y_j) \right] \\
    & - \mathbb{E}_{ S_{j} | S_{[i-1]}=s_{[i-1]}, S_{i}=s'_i, S_{(i,j-1]}=s_{(i,j-1]}} \left[ \ell(f_w((x_{[i-1]},x'_i,x_{(i,j-1]},X_j,x_{(j,n]}), \tilde{a}_{i+1,[n]}), Y_j) \right] \Big\vert \\
    \leq & 2 c_x c_w \tilde{a}_{j,i} L.
\end{aligned}
\end{equation}
Plugging Eqs.~(\ref{exp_phi5}, \ref{exp_phi6}) into Eq.~(\ref{exp_phi4}), we obtain that, for any $j \in [i+1, n]$:
\begin{equation}\label{exp_phi7}
\resizebox{0.98\hsize}{!}{$
\begin{aligned}
    & \Big\vert \mathbb{E}_{ S_{j} | S_{[j-1]}=s_{[j-1]}} \left[ \ell(f_w((x_{[j-1]},X_j,x_{(j,n]}), \tilde{a}_{i+1,[n]}), Y_j) \right] - \mathbb{E}_{ S_{j} | S_{[j-1]}=\widetilde{s}_{[j-1]}} \left[ \ell(f_w((\widetilde{s}_{[j-1]},X_j,x_{(j,n]}), \tilde{a}_{i+1,[n]}), Y_j) \right] \Big\vert \\
    \leq & 2 c_x c_w \tilde{a}_{j,i} L + M \widetilde{\Gamma}_{i,j}.
\end{aligned}$}
\end{equation}
Combining Eq.~(\ref{exp_phi4}) and Eq.~(\ref{exp_phi7}), for any fixed realization $W=w$, $(A_{i,j})_{n \times n} = (a_{i,j})_{n \times n}$ and any $i \in [n]$, we have
\begin{equation}\label{exp_phi8}
\begin{aligned}
    \left\vert \widetilde{\mathcal{E}}_2(w, s_{[n]}, (a_{i,j})_{n \times n}) - \widetilde{\Psi}_2(w, s^{(i)}_{[n]}, (a_{i,j})_{n \times n}) \right\vert \leq & \frac{\mathds{1}\{i \ne n \}}{n} \sum_{j=i+1}^n \left( 2 c_x c_w \tilde{a}_{j,i} L + M \widetilde{\Gamma}_{i,j} \right) \\
    \leq & \frac{\mathds{1}\{i \ne n \}}{n} \bigg( 2 c_x c_w c_a L + M \Vert \widetilde{\bm{\Gamma}} \Vert_{\infty} \bigg).
\end{aligned}
\end{equation}
Now we are in a place to derive the final result by applying Eq.~(\ref{final_1}). For each $i \in [n]$, Let $c_i$ be the right hand side of Eq.~(\ref{exp_phi8}), we have
\begin{equation}
\begin{aligned}
    \Vert \bm{c} \Vert^2_2 = & \frac{1}{n^2}\sum_{i=1}^{n-1} \left( 2c_x c_w c_a L + M \Vert \widetilde{\bm{\Gamma}} \Vert_{\infty} \right)^2 \leq \frac{(2 c_x c_w c_a L + M \Vert \widetilde{\bm{\Gamma}} \Vert_{\infty})^2}{n}.
\end{aligned}
\end{equation}
Now using the inequality that $\Vert \bm{\Gamma} \bm{c} \Vert_2 \leq \Vert \bm{\Gamma} \Vert \Vert \bm{c} \Vert_2$ and applying Eq.~(\ref{final_1}), for any fixed realization $W=w$ and $(A_{i,j})_{n \times n} = (a_{i,j})_{n \times n}$:
\begin{equation}
    \mathbb{E}_{S_{[n]}|(A_{i,j})_{n \times n} = (a_{i,j})_{n \times n}} \left[ e^{\lambda \alpha \big(\widetilde{\mathcal{E}}_2(w, S_{[n]}, (a_{i,j})_{n \times n}) - \mathbb{E}_{S'_{[n]}} [ \widetilde{\mathcal{E}}_2(w, s_{[n]}, (a_{i,j})_{n \times n}) ] \big)} \right] \leq \exp \left\{ \frac{\lambda^2 \alpha^2 \Vert \bm{\Gamma} \Vert^2 \Vert \bm{c} \Vert^2}{8} \right\}.
\end{equation}
Taking expectation over $(A_{i,j})_{n \times n}$ on both side yields
\begin{equation}
    \mathbb{E}_{S_{[n]}, (A_{i,j})_{n \times n}} \left[ e^{\lambda \alpha \big(\widetilde{\mathcal{E}}_2(w, S_{[n]}, (a_{i,j})_{n \times n}) - \mathbb{E}_{S'_{[n]}} [ \widetilde{\mathcal{E}}_2(w, s_{[n]}, (a_{i,j})_{n \times n}) ] \big)} \right] \leq \exp \left\{ \frac{\lambda^2 \alpha^2 \Vert \bm{\Gamma} \Vert^2 \Vert \bm{c} \Vert^2}{8} \right\}.
\end{equation}
Following the process of deriving Eq~(\ref{main_pro_bound1}), for any $\delta \in (0,1)$, with probability at least $1-\delta/2$ over the randomness of $S_{[n]}$ and $(A_{i,j})_{n \times n}$:
\begin{equation}\label{ineq}
\begin{aligned}
    & \mathbb{E}_{W \sim Q} \left[ \widetilde{\mathcal{E}}_2(W, S_{[n]}, (A_{i,j})_{n \times n}) \right] \\
    \leq & \mathbb{E}_{W \sim Q} \mathbb{E}_{S'_{[n]}} \left[ \widetilde{\mathcal{E}}_2(W, S'_{[n]}, (A_{i,j})_{n \times n}) \right] + \sqrt{\frac{(2 c_x c_w c_a L + M \Vert \widetilde{\bm{\Gamma}} \Vert_{\infty})^2({\rm D}_{\alpha}(Q||P)+\log(2\sqrt{2n}/\delta) ) \Vert \bm{\Gamma} \Vert^2}{(2n-1)}} .
\end{aligned}
\end{equation}
The last step is providing an upper bound for the expectation term. For any fixed realization $W=w$, $(A_{i,j})_{n \times n}=(a_{i,j})_{n \times n}$ $S'_{[j-1]}=s'_{[j-1]}$ and any $j \in [n]$, we have
\begin{equation}
\begin{aligned}
    & \mathbb{E}_{S'_{[j,n]}|S'_{[j-1]}=s'_{[j-1]}} \Big[ \mathbb{E}_{S'_j |S'_{[j-1]}=s'_{[j-1]}} \left[ \ell(f_{w,j}((x'_{[j]},X'_{(j,n]}), (a_{i,j})_{n \times n}), Y'_j) \right] \\
    & - \mathbb{E}_{ S'_{[j,n]} | S_{[j-1]}=s'_{[j-1]}} \left[ \ell(f_{w,j}((x'_{[j-1]}, X_{[j,n]}), (a_{i,j})_{n \times n}), Y'_j) \right] \Big] \\
    = & \mathbb{E}_{S'_{(j,n]}|S'_{[j-1]}=s'_{[j-1]}} \left[ \mathbb{E}_{S'_j |S'_{[j-1]}=s'_{[j-1]}} \left[ \ell(f_{w,j}((x'_{[j]},X'_{(j,n]}), (a_{i,j})_{n \times n}), Y'_j) \right] \right] \\
    & - \mathbb{E}_{ S'_{[j,n]} | S_{[j-1]}=s'_{[j-1]}} \left[ \ell(f_{w,j}((x'_{[j-1]}, X'_{[j,n]}), (a_{i,j})_{n \times n}), Y'_j) \right] \Big] \\
    = & \mathbb{E}_{S'_{[j,n]} \sim P_{S'_{(j,n]}|S'_{[j-1]}=s'_{[j-1]}} \otimes P_{S_j |S'_{[j-1]}=s'_{[j-1]}}} \left[ \ell(f_{w,j}((x'_{[j]},X'_{(j,n]}), (a_{i,j})_{n \times n}), Y'_j) \right] \\
    & - \mathbb{E}_{ S'_{[j,n]} \sim P_{S'_{[j,n]} | S'_{[j-1]}=s'_{[j-1]}} } \left[ \ell(f_{w,j}((x'_{[j-1]}, X'_{[j,n]}), (a_{i,j})_{n \times n}), Y'_j) \right] \Big] \\
    \leq & M {\rm D}_{\rm TV} \left( P_{S'_{(j,n]}|S'_{[j-1]}=s'_{[j-1]}} \otimes P_{S_j |S'_{[j-1]}=s'_{[j-1]}}, P_{S'_{[j,n]} | S'_{[j-1]}=s'_{[j-1]}} \right),
\end{aligned}
\end{equation}
which implies that
\begin{equation}
\begin{aligned}
    & \mathbb{E}_{S'_{[j,n]}|S'_{[j-1]}} \left[ \mathbb{E}_{S'_j |S'_{[j-1]}} \left[ \ell(f_{w,j}(X'_{[n]}, (A_{i,j})_{n \times n}), Y'_j) \right] - \mathbb{E}_{ S'_{[j,n]} | S_{[j-1]}} \left[ \ell(f_{w,j}(X'_{[n]}, (A_{i,j})_{n \times n}), Y'_j) \right] \right] \\
    \leq & M {\rm D}_{\rm TV} \left( P_{S'_{(j,n]}|S'_{[j-1]}} \otimes P_{S_j |S'_{[j-1]}}, P_{S'_{[j,n]} | S'_{[j-1]}} \right).
\end{aligned}
\end{equation}
Thus, for any fixed realization $W=w$, we get
\begin{equation}\label{exp_phi9}
\begin{aligned}
    & \mathbb{E}_{S'} \Bigg[ \frac{1}{n} \sum_{j=1}^n \left( \mathbb{E}_{S'_j |S'_{[j-1]}} \left[ \ell(f_{w,j}(X'_{[n]}, (A_{i,j})_{n \times n}), Y'_j) \right] - \mathbb{E}_{ S'_{[j,n]} | S_{[j-1]}} \left[ \ell(f_{w,j}(X'_{[n]}, (A_{i,j})_{n \times n}), Y'_j) \right] \right) \Bigg] \\
    = & \frac{1}{n} \sum_{j=1}^n \mathbb{E}_{S'_{[j-1]}} \mathbb{E}_{S'_{[j,n]}|S'_{[j-1]}} \Big[ \mathbb{E}_{S'_j |S'_{[j-1]}} \Big[ \ell(f_{w,j}(X'_{[n]}, (A_{i,j})_{n \times n}), Y'_j) \Big] \\
    & - \mathbb{E}_{ S'_{[j,n]} | S_{[j-1]}} \Big[ \ell(f_{w,j}(X'_{[n]}, (A_{i,j})_{n \times n}), Y'_j) \Big] \Big] \\
    \leq & \frac{M}{n} \sum_{j=1}^n \mathbb{E}_{S'_{[j-1]}} \left[ {\rm D}_{\rm TV} \left( P_{S'_{(j,n]}|S'_{[j-1]}} \otimes P_{S'_j |S'_{[j-1]}}, P_{S'_{[j,n]} | S'_{[j-1]}} \right) \right].
\end{aligned}
\end{equation}
Combining Eq.~(\ref{exp_phi1}), (\ref{exp_phi3}), (\ref{ineq}), and (\ref{exp_phi9}), we get
\begin{equation}
\begin{aligned}
    & \mathbb{E}_{W \sim Q} \left[ \mathcal{E}_2(W, S_{[n]}, (A_{i,j})_{n \times n}) \right] \\
    \leq & M {\rm D}_{\rm TV} \left( P_{A_{n+1,[n+1]} | (A_{i,j})_{n \times n}}, Q_{A_{n+1,[n+1]} | (A_{i,j})_{n \times n}} \right) + \frac{M}{n} \sum_{i=1}^n {\rm D}_{\rm TV} \left( P_{S_{n+1}|S_{[n]}}, P_{S_i |S_{[i-1]}} \right) \\
    & + \frac{M}{n} \sum_{i=1}^n \mathbb{E}_{S_{[i-1]}} \left[ {\rm D}_{\rm TV} \left( P_{S_{(i,n]}|S_{[i-1]}} \otimes P_{S_i |S_{[i-1]}}, P_{S_{[i,n]} | S_{[i-1]}} \right) \right] \\
    & + \sqrt{\frac{(2 c_x c_w c_a L + M \Vert \widetilde{\bm{\Gamma}} \Vert_{\infty})^2({\rm D}_{\alpha}(Q||P)+\log(2\sqrt{2n}/\delta) ) \Vert \bm{\Gamma} \Vert^2}{(2n-1)}}.
\end{aligned}
\end{equation}
This finishes the proof.

\subsection{Proof of Proposition~\ref{pro3}}
For any $i \in [n]$ and $j > i$, by the definition of $\bm{\Gamma}$ in Eq.~(\ref{gamma_matrix}) and definition of total variation distance,
\begin{equation}
\begin{aligned}
    \Gamma_{i,j} = & \mathop{\rm sup}_{x, z \in \Omega, x^{[n]\setminus \{i\}} = z^{[n]\setminus \{i\}}} {\rm D}_{\rm TV} \left( P_{[j,n]} \big( \cdot {\big |} x_{[i]} \big) - P_{[j,n]} \big(\cdot {\big |} z_{[i]} \big) \right) \\
    = & \mathop{\rm sup}_{x_i, z_i} \frac{1}{2} \int_{x_{[j,n]}} \left\vert \dif P_{X_{[j,n]}|X_{[i-1]}=x_{[i-1]},X_i=x_i} \big( x_{[j,n]} \big) - \dif P_{X_{[j,n]}|X_{[i-1]}=x_{[i-1]},X_i=z_i} \big( x_{[j,n]} \big) \right\vert.
\end{aligned}
\end{equation}
By the property of Markov chain,
\begin{equation}\label{gamma_bound}
\begin{aligned}
    & \frac{1}{2} \int_{x_{[j,n]}} \left\vert \dif P_{X_{[j,n]}|X_{[i-1]}=x_{[i-1]},X_i=x_i} \big( x_{[j,n]} \big) - \dif P_{X_{[j,n]}|X_{[i-1]}=x_{[i-1]},X_i=z_i} \big( x_{[j,n]} \big) \right\vert \\
    = & \frac{1}{2} \int_{x_{[j,n]}} \left\vert \dif P_{X_{[j,n]}|X_i=x_i} \big( x_{[j,n]} \big) - \dif P_{X_{[j,n]}|X_i=z_i} \big( x_{[j,n]} \big) \right\vert \\
    = & \frac{1}{2} \int_{x_j} \int_{x_{(j,n]}} \Big\vert \dif P_{X_j|X_i=x_i} \big( x_j \big) \dif P_{X_{(j,n]}|X_j=x_j} \big( x_{(j,n]} \big) - \dif P_{X_j|X_i=z_i} \big( x_j \big) \dif P_{X_{(j,n]}|X_j=x_j} \big( x_{(j,n]} \big) \Big\vert \\
    = & \frac{1}{2} \int_{x_j} \bigg( \int_{x_{(j,n]}} \dif P_{X_{(j,n]}|X_j=x_j} \big( x_{(j,n]} \big) \bigg) \Big\vert \dif P_{X_j|X_i=x_i}(x_j) - \dif P_{X_j|X_i=z_i} \big( x_j \big) \Big\vert \\
    = & \frac{1}{2} \int_{x_j} \Big\vert \dif P_{X_j|X_i=x_i}(x_j) - \dif P_{X_j|X_i=z_i} \big( x_j \big) \Big\vert \\
    \leq & \frac{1}{2} \int_{x_j} \Big\vert \dif P_{X_j|X_i=x_i}(x_j) - \dif \pi(x_j) \Big\vert + \frac{1}{2} \int_{x_j} \Big\vert \dif \pi(x_j) - \dif P_{X_j|X_i=z_i} \big( x_j \big) \Big\vert \\
    = & {\rm D}_{\rm TV} \left( P^{j-i}(x_i, \cdot), \pi \right) + {\rm D}_{\rm TV} \left( P^{j-i}(z_i, \cdot), \pi \right) \leq \rho^{j-i}(M(x_i)+M(z_i)).
\end{aligned}
\end{equation}
Thus, we have $\Gamma_{i,j} \leq 2 \rho^{j-i} \sup_{x} M(x)$. Similarly, one can find that
\begin{equation}\label{gamma_bound2}
\begin{aligned}
    & {\rm D}_{\rm TV} \big( P_{S_{[j,n]} | S_{[i-1]}=s_{[i-1]}, S_{i}=s_i, S_{(i,j-1]}=s_{(i,j-1]}}, P_{S_{[j,n]} | S_{[i-1]}=s_{[i-1]}, S_{i}=s'_i, S_{(i,j-1]}=s_{(i,j-1]}} \big) \\
    = & \begin{cases}
        {\rm D}_{\rm TV} \big( P_{S_{(i,n]} | S_{i}=s_i}, P_{S_{(i,n]} | S_{i}=s'_i} \big), & j = i+1 \\
        {\rm D}_{\rm TV} \big( P_{S_{[j,n]} | S_j=s_j}, P_{S_{[j,n]} | S_j=s_j} \big), & i+1 < j \leq n \\
    \end{cases}.
\end{aligned}
\end{equation}
By the property of Markov chain, ${\rm D}_{\rm TV} \big( P_{S_{[j,n]} | S_j=s_j}, P_{S_{[j,n]} | S_j=s_j} \big)=0$ for any $i+1 < j \leq n$. Thus, we only need to consider the case that $j=i+1$. Combining Eq.~(\ref{gamma_bound}) and Eq.~(\ref{gamma_bound2}), we have
\begin{equation}
\begin{aligned}
    \widetilde{\Gamma}_{i,i+1} = \sup_{s_i, s'_i} {\rm D}_{\rm TV} \big( P_{S_{(i,n]} | S_{i}=s_i}, P_{S_{(i,n]} | S_{i}=s'_i} \big) \leq 2 \rho \sup_{x} M(x).
\end{aligned}
\end{equation}
Thus, we have the following upper bounds
\begin{equation}
\begin{aligned}
    & \Vert \bm{\Gamma} \Vert \leq \Vert \bm{\Gamma} \Vert_{\infty} = \sup_{i \in [n]} \sum_{j=1}^n \Gamma_{i, j} \leq 1 + 2 \sup_{i \in [n]} \frac{1-\rho^{n-i}}{1-\rho} \sup_{x} M(x) = 1 + \frac{2 \sup_{x} M(x)}{1 - \rho}, \\
    & \Vert \widetilde{\bm{\Gamma}} \Vert_{\infty} = \sup_{i \in [n]} \sum_{j=1}^n \widetilde{\Gamma}_{i,j} = \sup_{i \in [n]} \left\{ 2 \rho \sup_{x} M(x) \right\} = 2 \rho \sup_{x} M(x).
\end{aligned}
\end{equation}
This completes the proof.

\subsection{Proof of Proposition~\ref{pro4}}

By definition, for any $i \in [n]$ and fixed realization $S_{[i-1]}=s_{[i-1]}$:
\begin{equation}\label{term2_bound1}
\begin{aligned}
    & {\rm D}_{\rm TV} \left( P_{S_{(i,n]}|S_{[i-1]}=s_{[i-1]}} \otimes P_{S_i |S_{[i-1]}=s_{[i-1]}}, P_{S_{[i,n]} | S_{[i-1]}=s_{[i-1]}} \right) \\
    = & \frac{1}{2} \int_{s_{(i,n]}} \left\vert \dif P_{S_{(i,n]}|S_{[i-1]}=s_{[i-1]}} \big( s_{(i,n]} \big) \dif P_{S_i|S_{[i-1]}=s_{[i-1]}} \big( s_i \big) - \dif P_{S_{[i,n]} | S_{[i-1]}=s_{[i-1]}} \big( s_{[i,n]} \big) \right\vert \\
    = & \frac{1}{2} \int_{s_{(i,n]}} \left\vert \dif P_{S_{(i,n]}|S_{[i-1]}=s_{[i-1]}} \big( s_{(i,n]} \big) \dif P_{S_i|S_{[i-1]}=s_{[i-1]}} \big( s_i \big) - \dif P_{S_{(i,n]} | S_{i}=s_{i}} \big( s_{(i,n]} \big) \dif P_{S_i|S_{[i-1]}=s_{[i-1]}} \big( s_i \big) \right\vert \\
    = & \frac{1}{2} \int_{s_i} \bigg( \int_{s_{(i,n]}} \left\vert \dif P_{S_{(i,n]}|S_{[i-1]}=s_{[i-1]}} \big( s_{(i,n]} \big) - \dif P_{S_{(i,n]} | S_{i}=s_{i}} \big( s_{(i,n]} \big) \right\vert \bigg) \dif P_{S_i|S_{[i-1]}=s_{[i-1]}} \big( s_i \big).
\end{aligned}
\end{equation}
Notice that
\begin{equation}\label{term2_bound2}
\begin{aligned}
    & \int_{s_{(i,n]}} \bigg\vert \dif P_{S_{(i,n]}|S_{[i-1]}=s_{[i-1]}} \big( s_{(i,n]} \big) - \dif P_{S_{(i,n]} | S_{i}=s_{i}} \big( s_{(i,n]} \big) \bigg\vert \\
    = & \int_{s_{(i,n]}} \bigg\vert \bigg( \int_{s'_i} \dif P_{S_i,S_{(i,n]}|S_{[i-1]}=s_{[i-1]}} \big(s'_i, s_{(i,n]} \big) \bigg) - \dif P_{S_{(i,n]} | S_{i}=s_{i}} \big( s_{(i,n]} \big) \bigg\vert \\
    = & \int_{s_{(i,n]}} \bigg\vert \bigg( \int_{s'_i} \dif P_{S_i|S_{[i-1]}=s_{[i-1]}} \big( s'_i \big) \dif P_{S_{(i,n]}|S_i=s'_i}  \big(s_{(i,n]} \big) \bigg) - \dif P_{S_{(i,n]} | S_{i}=s_{i}} \big( s_{(i,n]} \big) \bigg\vert \\
    = & \int_{s_{(i,n]}} \bigg\vert \int_{s'_i} \dif P_{S_i|S_{[i-1]}=s_{[i-1]}} \big( s'_i \big) \bigg(\dif P_{S_{(i,n]}|S_i=s'_i}  \big(s_{(i,n]} \big) - \dif P_{S_{(i,n]} | S_{i}=s_{i}} \big( s_{(i,n]} \big) \bigg) \bigg\vert \\
    \leq & \int_{s_{(i,n]}} \int_{s'_i} \dif P_{S_i|S_{[i-1]}=s_{[i-1]}} \big( s'_i \big) \bigg\vert \dif P_{S_{(i,n]}|S_i=s'_i}  \big(s_{(i,n]} \big) - \dif P_{S_{(i,n]} | S_{i}=s_{i}} \big( s_{(i,n]} \big) \bigg\vert \\
    = & \int_{s_{(i,n]}} \int_{s'_i} \dif P_{S_i|S_{[i-1]}=s_{[i-1]}} \big( s'_i \big) \bigg\vert \dif P_{S_{(i,n]}|S_i=s'_i}  \big(s_{(i,n]} \big) - \dif P_{S_{(i,n]} | S_{i}=s_{i}} \big( s_{(i,n]} \big) \bigg\vert \\
    = & \int_{s'_i} \bigg( \int_{s_{(i,n]}} \bigg\vert \dif P_{S_{(i,n]}|S_i=s'_i}  \big(s_{(i,n]} \big) - \dif P_{S_{(i,n]} | S_{i}=s_{i}} \big( s_{(i,n]} \big) \bigg\vert \bigg) \dif P_{S_i|S_{[i-1]}=s_{[i-1]}} \big( s'_i \big) \\
    = & \int_{s'_i} \bigg( \int_{s_{i+1}} \Big\vert \dif P_{S_{i+1}|S_i=s'_i} \big(s_{i+1} \big) - \dif P_{S_{i+1} | S_{i}=s_{i}} \big( s_{i+1} \big) \Big\vert \bigg) \dif P_{S_i|S_{[i-1]}=s_{[i-1]}} \big( s'_i \big),
\end{aligned}
\end{equation}
where the last line is obtained by Eq.~(\ref{gamma_bound}). Plugging Eq.~(\ref{term2_bound2}) into Eq.~(\ref{term2_bound1}), for any $i \in [2, n]$ we have
\begin{equation}
\begin{aligned}
    & {\rm D}_{\rm TV} \left( P_{S_{(i,n]}|S_{[i-1]}=s_{[i-1]}} \otimes P_{S_i |S_{[i-1]}=s_{[i-1]}}, P_{S_{[i,n]} | S_{[i-1]}=s_{[i-1]}} \right) \\
    \leq & \frac{1}{2} \int_{s_i} \int_{s'_i} \bigg( \int_{s_{i+1}} \Big\vert \dif P_{S_{i+1}|S_i=s'_i} \big(s_{i+1} \big) - \dif P_{S_{i+1} | S_{i}=s_{i}} \big( s_{i+1} \big) \Big\vert \bigg) \dif P_{S_i|S_{[i-1]}=s_{[i-1]}} \big( s'_i \big) \dif P_{S_i|S_{[i-1]}=s_{[i-1]}} \big( s_i \big) \\
    = & \int_{s_i} \int_{s'_i} {\rm D}_{\rm TV} \left( P_{S_{i+1}|S_i=s'_i}, P_{S_{i+1}|S_i=s_i} \right) \dif P_{S_i|S_{i-1}=s_{i-1}} \big( s'_i \big) \dif P_{S_i|S_{i-1}=s_{i-1}} \big( s_i \big) \\
    \leq & \int_{s_i} \int_{s'_i} \left[ {\rm D}_{\rm TV} \left( P(s'_i,\cdot), \pi \right) + {\rm D}_{\rm TV} \left( \pi, P(s_i,\cdot) \right) \right] \dif P_{S_i|S_{i-1}=s_{i-1}} \big( s'_i \big) \dif P_{S_i|S_{i-1}=s_{i-1}} \big( s_i \big) \\
    = & 2 \mathbb{E}_{S \sim P_{S_i|S_{i-1}=s_{i-1}}} \left[ {\rm D}_{\rm TV} \left( P(S, \cdot), \pi \right) \right] \triangleq 2g(s_{i-1}).
\end{aligned}
\end{equation}
Particularly, we stipulate that $g(s_0) = \mathbb{E}_{S \sim P_{S_1}} [ {\rm D}_{\rm TV} ( P(S, \cdot), \pi )]$. By the definition of total variation distance, for any $i \in [3, n]$:
\begin{equation}\label{term2_bound3}
\begin{aligned}
    & \mathbb{E} \left[ {\rm D}_{\rm TV} \left( P_{S_{(i,n]}|S_{[i-1]}=s_{[i-1]}} \otimes P_{S_i |S_{[i-1]}=s_{[i-1]}}, P_{S_{[i,n]} | S_{[i-1]}=s_{[i-1]}} \right) \right] \\
    \leq & 2 \int_{s_{i-1}} g(s_{i-1}) \dif P_{S_{i-1}}(s_{i-1}) = 2 \int_{s_1, s_{i-1}} g(s_{i-1}) \dif P_{S_1,S_{i-1}}(s_1,s_{i-1}) \\
    = & 2 \int_{s_1} \bigg( \int_{s_{i-1}} g(s_{i-1}) \dif P_{S_{i-1}|S_1=s_1}(s_{i-1}) \bigg) \dif P_{S_1}(s_1) \\
    = & 2 \int_{s_1} \bigg( \int_{s_{i-1}} g(s_{i-1}) \dif P_{S_{i-1}|S_1=s_1}(s_{i-1}) - \int_{s_{i-1}} g(s_{i-1}) \dif \pi(s_{i-1}) + \int_{s_{i-1}} g(s_{i-1}) \dif \pi(s_{i-1}) \bigg) \dif P_{S_1}(s_1) \\
    \leq & 2 \int_{s_1} \bigg\vert \int_{s_{i-1}} g(s_{i-1}) \dif P_{S_{i-1}|S_1=s_1}(s_{i-1}) - \int_{s_{i-1}} g(s_{i-1}) \dif \pi(s_{i-1}) \bigg\vert \dif P_{S_1}(s_1) + 2 \int_{s_{i-1}} g(s_{i-1}) \dif \pi(s_{i-1}) \\
    \leq & 2 \int_{s_1} {\rm D}_{\rm TV} \left( P^{i-2}(s_1, \cdot), \pi \right) \dif \pi(s_1) + 2 \mathbb{E}_{S \sim \pi} \left[ g(S) \right] \leq 2\int_{s_1} \rho^{i-2} M(s_1) \dif \pi(s_1) + 2 \mathbb{E}_{S_{i-1} \sim \pi} \left[ g(S_{i-1}) \right] \\
    = & 2\rho^{i-2} \mathbb{E} \left[ M(S_1) \right] + 2 \mathbb{E}_{S_{i-1} \sim \pi} \left[ g(S_{i-1}) \right].
\end{aligned}
\end{equation}
For $i = 2$, we have
\begin{equation}\label{term2_bound4}
\begin{aligned}
    & \mathbb{E} \left[ {\rm D}_{\rm TV} \left( P_{S_{[3,n]}|S_{1}=s_{1}} \otimes P_{S_2 |S_{1}=s_{1}}, P_{S_{[2,n]} | S_{1}=s_{1}} \right) \right] \\
    \leq & 2 \int_{s_1} g(s_1) \dif P_{S_1}(s_1) = 2 \int_{s_1} g(s_1) \dif P_{S_1}(s_1) - 2 \int_{s_1} g(s_1) \dif \pi(s_1) + 2 \int_{s_1} g(s_1) \dif \pi(s_1) \\
    \leq & 2 \bigg\vert \int_{s_1} g(s_1) \dif P_{S_1}(s_1) - \int_{s_1} g(s_1) \dif \pi(s_1) \bigg\vert + 2 \int_{s_1} g(s_1) \dif \pi(s_1) \\
    \leq & 2{\rm D}_{\rm TV} \left( P_{S_1}, \pi \right) + 2 \mathbb{E}_{S_1 \sim \pi} \left[ g(S_1) \right].
\end{aligned}
\end{equation}
For $i=1$, we have
\begin{equation}\label{term2_bound5}
\begin{aligned}
    & {\rm D}_{\rm TV} \left( P_{S_{[2,n]}} \otimes P_{S_1}, P_{S_{[1,n]}} \right) \\
    \leq & \left\vert 2\mathbb{E}_{S \sim P_{S_1}} [ {\rm D}_{\rm TV} ( P(S, \cdot), \pi )] - 2\mathbb{E}_{S \sim \pi} [ {\rm D}_{\rm TV} ( P(S, \cdot), \pi )] \right\vert + 2\mathbb{E}_{S \sim \pi} [ {\rm D}_{\rm TV} ( P(S, \cdot), \pi )]  \\
    \leq & 2{\rm D}_{\rm TV} \left( P_{S_1}, \pi \right) + 2 \mathbb{E}_{S \sim \pi} [{\rm D}_{\rm TV} ( P(S, \cdot), \pi )].
\end{aligned}
\end{equation}
Since $\pi$ is the stationary distribution, for any $i \in [2,n]$ we have
\begin{equation}
    \mathbb{E}_{S_{i-1} \sim \pi} \left[ g(S_{i-1}) \right] = \mathbb{E}_{S_{i-1} \sim \pi} \mathbb{E}_{S_{i} \sim P(S_{i-1}, \cdot)} \left[ {\rm D}_{\rm TV} ( P(S_i, \cdot), \pi ) \right] = \mathbb{E}_{S \sim \pi} [{\rm D}_{\rm TV} ( P(S, \cdot), \pi )].
\end{equation}
Combining Eq.~(\ref{term2_bound3}), Eq.~(\ref{term2_bound4}), and Eq.~(\ref{term2_bound5}) we get the following upper bound for the term in Eq.~(\ref{term2}):
\begin{equation}
\begin{aligned}
    & \mathbb{E}_S \left[ \frac{1}{n}\sum_{i=1}^n {\rm D}_{\rm TV} \left( P_{S_{(i,n]}|S_{[i-1]}} \otimes P_{S_i |S_{[i-1]}}, P_{S_{[i,n]} | S_{[i-1]}} \right) \right] \\
    \leq & \frac{2\mathbb{E} \left[ M(S_1) \right]}{n} \sum_{i=3}^n \rho^{i-2} + \frac{4 {\rm D}_{\rm TV} \left( P_{S_1}, \pi \right)}{n} + 2 \mathbb{E}_{S \sim \pi} \left[{\rm D}_{\rm TV} ( P(S, \cdot), \pi ) \right] \\
    = & \frac{1}{n}\left( \frac{2\rho (1-\rho^{n-2}) \mathbb{E} \left[ M(S_1) \right]}{1-\rho} + 4{\rm D}_{\rm TV} \left( P_{S_1}, \pi \right) \right) + 2 \mathbb{E}_{S \sim \pi} \left[{\rm D}_{\rm TV} ( P(S, \cdot), \pi )\right].
\end{aligned}
\end{equation}
For the term in Eq.~(\ref{term1}), we have
\begin{equation}
\begin{aligned}
    \frac{1}{n}\sum_{i=1}^n {\rm D}_{\rm TV} \left( P_{S_{n+1}|S}, P_{S_i |S_{[i-1]}} \right) \leq & \frac{1}{n}\sum_{i=1}^n \left( {\rm D}_{\rm TV} \left( P_{S_{n+1}|S}, \pi \right) + {\rm D}_{\rm TV} \left( P_{S_i |S_{[i-1]}}, \pi \right) \right) \\
    = & {\rm D}_{\rm TV} \left( P_{S_{n+1}|S_n}, \pi \right) + \frac{1}{n}\sum_{i=1}^n {\rm D}_{\rm TV} \left( P_{S_i |S_{i-1}}, \pi \right).
\end{aligned}
\end{equation}
This completes the proof.

\subsection{Proof of Theorem~\ref{two_layer_bound}}

The proof process generally follows that of Theorem~\ref{one_layer_bound}. The first step is decomposing the generalization gap $\mathcal{E}(W,S_{[n]},(A_{i,j})_{n \times n})$ into the following two terms
\begin{equation}
\begin{aligned}
    \mathcal{E}_1(W,S_{[n]},(A_{i,j})_{n \times n}) = & \frac{1}{n} \sum_{\imath = 1}^n \mathbb{E}_{S_{[\imath,n]}} \big[ \ell(f_{W,\imath}(X_{[n]}, (\hat{A}_{i,j})_{n \times n}), Y_{\imath}) \big\vert S_{[\imath-1]} \big] - \frac{1}{n} \sum_{\imath = 1}^n \ell(f_{W,\imath}(X_{[n]}, (\hat{A}_{i,j})_{n \times n}), Y_{\imath}), \\
    \mathcal{E}_2(W,S_{[n]},(A_{i,j})_{n \times n}) = & \mathbb{E}_{S_{n+1}, A_{n+1,[n+1]}} \big[ \ell(f_W(X_{[n+1]}, (\hat{A}_{i,j})_{n \times n}, A_{n+1,[n+1]}),Y_{n+1}) \big\vert S_{[n]},(A_{i,j})_{n \times n} \big] \\
    & - \frac{1}{n} \sum_{\imath = 1}^n \mathbb{E}_{S_{[\imath,n]}} \big[ \ell(f_{W,\imath}(X_{[n]}, (\hat{A}_{i,j})_{n \times n}), Y_{\imath}) \big\vert S_{[\imath-1]} \big].
\end{aligned}
\end{equation}
Denote by $s^{(i)}_{[n]}=(s_1,\ldots,s_{i-1},s'_i,s_{i+1},\ldots,s_n)$ the sequence of data points that differs from $s_{[n]}=(s_1,\ldots,s_n)$ only in the $i$-th entry. We need to derive an upper bound of $\vert \mathcal{E}_1(w,s_{[n]}, (a_{i,j})_{n \times n}) - \mathcal{E}_1(w,s^{(i)}_n, (a_{i,j})_{n \times n}) \vert$ for any fixed realization $W=w$ and $(A_{i,j})_{n \times n}=(a_{i,j})_{n \times n}$. By definition,
\begin{equation}\label{bound_difference_21}
\begin{aligned}
    & \left\vert \mathcal{E}_1(w,s_{[n]},(a_{i,j})_{n \times n}) - \mathcal{E}_1(w,s^{(i)}_{[n]},(a_{i,j})_{n \times n}) \right\vert \\
    \leq & \Bigg\vert \frac{1}{n} \sum_{\imath=1}^n \left( \ell(f_{w,\imath}(x_{[n]}, (\hat{a}_{i,j})_{n \times n}), y_{\imath}) - \ell(f_w(\widetilde{x}_{[n]}, (\hat{a}_{i,j})_{n \times n}), \widetilde{y}_{\imath}) \right) \Bigg\vert \\
    & + \Bigg \vert \frac{1}{n} \sum_{\imath=1}^n \Big( \mathbb{E}_{ S_{[\imath,n]}} \big[ \ell(f_{w,\imath}((x_{[\imath-1]},X_{[\imath,n]}), (\hat{a}_{i,j})_{n \times n}), Y_{\imath}) \big\vert S_{[\imath-1]}=s_{[\imath-1]} \big] \\
    & - \mathbb{E}_{ S_{[\imath,n]}} \big[ \ell(f_{w,\imath}((\widetilde{x}_{[\imath-1]},X_{[\imath,n]}), (\hat{a}_{i,j})_{n \times n}), Y_{\imath}) \big\vert S_{[\imath-1]}=\widetilde{s}_{[\imath-1]} \big] \Big) \Bigg\vert.
\end{aligned}
\end{equation}
For any $\imath \in [n]$ such that $\imath \ne i$, we have $\widetilde{x}_{\imath} = x_{\imath}$ and $\widetilde{y}_{\imath} = y_{\imath}$. If $v_i$ is the first-order or second-order neighbor of $v_{\imath}$, then we have
\begin{equation}\label{bound_first}
\begin{aligned}
    & \left\vert \ell(f_{w,\imath}(x_{[n]}, (\hat{a}_{i,j})_{n \times n}), y_{j}) - \ell(f_{w,\imath}(\widetilde{x}_{[n]}, (\hat{a}_{i,j})_{n \times n}), \widetilde{y}_{j}) \right\vert \\
    = & \left\vert \ell(f_{w,\imath}(x_{[n]}, (\hat{a}_{i,j})_{n \times n}), y_{j}) - \ell(f_{w,\imath}(\widetilde{x}_{[n]}, (\hat{a}_{i,j})_{n \times n}), y_{j}) \right\vert \\
    \leq & L \Bigg\Vert \phi \Bigg( \sum_{j=1}^n \hat{a}_{\imath, j} \Bigg( \phi \left( \sum_{k=1}^n \hat{a}_{j,k} x_k w_1 \right) w_2 \Bigg) - \phi \Bigg( \sum_{j=1}^n \hat{a}_{\imath, j} \phi \left( \sum_{k=1}^n \hat{a}_{j,k} \widetilde{x}_k w_1 \right) \Bigg) w_2 \Bigg\Vert \\
    \leq & L L_{\phi} \Bigg\Vert \sum_{j=1}^n \hat{a}_{\imath, j} \Bigg( \phi \left( \sum_{k=1}^n \hat{a}_{j,k} x_k w_1 \right) - \phi \left( \sum_{k=1}^n \hat{a}_{j,k} \widetilde{x}_k w_1 \right) \Bigg) w_2 \Bigg\Vert \\
    \leq & c_w L L^2_{\phi} \sum_{j=1}^n \hat{a}_{\imath, j} \Bigg\Vert \sum_{k=1}^n \hat{a}_{j,k} \big( x_k - \widetilde{x}_k \big) w_1 \Bigg\Vert \\
    = & c_w L L^2_{\phi} \sum_{j=1}^n \hat{a}_{\imath, j} \left\Vert \hat{a}_{j,i} ( x_i - x'_i ) w_1 \right\Vert \leq c^2_w L L^2_{\phi} \sum_{j=1}^n \hat{a}_{\imath, j} \hat{a}_{j,i} \Vert x_i - x'_i \Vert \leq 2 c_x c^2_w L L^2_{\phi} \sum_{j=1}^n \hat{a}_{\imath, j} \hat{a}_{j,i}.
\end{aligned}
\end{equation}
Otherwise, if $v_i$ is neither the first-order nor the second-order neighbor of $v_{\imath}$, then we have
\begin{equation}
    f_{w,\imath}(x_{[n]}, (\hat{a}_{i,j})_{n \times n}) = f_{w,\imath}(\widetilde{x}_{[n]}, (\hat{a}_{i,j})_{n \times n})
\end{equation}
and thus
\begin{equation}
\begin{aligned}
    & \left\vert \ell(f_{w,\imath}(x_{[n]}, (\hat{a}_{i,j})_{n \times n}), y_{\imath}) - \ell(f_{w,\imath}(\widetilde{x}_{[n]}, (\hat{a}_{i,j})_{n \times n}), \widetilde{y}_{\imath}) \right\vert \\
    = & \left\vert \ell(f_{w,\imath}(x_{[n]}, (\hat{a}_{i,j})_{n \times n}), y_{\imath}) - \ell(f_{w,\imath}(\widetilde{x}_{[n]}, (\hat{a}_{i,j})_{n \times n}), y_{\imath}) \right\vert = 0.
\end{aligned}
\end{equation}
For the case that $\imath=i$, by the assumption that $\ell(\cdot, \cdot)$ is bounded by $M$, we have
\begin{equation}
\begin{aligned}
    \left\vert \ell(f_{w,\imath}(x_{[n]}, (\hat{a}_{i,j})_{n \times n}), y_{\imath}) - \ell(f_{w,\imath}(\widetilde{x}_{[n]}, (\hat{a}_{i,j})_{n \times n}), \widetilde{y}_{\imath}) \right\vert \leq M.
\end{aligned}
\end{equation}
Combining the above inequalities, we get
\begin{equation}
\begin{aligned}
    & \Bigg\vert \frac{1}{n} \sum_{\imath=1}^n \left( \ell(f_{w,\imath}(x_{[n]}, (\hat{a}_{i,j})_{n \times n}), y_{\imath}) - \ell(f_w(\widetilde{x}_{[n]}, (\hat{a}_{i,j})_{n \times n}), \widetilde{y}_{\imath}) \right) \Bigg\vert \\
    \leq & \frac{1}{n} \sum_{\imath=1}^n \left\vert \ell(f_w(X_{[n]}, \hat{a}_{j,[n]}), Y_j) - \frac{1}{n} \sum_{\imath=1}^n \ell(f_w(\widetilde{X}_{[n]}, \hat{a}_{j,[n]}), \widetilde{Y}_j) \right\vert \\
    = & \frac{1}{n} \sum_{\imath \ne i} \left\vert \ell(f_w(X_{[n]}, \hat{a}_{j,[n]}), Y_j) - \frac{1}{n} \sum_{\imath=1}^n \ell(f_w(\widetilde{X}_{[n]}, \hat{a}_{j,[n]}), \widetilde{Y}_j) \right\vert \\
    & + \frac{1}{n} \left\vert \ell(f_w(X_{[n]}, (\hat{a}_{i,[n]})), Y_i) - \ell(f_w(\widetilde{X}_{[n]}, \hat{a}_{j,[n]}), \widetilde{Y}_i) \right\vert \\
    \leq & \frac{2 c_x c^2_w L L^2_{\phi}}{n} \sum_{\imath \ne i} \sum_{j=1}^n \hat{a}_{\imath, j} \hat{a}_{j,i} + \frac{M}{n} \leq \frac{2 c_x c^2_w c^2_a L L^2_{\phi} + M}{n},
\end{aligned}
\end{equation}
where the last inequality is obtained by the fact that
\begin{equation}
    \sum_{\imath \ne i} \sum_{j=1}^n \hat{a}_{\imath, j} \hat{a}_{j,i} \leq \sum_{\imath=1}^n \sum_{j=1}^n \hat{a}_{\imath, j} \hat{a}_{j,i} \leq \left\Vert ((\hat{a}_{ij})_{n \times n})^2 \right\Vert_1 \leq \left\Vert (\hat{a}_{ij})_{n \times n} \right\Vert^2_1 = \left\Vert (\hat{a}_{ij})_{n \times n} \right\Vert^2_{\infty} \leq c^2_a.
\end{equation}
Now we analyze the term
\begin{equation}
\begin{aligned}
    & \Bigg \vert \frac{1}{n} \sum_{\imath=1}^n \Big( \mathbb{E}_{ S_{[\imath,n]}} \big[ \ell(f_{w,\imath}((x_{[\imath-1]},X_{[\imath,n]}), (\hat{a}_{i,j})_{n \times n}), Y_{\imath}) \big\vert S_{[\imath-1]}=s_{[\imath-1]} \big] \\
    & - \mathbb{E}_{ S_{[\imath,n]}} \big[ \ell(f_{w,\imath}((\widetilde{x}_{[\imath-1]},X_{[\imath,n]}), (\hat{a}_{i,j})_{n \times n}), Y_{\imath}) \big\vert S_{[\imath-1]}=\widetilde{s}_{[\imath-1]} \big] \Big) \Bigg\vert.
\end{aligned}
\end{equation}
If $i = n$, then we have $\widetilde{s}_j = s_j$ for $j \in [n-1]$, which implies that $P_{S_{[j,n]} | S_{[j-1]}=s_{[j-1]}} = P_{S_{[j,n]} | S_{[j-1]}=\widetilde{s}_{[j-1]}}$ for each $j \in [n-1]$. Then we have
\begin{equation}
\begin{aligned}
    & \Bigg \vert \frac{1}{n} \sum_{\imath=1}^n \Big( \mathbb{E}_{ S_{[\imath,n]}} \big[ \ell(f_{w,\imath}((x_{[\imath-1]},X_{[\imath,n]}), (\hat{a}_{i,j})_{n \times n}), Y_{\imath}) \big\vert S_{[\imath-1]}=s_{[\imath-1]} \big] \\
    & - \mathbb{E}_{ S_{[\imath,n]}} \big[ \ell(f_{w,\imath}((\widetilde{x}_{[\imath-1]},X_{[\imath,n]}), (\hat{a}_{i,j})_{n \times n}), Y_{\imath}) \big\vert S_{[\imath-1]}=\widetilde{s}_{[\imath-1]} \big] \Big) \Bigg\vert \\
    = & \Bigg \vert \frac{1}{n} \sum_{\imath=1}^n \Big( \mathbb{E}_{ S_{[\imath,n]}} \big[ \ell(f_{w,\imath}((x_{[\imath-1]},X_{[\imath,n]}), (\hat{a}_{i,j})_{n \times n}), Y_{\imath}) \big\vert S_{[\imath-1]}=s_{[\imath-1]} \big] \\
    & - \mathbb{E}_{ S_{[\imath,n]}} \big[ \ell(f_{w,\imath}((x_{[\imath-1]},X_{[\imath,n]}), (\hat{a}_{i,j})_{n \times n}), Y_{\imath}) \big\vert S_{[\imath-1]}=s_{[\imath-1]} \big] \Big) \Bigg\vert = 0.
\end{aligned}
\end{equation}
If $i \ne n$, then we have $\widetilde{s}_{\imath} = s_{\imath}$ for $\imath \in [i-1]$, which implies that $P_{S_{[\imath,n]} | S_{[\imath-1]}=s_{[\imath-1]}} = P_{S_{[\imath,n]} | S_{[\imath-1]}=\widetilde{s}_{[\imath-1]}}$ for each $\imath \in [i-1]$. Now we further decompose this term into the following two terms
\begin{equation}
\begin{aligned}
    & \Bigg \vert \frac{1}{n} \sum_{\imath=1}^n \Big( \mathbb{E}_{ S_{[\imath,n]}} \big[ \ell(f_{w,\imath}((x_{[\imath-1]},X_{[\imath,n]}), (\hat{a}_{i,j})_{n \times n}), Y_{\imath}) \big\vert S_{[\imath-1]}=s_{[\imath-1]} \big] \\
    & - \mathbb{E}_{ S_{[\imath,n]}} \big[ \ell(f_{w,\imath}((\widetilde{x}_{[\imath-1]},X_{[\imath,n]}), (\hat{a}_{i,j})_{n \times n}), Y_{\imath}) \big\vert S_{[\imath-1]}=\widetilde{s}_{[\imath-1]} \big] \Big) \Bigg\vert \\
    \leq & \Bigg \vert \frac{1}{n} \sum_{\imath=1}^{i} \Big( \mathbb{E}_{ S_{[\imath,n]}} \big[ \ell(f_{w,\imath}((x_{[\imath-1]},X_{[\imath,n]}), (\hat{a}_{i,j})_{n \times n}), Y_{\imath}) \big\vert S_{[\imath-1]}=s_{[\imath-1]} \big] \\
    & - \mathbb{E}_{ S_{[\imath,n]}} \big[ \ell(f_{w,\imath}((\widetilde{x}_{[\imath-1]},X_{[\imath,n]}), (\hat{a}_{i,j})_{n \times n}), Y_{\imath}) \big\vert S_{[\imath-1]}=\widetilde{s}_{[\imath-1]} \big] \Big) \Bigg\vert \\
    & + \Bigg \vert \frac{1}{n} \sum_{\imath=i+1}^{n} \Big( \mathbb{E}_{ S_{[\imath,n]}} \big[ \ell(f_{w,\imath}((x_{[\imath-1]},X_{[\imath,n]}), (\hat{a}_{i,j})_{n \times n}), Y_{\imath}) \big\vert S_{[\imath-1]}=s_{[\imath-1]} \big] \\
    & - \mathbb{E}_{ S_{[\imath,n]}} \big[ \ell(f_{w,\imath}((\widetilde{x}_{[\imath-1]},X_{[\imath,n]}), (\hat{a}_{i,j})_{n \times n}), Y_{\imath}) \big\vert S_{[\imath-1]}=\widetilde{s}_{[\imath-1]} \big] \Big) \Bigg\vert.
\end{aligned}
\end{equation}
It can be verified that the first term equals zero, thus it is sufficient to analyze the second term. For any $\imath \in [i+1, n]$, we have
\begin{equation}
\begin{aligned}
    & \Big\vert \mathbb{E}_{ S_{[\imath,n]}} \big[ \ell(f_{w,\imath}((x_{[\imath-1]},X_{[\imath,n]}), (\hat{a}_{i,j})_{n \times n}), Y_{\imath}) \big\vert S_{[\imath-1]}=s_{[\imath-1]} \big] \\
    & - \mathbb{E}_{ S_{[\imath,n]}} \big[ \ell(f_{w,\imath}((\widetilde{x}_{[\imath-1]},X_{[\imath,n]}), (\hat{a}_{i,j})_{n \times n}), Y_\imath) \big\vert S_{[\imath-1]}=\widetilde{s}_{[\imath-1]} \big] \\
    = & \Big\vert \mathbb{E}_{ S_{[\imath,n]}} \big[ \ell(f_{w,\imath}((x_{[i-1]},x_i,x_{(i,\imath-1]},X_{[\imath,n]}), (\hat{a}_{i,j})_{n \times n}), Y_{\imath}) \big\vert S_{[i-1]}=s_{[i-1]}, S_{i}=s_i, S_{(i,\imath-1]}=s_{(i,\imath-1]} \big] \\
    & - \mathbb{E}_{ S_{[\imath,n]}} \big[ \ell(f_{w,\imath}((x_{[i-1]},x'_i,x_{(i,\imath-1]},X_{[\imath,n]}), (\hat{a}_{i,j})_{n \times n}), Y_\imath) \big\vert S_{[i-1]}=s_{[i-1]}, S_{i}=s'_i, S_{(i,\imath-1]}=s_{(i,\imath-1]} \big] \\
    \leq & \Big\vert \mathbb{E}_{ S_{[\imath,n]}} \big[ \ell(f_{w,\imath}((x_{[i-1]},x_i,x_{(i,\imath-1]},X_{[\imath,n]}), (\hat{a}_{i,j})_{n \times n}), Y_{\imath}) \big\vert S_{[i-1]}=s_{[i-1]}, S_{i}=s_i, S_{(i,\imath-1]}=s_{(i,\imath-1]} \big] \\
    & - \mathbb{E}_{ S_{[\imath,n]}} \big[ \ell(f_{w,\imath}((x_{[i-1]},x_i,x_{(i,\imath-1]},X_{[\imath,n]}), (\hat{a}_{i,j})_{n \times n}), Y_{\imath}) \big\vert S_{[i-1]}=s_{[i-1]}, S_{i}=s'_i, S_{(i,\imath-1]}=s_{(i,\imath-1]} \big] \Big\vert \\
    & + \Big\vert \mathbb{E}_{ S_{[\imath,n]}} \big[ \ell(f_{w,\imath}((x_{[i-1]},x_i,x_{(i,\imath-1]},X_{[\imath,n]}), (\hat{a}_{i,j})_{n \times n}), Y_{\imath}) \big\vert S_{[i-1]}=s_{[i-1]}, S_{i}=s'_i, S_{(i,\imath-1]}=s_{(i,\imath-1]} \big] \\
    & - \mathbb{E}_{ S_{[\imath,n]}} \big[ \ell(f_{w,\imath}((x_{[i-1]},x'_i,x_{(i,\imath-1]},X_{[\imath,n]}), (\hat{a}_{i,j})_{n \times n}), Y_\imath) \big\vert S_{[i-1]}=s_{[i-1]}, S_{i}=s'_i, S_{(i,\imath-1]}=s_{(i,\imath-1]} \big] \Big\vert,
\end{aligned}
\end{equation}
where we have stipulated that $(i,\imath-1] \triangleq \varnothing$ if $\imath=i+1$. By the definition of total variation distance, for any $\imath \in (i,n]$ we have
\begin{equation}\label{pre1}
\begin{aligned}
    & \Big\vert \mathbb{E}_{ S_{[\imath,n]}} \big[ \ell(f_{w,\imath}((x_{[i-1]},x_i,x_{(i,\imath-1]},X_{[\imath,n]}), (\hat{a}_{i,j})_{n \times n}), Y_{\imath}) \big\vert S_{[i-1]}=s_{[i-1]}, S_{i}=s_i, S_{(i,\imath-1]}=s_{(i,\imath-1]} \big] \\
    & - \mathbb{E}_{ S_{[\imath,n]}} \big[ \ell(f_{w,\imath}((x_{[i-1]},x_i,x_{(i,\imath-1]},X_{[\imath,n]}), (\hat{a}_{i,j})_{n \times n}), Y_{\imath}) \big\vert S_{[i-1]}=s_{[i-1]}, S_{i}=s'_i, S_{(i,\imath-1]}=s_{(i,\imath-1]} \big] \Big\vert \\
    = & \bigg\vert \int_{s_{[\imath,n]}} \ell(f_{w,\imath}((x_{[\imath-1]},x_{[\imath,n]}), (\hat{a}_{i,j})_{n \times n}), y_{\imath}) \dif P_{S_{[\imath,n]} | S_{[i-1]}=s_{[i-1]}, S_{i}=s_i, S_{(i,\imath-1]}=s_{(i,\imath-1]}}(s_{[\imath,n]}) \\
    & - \int_{s_{[\imath,n]}} \ell(f_{w,\imath}((x_{[\imath-1]},x_{[\imath,n]}), (\hat{a}_{i,j})_{n \times n}), y_{\imath}) \dif P_{S_{[\imath,n]} | S_{[i-1]}=s_{[i-1]}, S_{i}=s'_i, S_{(i,\imath-1]}=s_{(i,\imath-1]}}(s_{[\imath,n]}) \bigg\vert \\
    \leq & M {\rm D}_{\rm TV} \left( P_{S_{[\imath,n]} | S_{[i-1]}=s_{[i-1]}, S_{i}=s_i, S_{(i,\imath-1]}=s_{(i,\imath-1]}}, P_{S_{[\imath,n]} | S_{[i-1]}=s_{[i-1]}, S_{i}=s'_i, S_{(i,\imath-1]}=s_{(i,\imath-1]}} \right) \\
    \leq & M \sup_{s_i, s'_i} {\rm D}_{\rm TV} \left( P_{S_{[\imath,n]} | S_{[i-1]}=s_{[i-1]}, S_{i}=s_i, S_{(i,\imath-1]}=s_{(i,\imath-1]}}, P_{S_{[\imath,n]} | S_{[i-1]}=s_{[i-1]}, S_{i}=s'_i, S_{(i,\imath-1]}=s_{(i,\imath-1]}} \right).
\end{aligned}
\end{equation}
For any $\imath \in (i,n]$ and fixed realization $S_{[\imath,n]}=s_{[\imath,n]}$, following the same procedure of deriving Eq.~(\ref{bound_first}), we have
\begin{equation}
\begin{aligned}
    & \left\vert \ell(f_{w,\imath}((x_{[i-1]},x_i,x_{(i,\imath-1]},x_{[\imath,n]}), (\hat{a}_{i,j})_{n \times n})), y_\imath) - \ell(f_{w,\imath}((x_{[i-1]},x'_i,x_{(i,\imath-1]},x_{[\imath,n]}), (\hat{a}_{i,j})_{n \times n})), y_\imath) \right\vert \\
    \leq & 2 c_x c^2_w L L^2_{\phi} \sum_{j=1}^n \hat{a}_{\imath,j} \hat{a}_{j,i},
\end{aligned}
\end{equation}
which implies that
\begin{equation}\label{pre2}
\begin{aligned}
    & \Big\vert \mathbb{E}_{ S_{[\imath,n]}} \big[ \ell(f_{w,\imath}((x_{[i-1]},x_i,x_{(i,\imath-1]},X_{[\imath,n]}), (\hat{a}_{i,j})_{n \times n}), Y_{\imath}) \big\vert S_{[i-1]}=s_{[i-1]}, S_{i}=s'_i, S_{(i,\imath-1]}=s_{(i,\imath-1]} \big] \\
    & - \mathbb{E}_{ S_{[\imath,n]}} \big[ \ell(f_{w,\imath}((x_{[i-1]},x'_i,x_{(i,\imath-1]},X_{[\imath,n]}), (\hat{a}_{i,j})_{n \times n}), Y_\imath) \big\vert S_{[i-1]}=s_{[i-1]}, S_{i}=s'_i, S_{(i,\imath-1]}=s_{(i,\imath-1]} \big] \Big\vert \\
    \leq & \int_{s_{[\imath,n]}} \big\vert \ell(f_{w,\imath}((x_{[i-1]},x_i,x_{(i,\imath-1]},x_{[\imath,n]}), (\hat{a}_{i,j})_{n \times n}), y_{\imath}) \\
    & - \ell(f_{w,\imath}((x_{[i-1]},x'_i,x_{(i,\imath-1]},x_{[\imath,n]}), (\hat{a}_{i,j})_{n \times n}), y_{\imath}) \big\vert \dif P_{S_{[\imath,n]} | S_{[i-1]}=s_{[i-1]}, S_{i}=s'_i, S_{(i,\imath-1]}=s_{(i,\imath-1]}}(s_{[\imath,n]}) \\
    \leq & 2 c_x c^2_w L L^2_{\phi} \sum_{j=1}^n \hat{a}_{\imath,j} \hat{a}_{j,i}.
\end{aligned}
\end{equation}
Combining the above ingredients, for any $\imath \in [i+1, n]$:
\begin{equation}\label{pre0}
\begin{aligned}
    & \Big\vert \mathbb{E}_{ S_{[\imath,n]}} \big[ \ell(f_{w,\imath}((x_{[\imath-1]},X_{[\imath,n]}), (\hat{a}_{i,j})_{n \times n}), Y_{\imath}) \big\vert S_{[\imath-1]}=s_{[\imath-1]} \big] \\
    & - \mathbb{E}_{ S_{[\imath,n]}} \big[ \ell(f_{w,\imath}((\widetilde{x}_{[\imath-1]},X_{[\imath,n]}), (\hat{a}_{i,j})_{n \times n}), Y_\imath) \big\vert S_{[\imath-1]}=\widetilde{s}_{[\imath-1]} \big] \leq 2 c_x c^2_w L L^2_{\phi} \sum_{j=1}^n \hat{a}_{\imath,j} \hat{a}_{j,i} + M \widetilde{\Gamma}_{i,\imath}.
\end{aligned}
\end{equation}
Then we get
\begin{equation}
\begin{aligned}
    & \Bigg \vert \frac{1}{n} \sum_{\imath=1}^n \Big( \mathbb{E}_{ S_{[\imath,n]}} \big[ \ell(f_{w,\imath}((x_{[\imath-1]},X_{[\imath,n]}), (\hat{a}_{i,j})_{n \times n}), Y_{\imath}) \big\vert S_{[\imath-1]}=s_{[\imath-1]} \big] \\
    & - \mathbb{E}_{ S_{[\imath,n]}} \big[ \ell(f_{w,\imath}((\widetilde{x}_{[\imath-1]},X_{[\imath,n]}), (\hat{a}_{i,j})_{n \times n}), Y_{\imath}) \big\vert S_{[\imath-1]}=\widetilde{s}_{[\imath-1]} \big] \Big) \Bigg\vert \\
    \leq & \frac{\mathds{1}\{i \neq n\}}{n} \sum_{\imath = i+1}^n \bigg( 2 c_x c^2_w L L^2_{\phi} \sum_{j=1}^n \hat{a}_{\imath,j} \hat{a}_{j,i} + M \widetilde{\Gamma}_{i,\imath} \bigg) \\
    = & \frac{\mathds{1}\{i \neq n\}}{n} \bigg( 2 c_x c^2_w L L^2_{\phi} \sum_{\imath = i+1}^n \sum_{j=1}^n \hat{a}_{\imath,j} \hat{a}_{j,i} + M \sum_{\imath = 1}^n \widetilde{\Gamma}_{i,\imath} \bigg) \\
    \leq & \frac{\mathds{1}\{i \neq n\}}{n} \bigg( 2 c_x c^2_w L L^2_{\phi} \sum_{\imath = 1}^n \sum_{j=1}^n \hat{a}_{\imath,j} \hat{a}_{j,i} + M \sum_{\imath = 1}^n \widetilde{\Gamma}_{i,\imath} \bigg) \leq \frac{\mathds{1}\{i \neq n\}}{n} \left(2 c_x c^2_w c^2_a L L^2_{\phi} + M \Vert \widetilde{\bm{\Gamma}} \Vert_{\infty} \right).
\end{aligned}
\end{equation}
Till now, we have proved that
\begin{equation}\label{bd1}
\begin{aligned}
    \left\vert \Psi_1(w, s) - \Psi_1(w, s^{(i)}) \right\vert \leq & \frac{2 c_x c^2_w c^2_a L L^2_{\phi} + M}{n} + \frac{\mathds{1}\{i \neq n\}}{n} \left(2 c_x c^2_w c^2_a L + M \Vert \widetilde{\bm{\Gamma}} \Vert_{\infty} \right) \\
    \leq & \frac{2(2 c_x c^2_w c^2_a L L^2_{\phi} + M \max (1, \Vert \widetilde{\bm{\Gamma}} \Vert_{\infty}))}{n}.
\end{aligned}
\end{equation}
Now we are in a place to derive the final result by applying Eq.~(\ref{final_1}). For each $i \in [n]$, Let $c_i$ be the right hand side of Eq.~(\ref{bd1}), we have
\begin{equation}
\begin{aligned}
    \Vert \bm{c} \Vert^2_2 \leq & \frac{4}{n^2} \sum_{\imath = 1}^{n} \left( 2 c_x c^2_w c^2_a L L^2_{\phi} + M \max (1, \Vert \widetilde{\bm{\Gamma}} \Vert_{\infty}) \right)^2 = \frac{4(2 c_x c^2_w c^2_a L L^2_{\phi} + M \max (1, \Vert \widetilde{\bm{\Gamma}} \Vert_{\infty}))^2}{n}.
\end{aligned}
\end{equation}
Now using the inequality that $\Vert \bm{\Gamma} \bm{c} \Vert_2 \leq \Vert \bm{\Gamma} \Vert \Vert \bm{c} \Vert_2$ and applying Eq.~(\ref{final_1}), for any fixed realization $W=w$ and $(A_{i,j})_{n \times n} = (a_{i,j})_{n \times n}$:
\begin{equation}
    \mathbb{E}_{S_{[n]}|(A_{i,j})_{n \times n} = (a_{i,j})_{n \times n}} \left[ e^{\lambda \alpha \big(\mathcal{E}_1(w, S_{[n]}, (a_{i,j})_{n \times n}) - \mathbb{E}_{S'_{[n]}} [ \mathcal{E}_1(w, s_{[n]}, (a_{i,j})_{n \times n}) ] \big)} \right] \leq \exp \left\{ \frac{\lambda^2 \alpha^2 \Vert \bm{\Gamma} \Vert^2 \Vert \bm{c} \Vert^2}{8} \right\}.
\end{equation}
Taking expectation over $(A_{i,j})_{n \times n}$ on both side yields
\begin{equation}
    \mathbb{E}_{S_{[n]}, (A_{i,j})_{n \times n}} \left[ e^{\lambda \alpha \big(\mathcal{E}_1(w, S_{[n]}, (a_{i,j})_{n \times n}) - \mathbb{E}_{S'_{[n]}} [ \mathcal{E}_1(w, s_{[n]}, (a_{i,j})_{n \times n}) ] \big)} \right] \leq \exp \left\{ \frac{\lambda^2 \alpha^2 \Vert \bm{\Gamma} \Vert^2 \Vert \bm{c} \Vert^2}{8} \right\}.
\end{equation}
Notice that $\mathbb{E}_{W \sim Q} \mathbb{E}_{S'_{[n]}}[\mathcal{E}_1(W, S'_{[n]}, (a_{i,j})_{n \times n})] = 0$ for any fixed realization $(A_{i,j})_{n \times n}=(a_{i,j})_{n \times n}$. Following the process of deriving Eq~(\ref{main_pro_bound1}), for any $\delta \in (0,1)$, with probability at least $1-\delta/2$ over the randomness of $S_{[n]}$ and $(A_{i,j})_{n \times n}$:
\begin{equation}\label{b0}
    \mathbb{E}_{W \sim Q} \left[ \mathcal{E}_1(W,S_{[n]},(A_{i,j})_{n \times n}) \right] \leq 2\sqrt{\frac{(2 c_x c^2_w c^2_a L L^2_{\phi} + M \max (1, \Vert \widetilde{\bm{\Gamma}} \Vert_{\infty}))^2({\rm D}_{\alpha}(Q||P)+\log(2\sqrt{2n}/\delta) ) \Vert \bm{\Gamma} \Vert^2}{(2n-1)}}.
\end{equation}
Now we turn to the analysis of the second term $\mathcal{E}_2(W,S_{[n]},(A_{i,j})_{n \times n})$. For any fixed realization $W=w$ and $(A_{i,j})_{n \times n} = (a_{i,j})_{n \times n}$, it can be decomposed it into the following terms
\begin{equation}
\begin{aligned}
    & \mathbb{E}_{S_{n+1}, A_{n+1,[n+1]}|S_{[n]},(A_{i,j})_{n \times n}=(a_{i,j})_{n \times n}} \big[ \ell(f_w(X_{[n+1]}, (\hat{a}_{i,j})_{n \times n}, A_{n+1,[n+1]}),Y_{n+1}) \big] \\
    & - \frac{1}{n} \sum_{\imath=1}^n \mathbb{E}_{S_{[\imath,n]}|S_{[\imath-1]}} \big[ \ell(f_{w,\imath}(X_{[n]}, (\hat{a}_{i,j})_{n \times n}), Y_{\imath}) \big] \\
    = & \mathbb{E}_{P_{A_{n+1,[n+1]}}|(A_{i,j})_{n \times n} = (a_{i,j})_{n \times n}} \mathbb{E}_{S_{n+1}|S_{[n]}} \big[ \ell(f_w(X_{[n+1]}, (\hat{a}_{i,j})_{n \times n}, A_{n+1,[n+1]}),Y_{n+1}) \big] \\
    & - \frac{1}{n} \sum_{\imath=1}^n \mathbb{E}_{S_{[\imath,n]}|S_{[\imath-1]}} \big[ \ell(f_{w,\imath}(X_{[n]}, (\hat{a}_{i,j})_{n \times n}), Y_{\imath}) \big] \\
    = & \mathbb{E}_{P_{A_{n+1,[n+1]}}|(A_{i,j})_{n \times n} = (a_{i,j})_{n \times n}} \mathbb{E}_{S_{n+1}|S_{[n]}} \big[ \ell(f_w(X_{[n+1]}, (\hat{a}_{i,j})_{n \times n}, A_{n+1,[n+1]}),Y_{n+1}) \big] \\
    & - \mathbb{E}_{Q_{A_{n+1,[n+1]}}|(A_{i,j})_{n \times n} = (a_{i,j})_{n \times n}} \mathbb{E}_{S_{n+1}|S_{[n]}} \big[ \ell(f_w(X_{[n+1]}, (\hat{a}_{i,j})_{n \times n}, A_{n+1,[n+1]}),Y_{n+1}) \big] \\
    & + \frac{1}{n} \sum_{\imath=1}^n \mathbb{E}_{S_{n+1}|S_{[n]}} \big[ \ell(f_w(X_{[n+1]}, (\hat{a}_{i,j})_{n \times n}, (\hat{a}_{\imath,[\imath-1]},0,\hat{a}_{\imath,(\imath,n]},\hat{a}_{\imath,\imath})),Y_{n+1}) \big] \\
    & - \frac{1}{n} \sum_{\imath=1}^n \mathbb{E}_{S_{[\imath,n]}|S_{[\imath-1]}} \big[ \ell(f_{w,\imath}(X_{[n]}, (\hat{a}_{i,j})_{n \times n}), Y_{\imath}) \big].
\end{aligned}
\end{equation}
By Assumption~\ref{assump1}, the first term can be bounded by
\begin{equation}\label{b1}
\begin{aligned}
    & \mathbb{E}_{P_{A_{n+1,[n+1]}}|(A_{i,j})_{n \times n} = (a_{i,j})_{n \times n}} \mathbb{E}_{S_{n+1}|S_{[n]}} \big[ \ell(f_w(X_{[n+1]}, (a_{i,j})_{n \times n}, A_{n+1,[n+1]}),Y_{n+1}) \big] \\
    & - \mathbb{E}_{Q_{A_{n+1,[n+1]}}|(A_{i,j})_{n \times n} = (a_{i,j})_{n \times n}} \mathbb{E}_{S_{n+1}|S_{[n]}} \big[ \ell(f_w(X_{[n+1]}, (a_{i,j})_{n \times n}, A_{n+1,[n+1]}),Y_{n+1}) \big] \\
    \leq & M {\rm D}_{\rm TV} \left( P_{A_{n+1,[n+1]} | (A_{i,j})_{n \times n} = (a_{i,j})_{n \times n}}, Q_{A_{n+1,[n+1]} | (A_{i,j})_{n \times n} = (a_{i,j})_{n \times n}} \right).
\end{aligned}
\end{equation}
For the second term, notice that
\begin{equation}
\begin{aligned}
    & \frac{1}{n} \sum_{\imath=1}^n \Big( \mathbb{E}_{ S_{n+1} | S_{[n]}} \left[ \ell(f_{w}(X_{[n+1]}, (\hat{a}_{i,j})_{n \times n}, (\hat{a}_{\imath,[\imath-1]},0,\hat{a}_{\imath,[\imath+1,n]},\hat{a}_{\imath,\imath})), Y_{n+1}) \right] \\
    & - \mathbb{E}_{ S_{[\imath,n]} | S_{[\imath-1]}} \left[ \ell(f_{w,\imath}(X_{[n]}, (\hat{a}_{i,j})_{n \times n}), Y_{\imath}) \right] \Big) \\
    = & \frac{1}{n} \sum_{\imath=1}^n \Big( \mathbb{E}_{ S_{n+1} | S_{[n]}} \left[ \ell(f_{w}(X_{[n+1]}, (\hat{a}_{i,j})_{n \times n}, (\hat{a}_{\imath,[\imath-1]},0,\hat{a}_{\imath,[\imath+1,n]},\hat{a}_{\imath,\imath})), Y_{n+1}) \right] \\
    & - \mathbb{E}_{ S_{\imath} | S_{[\imath-1]}} \left[ \ell(f_{w,\imath}(X_{[n]}, (\hat{a}_{i,j})_{n \times n}), Y_{\imath}) \right] \Big) \\ 
    & + \frac{1}{n} \sum_{\imath=1}^n \Big( \mathbb{E}_{ S_{\imath} | S_{[\imath-1]}} \left[ \ell(f_{w,\imath}(X_{[n]}, (\hat{a}_{i,j})_{n \times n}), Y_{\imath}) \right] - \mathbb{E}_{ S_{[\imath,n]} | S_{[\imath-1]}} \left[ \ell(f_{w,\imath}(X_{[n]}, (\hat{a}_{i,j})_{n \times n}), Y_{\imath}) \right] \Big) \\
    = & \frac{1}{n} \sum_{\imath=1}^n \Big( \mathbb{E}_{ S_{n+1} | S_{[n]}} \big[ \ell(f_{w}(X_{[n+1]}, (\hat{a}_{i,j})_{n \times n}, (\hat{a}_{\imath,[\imath-1]},0,\hat{a}_{\imath,[\imath+1,n]},\hat{a}_{\imath,\imath})), Y_{n+1}) \big] \\
    & - \mathbb{E}_{S_{n+1} | S_{[n]}} \big[ \ell(\tilde{f}_{w,\imath}(X_{[n+1]}, (\hat{a}_{i,j})_{n \times n}), Y_{n+1}) \big] \Big) \\ 
    & + \frac{1}{n} \sum_{\imath=1}^n \Big( \mathbb{E}_{ S_{n+1} | S_{[n]}} \big[ \ell(\tilde{f}_{w,\imath}(X_{[n+1]}, (\hat{a}_{i,j})_{n \times n}), Y_{n+1}) \big] - \mathbb{E}_{ S_{\imath} | S_{[\imath-1]}} \big[ \ell(f_{w,\imath}(X_{[n]}, (\hat{a}_{i,j})_{n \times n}), Y_{\imath}) \big] \Big) \\
    & + \frac{1}{n} \sum_{\imath=1}^n \Big( \mathbb{E}_{ S_{\imath} | S_{[\imath-1]}} \left[ \ell(f_{w,\imath}(X_{[n]}, (\hat{a}_{i,j})_{n \times n}), Y_{\imath}) \right] - \mathbb{E}_{ S_{[\imath,n]} | S_{[\imath-1]}} \left[ \ell(f_{w,\imath}(X_{[n]}, (\hat{a}_{i,j})_{n \times n}), Y_{\imath}) \right] \Big),
\end{aligned}
\end{equation}
where we define
\begin{equation}
    \tilde{f}_{w,\imath}(X_{[n+1]},(\hat{a}_{i,j})_{n \times n}) = \phi \left( \sum_{j \ne \imath} \hat{a}_{\imath, j} \phi \left( \sum_{k \ne \imath} \hat{a}_{j,k} X_k w_1 + \hat{a}_{j,\imath} X_{n+1} w_1 \right) w_2 \right).
\end{equation}
Recall that for any fixed $\imath \in [n]$, we have
\begin{equation}
    f_{w,\imath}(X_{[n]}, (\hat{a}_{i,j})_{n \times n}) = \phi \left( \sum_{j \ne \imath} \hat{a}_{\imath, j} \phi \left( \sum_{k \ne \imath} \hat{a}_{j,k} X_k w_1 + \hat{a}_{j,\imath} X_{\imath} w_1 \right) w_2 \right).
\end{equation}
Since $A_{j,n+1} = A_{n+1,j}$, for any fixed $\imath \in [n]$, we have
\begin{equation}
\begin{aligned}
    f_{w}(X_{[n+1]}, (\hat{a}_{i,j})_{n \times n}, (\hat{a}_{\imath,[\imath-1]},0,\hat{a}_{\imath,[\imath+1,n]},\hat{a}_{\imath,\imath})) = \phi \left( \sum_{j \ne \imath} \hat{a}_{\imath, j} \phi \left( \sum_{k=1}^n \hat{a}_{j,k} X_k w_1 + \hat{a}_{j,\imath} X_{n+1} w_1 \right) w_2 \right).
\end{aligned}
\end{equation}
For any fixed realization $S = s = ((x_1, y_1), \ldots, (x_n,y_n))$ and any unknown data point $S = (X,Y)$, we define
\begin{equation}
\begin{aligned}
    \hat{y}(X) & = \phi \left( \sum_{j \ne \imath} \hat{a}_{\imath, j} \phi \left( \sum_{k \ne \imath} \hat{a}_{j,k} x_k w_1 + \hat{a}_{j,\imath} X w_1 \right) w_2 \right), \ h(S) = \ell(\hat{y}(X),Y).
\end{aligned}
\end{equation}
Then we have
\begin{equation}
\begin{aligned}
    & h(S_{n+1}) = \ell(\tilde{f}_{w,\imath}((x_{[n]},X_{n+1}),(\hat{a}_{i,j})_{n \times n}), Y_{n+1}), \\
    & h(S_{\imath}) = \ell(f_{w,\imath}((x_{[\imath-1]},X_{\imath},x_{[\imath+1,n]}),(\hat{a}_{i,j})_{n \times n}), Y_{\imath}).
\end{aligned}
\end{equation}
For any fixed realization $S = s = ((x_1, y_1), \ldots, (x_n,y_n))$ and any $\imath \in [n]$, we have
\begin{equation}
\begin{aligned}
    & \mathbb{E}_{ S_{n+1} | S_{[n]}} \big[ \ell(\tilde{f}_{w,\imath}(X_{[n+1]}, (\hat{a}_{i,j})_{n \times n}), Y_{n+1}) \big] - \mathbb{E}_{ S_{\imath} | S_{[\imath-1]}} \big[ \ell(f_{w,\imath}(X_{[n]}, (\hat{a}_{i,j})_{n \times n}), Y_{\imath}) \big] \\
    = & \mathbb{E}_{S_{n+1}|S_{[n]}=s_{[n]}} \left[ h(S_{n+1}) \right] - \mathbb{E}_{S_{\imath} |S_{[\imath-1]} = s_{[\imath-1]}} \left[ h(S_{\imath}) \right] \\
    = & \int_{s_{n+1}} h(s_{n+1}) \dif P_{S_{n+1}|S=s}(s_{n+1}) - \int_{s_{\imath}} h(s_{\imath}) \dif P_{S_{\imath} | S_{[\imath-1]} = s_{[\imath-1]}}(s_{\imath}) \\
    \leq & \left\vert \int_{s_{n+1}} h(s_{n+1}) \dif P_{S_{n+1}|S=s}(s_{n+1}) - \int_{s_{\imath}} h(s_{\imath}) \dif P_{S_{\imath} | S_{[\imath-1]} = s_{[\imath-1]}}(s_{\imath}) \right\vert \\
    \leq & M {\rm D}_{\rm TV} \left( P_{S_{n+1}|S_{[n]}=s_{[n]}}, P_{S_{\imath} |S_{[\imath-1]} = s_{[\imath-1]}} \right),
\end{aligned}
\end{equation}
which implies that
\begin{equation}
\begin{aligned}
    & \mathbb{E}_{ S_{n+1} | S} \big[ \ell(\tilde{f}_{w,\imath}(X_{[n+1]}), Y_{n+1}) \big] - \mathbb{E}_{ S_{\imath} | S_{[\imath-1]}} \big[ \ell(f_{w,\imath}(X_{[n]}), Y_{\imath}) \big] \leq M {\rm D}_{\rm TV} \left( P_{S_{n+1}|S_{[n]}}, P_{S_{\imath} |S_{[\imath-1]}} \right).
\end{aligned}
\end{equation}
Therefore, we have
\begin{equation}\label{b2}
\begin{aligned}
    & \frac{1}{n} \sum_{\imath=1}^n \Big( \mathbb{E}_{ S_{n+1} | S} \big[ \ell(\tilde{f}_{w,\imath}(X_{[n+1]}), Y_{n+1}) \big] - \mathbb{E}_{ S_{\imath} | S_{[\imath-1]}} \big[ \ell(f_{w,\imath}(X_{[n]}), Y_{\imath}) \big] \Big) \\
    \leq & \frac{1}{n} \sum_{i=1}^n M {\rm D}_{\rm TV} \left( P_{S_{n+1}|S_{[n]}}, P_{S_{i} |S_{[i-1]}} \right).
\end{aligned}
\end{equation}
Besides, notice that
\begin{equation}
\begin{aligned}
    & \left\vert \ell(f_{w}(X_{[n+1]}, (\hat{a}_{i,j})_{n \times n}, (\hat{a}_{\imath,[\imath-1]},0,\hat{a}_{\imath,[\imath+1,n]},\hat{a}_{\imath,\imath})), Y_{n+1}) - \ell(\tilde{f}_{w,\imath}(X_{[n+1]}), Y_{n+1}) \right\vert \\
    \leq & c_w L L_{\phi} \Bigg\Vert \sum_{j \ne \imath} \hat{a}_{\imath, j} \phi \Bigg( \sum_{k=1}^n \hat{a}_{j,k} X_k w_1 + \hat{a}_{j,\imath} X_{n+1} w_1 \Bigg) - \sum_{j \ne \imath} \hat{a}_{\imath, j} \phi \Bigg( \sum_{k \ne \imath} \hat{a}_{j,k} X_k w_1 + \hat{a}_{j,\imath} X_{n+1} w_1 \Bigg) \Bigg\Vert \\
    \leq & c^2_w L L^2_{\phi} \sum_{j \ne \imath} \hat{a}_{\imath,j} \Vert \hat{a}_{j,\imath} X_{\imath} \Vert \leq c_x c^2_w L L^2_{\phi} \sum_{j \ne \imath} \hat{a}^2_{\imath,j}.
\end{aligned}
\end{equation}
Thus,
\begin{equation}\label{b3}
\begin{aligned}
    & \frac{1}{n} \sum_{\imath=1}^n \Big( \mathbb{E}_{ S_{n+1} | S_{[n]}} \big[ \ell(f_{w}(X_{[n+1]}, (\hat{a}_{i,j})_{n \times n}, (\hat{a}_{\imath,[\imath-1]},0,\hat{a}_{\imath,[\imath+1,n]},\hat{a}_{\imath,\imath})), Y_{n+1}) \big] \\
    & - \mathbb{E}_{S_{n+1} | S_{[n]}} \big[ \ell(\tilde{f}_{w,\imath}(X_{[n+1]}, (\hat{a}_{i,j})_{n \times n}), Y_{n+1}) \big] \Big) \\
    \leq & \frac{1}{n} \sum_{\imath=1}^n \mathbb{E}_{ S_{n+1} | S_{[n]}} \big[ \big\vert \ell(f_{w}(X_{[n+1]}, (\hat{a}_{i,j})_{n \times n}, (\hat{a}_{\imath,[\imath-1]},0,\hat{a}_{\imath,[\imath+1,n]},\hat{a}_{\imath,\imath})), Y_{n+1}) \big] \\
    & - \ell(\tilde{f}_{w,\imath}(X_{[n+1]}, (\hat{a}_{i,j})_{n \times n}), Y_{n+1}) \big\vert \big] \\
    \leq & \frac{c_x c^2_w L L^2_{\phi}}{n} \sum_{\imath=1}^n \sum_{j \ne \imath} \hat{a}^2_{\imath,j} = \frac{c_x c^2_w L L^2_{\phi} \Vert (\hat{a}_{ij})_{n \times n} \Vert^2_F}{n}.
\end{aligned}
\end{equation}
where we have used the fact that $\hat{a}_{i,i} = 0$ for any $i \in [n]$. 
Define
\begin{equation}
\begin{aligned}
    \widetilde{\mathcal{E}}_2(W,S_{[n]},(A_{i,j})_{n \times n}) = & \frac{1}{n} \sum_{\imath=1}^n \mathbb{E}_{S_{\imath} |S_{[\imath-1]}} \left[ \ell(f_{W,\imath}((X_{[\imath]},X_{(\imath,n]}), (\tilde{A}_{i,j})_{n \times n}), Y_{\imath}) \right] \\
    & - \frac{1}{n} \sum_{\imath=1}^n \mathbb{E}_{ S_{[\imath,n]} | S_{[\imath-1]}} \left[ \ell(f_{W,\imath}((X_{[\imath-1]}, X_{[\imath,n]}), (\tilde{A}_{i,j})_{n \times n}), Y_{\imath}) \right],
\end{aligned}
\end{equation}
our next step is applying the result in Theorem~\ref{main_pro2} to show that 
\begin{equation}
    \mathbb{E}_{W \sim Q} \left[ \widetilde{\mathcal{E}}_2(W,S_{[n]},(A_{i,j})_{n \times n}) \right]
\end{equation}
can be bounded by its expectation in high probability. To this end, we need to provide an upper bound for the term $\vert \widetilde{\mathcal{E}}_2(w, s_{[n]}, (a_{i,j})_{n \times n}) - \widetilde{\Psi}_2(w, s^{(i)}_{[n]}, (a_{i,j})_{n \times n}) \vert$ for any fixed realization $(A_{i,j})_{n \times n} = (a_{i,j})_{n \times n}$, where $s^{(i)}_{[n]}=(s_1,\ldots,s_{i-1},s'_i,s_{i+1},\ldots,s_n)$ is the sequence of data points that differs from $s_{[n]}=(s_1,\ldots,s_n)$ only in the $i$-th entry. If $i = n$, then we have $\widetilde{s}_j = s_j$ for $j \in [n-1]$, which implies that $P_{S_{[j,n]} | S_{[j-1]}=s_{[j-1]}} = P_{S_{[j,n]} | S_{[j-1]}=\widetilde{s}_{[j-1]}}$. One can verify that $\vert \widetilde{\mathcal{E}}_2(w, s_{[n]}, (a_{i,j})_{n \times n}) - \widetilde{\mathcal{E}}_2(w, s^{(i)}_{[n]}, (a_{i,j})_{n \times n}) \vert = 0$ holds. If $i < n$, for any $\imath \in [i+1, n]$:
\begin{equation}\label{b4}
\begin{aligned}
    & \Big\vert \mathbb{E}_{ S_{\imath} | S_{[\imath-1]}=s_{[\imath-1]}} \left[ \ell(f_{w,\imath}((x_{[\imath-1]},X_\imath,x_{(\imath,n]}), (\hat{a}_{i,j})_{n \times n}), Y_\imath) \right] \\
    & - \mathbb{E}_{ S_{\imath} | S_{[\imath-1]}=\widetilde{s}_{[\imath-1]}} \left[ \ell(f_{w,\imath}((\widetilde{s}_{[\imath-1]},X_\imath,x_{(\imath,n]}), (\hat{a}_{i,j})_{n \times n}), Y_\imath) \right] \Big\vert \\
    = & \Big\vert \mathbb{E}_{ S_{\imath} | S_{[i-1]}=s_{[i-1]}, S_{i}=s_i, S_{(i,\imath-1]}=s_{(i,\imath-1]}} \left[ \ell(f_{w,\imath}((x_{[i-1]},x_i,x_{(i,\imath-1]},X_\imath,x_{(\imath,n]}), (\hat{a}_{i,j})_{n \times n}), Y_\imath) \right] \\
    & - \mathbb{E}_{ S_{\imath} | S_{[i-1]}=s_{[i-1]}, S_{i}=s'_i, S_{(i,\imath-1]}=s_{(i,\imath-1]}} \left[ \ell(f_{w,\imath}((x_{[i-1]},x'_i,x_{(i,\imath-1]},X_\imath,x_{(\imath,n]}), (\hat{a}_{i,j})_{n \times n}), Y_\imath) \right] \Big\vert \\
    \leq & \Big\vert \mathbb{E}_{ S_{\imath} | S_{[i-1]}=s_{[i-1]}, S_{i}=s_i, S_{(i,\imath-1]}=s_{(i,\imath-1]}} \left[ \ell(f_{w,\imath}((x_{[i-1]},x_i,x_{(i,\imath-1]},X_\imath,x_{(\imath,n]}), (\hat{a}_{i,j})_{n \times n}), Y_\imath) \right] \\
    & - \mathbb{E}_{ S_{\imath} | S_{[i-1]}=s_{[i-1]}, S_{i}=s'_i, S_{(i,\imath-1]}=s_{(i,\imath-1]}} \left[ \ell(f_{w,\imath}((x_{[i-1]},x_i,x_{(i,\imath-1]},X_\imath,x_{(\imath,n]}), (\hat{a}_{i,j})_{n \times n}), Y_\imath) \right] \Big\vert \\
    & + \Big\vert \mathbb{E}_{ S_{\imath} | S_{[i-1]}=s_{[i-1]}, S_{i}=s'_i, S_{(i,\imath-1]}=s_{(i,\imath-1]}} \left[ \ell(f_{w,\imath}((x_{[i-1]},x_i,x_{(i,\imath-1]},X_\imath,x_{(\imath,n]}), (\hat{a}_{i,j})_{n \times n}), Y_\imath) \right] \\
    & - \mathbb{E}_{ S_{\imath} | S_{[i-1]}=s_{[i-1]}, S_{i}=s'_i, S_{(i,\imath-1]}=s_{(i,\imath-1]}} \left[ \ell(f_{w,\imath}((x_{[i-1]},x'_i,x_{(i,\imath-1]},X_\imath,x_{(\imath,n]}), (\hat{a}_{i,j})_{n \times n}), Y_\imath) \right] \Big\vert,
\end{aligned}
\end{equation}
where we have stipulated that $(i,\imath-1] \triangleq \varnothing$ if $\imath=i+1$ and $[\imath+1,\imath] = \varnothing$ if $\imath=n$. Following the procedure of deriving Eqs.~(\ref{pre1}, \ref{pre2}), we have
\begin{equation}
\begin{aligned}
    & \Big\vert \mathbb{E}_{ S_{\imath} | S_{[i-1]}=s_{[i-1]}, S_{i}=s_i, S_{(i,\imath-1]}=s_{(i,\imath-1]}} \left[ \ell(f_{w,\imath}((x_{[i-1]},x_i,x_{(i,\imath-1]},X_\imath,x_{(\imath,n]}), (\hat{a}_{i,j})_{n \times n}), Y_\imath) \right] \\
    & - \mathbb{E}_{ S_{\imath} | S_{[i-1]}=s_{[i-1]}, S_{i}=s'_i, S_{(i,\imath-1]}=s_{(i,\imath-1]}} \left[ \ell(f_{w,\imath}((x_{[i-1]},x_i,x_{(i,\imath-1]},X_\imath,x_{(\imath,n]}), (\hat{a}_{i,j})_{n \times n}), Y_\imath) \right] \Big\vert \\
    \leq & M \sup_{s_i, s'_i} {\rm D}_{\rm TV} \left( P_{S_{\imath} | S_{[i-1]}=s_{[i-1]}, S_{i}=s_i, S_{(i,\imath-1]}=s_{(i,\imath-1]}}, P_{S_{\imath} | S_{[i-1]}=s_{[i-1]}, S_{i}=s'_i, S_{(i,\imath-1]}=s_{(i,\imath-1]}} \right) \\
    \leq & M \sup_{s_i, s'_i} {\rm D}_{\rm TV} \left( P_{S_{[\imath,n]} | S_{[i-1]}=s_{[i-1]}, S_{i}=s_i, S_{(i,\imath-1]}=s_{(i,\imath-1]}}, P_{S_{[\imath,n]} | S_{[i-1]}=s_{[i-1]}, S_{i}=s'_i, S_{(i,\imath-1]}=s_{(i,\imath-1]}} \right),
\end{aligned}
\end{equation}
and
\begin{equation}
\begin{aligned}
    & \Big\vert \mathbb{E}_{ S_{\imath} | S_{[i-1]}=s_{[i-1]}, S_{i}=s'_i, S_{(i,\imath-1]}=s_{(i,\imath-1]}} \left[ \ell(f_{w,\imath}((x_{[i-1]},x_i,x_{(i,\imath-1]},X_\imath,x_{(\imath,n]}), (\hat{a}_{i,j})_{n \times n}), Y_\imath) \right] \\
    & - \mathbb{E}_{ S_{\imath} | S_{[i-1]}=s_{[i-1]}, S_{i}=s'_i, S_{(i,\imath-1]}=s_{(i,\imath-1]}} \left[ \ell(f_{w,\imath}((x_{[i-1]},x'_i,x_{(i,\imath-1]},X_\imath,x_{(\imath,n]}), (\hat{a}_{i,j})_{n \times n}), Y_\imath) \right] \Big\vert \\
    \leq & 2 c_x c^2_w L L^2_{\phi} \sum_{j=1}^n \hat{a}_{\imath,j} \hat{a}_{j,i}.
\end{aligned}
\end{equation}
Combining the above ingredients, for any $\imath \in [i+1, n]$:
\begin{equation}\label{pre3}
\begin{aligned}
    & \Big\vert \mathbb{E}_{ S_{\imath} | S_{[\imath-1]}=s_{[\imath-1]}} \left[ \ell(f_{w,\imath}((x_{[\imath-1]},X_\imath,x_{(\imath,n]}), (\hat{a}_{i,j})_{n \times n}), Y_\imath) \right] \\
    & - \mathbb{E}_{ S_{\imath} | S_{[\imath-1]}=\widetilde{s}_{[\imath-1]}} \left[ \ell(f_{w,\imath}((\widetilde{s}_{[\imath-1]},X_\imath,x_{(\imath,n]}), (\hat{a}_{i,j})_{n \times n}), Y_\imath) \right] \Big\vert \\
    \leq & 2 c_x c^2_w L L^2_{\phi} \sum_{j=1}^n \hat{a}_{\imath,j} \hat{a}_{j,i} + M \widetilde{\Gamma}_{i,\imath}.
\end{aligned}
\end{equation}
Combining Eq.~(\ref{b4}) and Eq.~(\ref{pre3}), for any fixed realization $W=w$, $(A_{i,j})_{n \times n}=(a_{i,j})_{n \times n}$ and any $i \in [n]$, we have
\begin{equation}\label{exp2}
\begin{aligned}
    \left\vert \widetilde{\mathcal{E}}_2(w, s_{[n]}, (a_{i,j})_{n \times n}) - \widetilde{\mathcal{E}}_2(w, s^{(i)}_{[n]}, (a_{i,j})_{n \times n}) \right\vert \leq \frac{\mathds{1}\{i \neq n\}(2 c_x c^2_w c^2_a L L^2_{\phi} + M \Vert \widetilde{\bm{\Gamma}} \Vert_{\infty})}{n}.
\end{aligned}
\end{equation}
Now we are in a place to derive the final result by applying Eq.~(\ref{final_1}). For each $i \in [n]$, Let $c_i$ be the right hand side of Eq.~(\ref{exp2}), we have
\begin{equation}
\begin{aligned}
    \Vert \bm{c} \Vert^2_2 = \frac{1}{n^2} \sum_{i=1}^{n-1} \left( 2 c_x c^2_w c^2_a L L^2_{\phi} + M \Vert \widetilde{\bm{\Gamma}} \Vert_{\infty} \right)^2 \leq \frac{(2 c_x c^2_w c^2_a L L^2_{\phi} + M \Vert \widetilde{\bm{\Gamma}} \Vert_{\infty})^2}{n}
\end{aligned}
\end{equation}
Now using the inequality that $\Vert \bm{\Gamma} \bm{c} \Vert_2 \leq \Vert \bm{\Gamma} \Vert \Vert \bm{c} \Vert_2$ and applying Eq.~(\ref{final_1}), for any fixed realization $W=w$ and $(A_{i,j})_{n \times n} = (a_{i,j})_{n \times n}$:
\begin{equation}
    \mathbb{E}_{S_{[n]}|(A_{i,j})_{n \times n} = (a_{i,j})_{n \times n}} \left[ e^{\lambda \alpha \big(\widetilde{\mathcal{E}}_2(w, S_{[n]}, (a_{i,j})_{n \times n}) - \mathbb{E}_{S'_{[n]}} [ \widetilde{\mathcal{E}}_2(w, s_{[n]}, (a_{i,j})_{n \times n}) ] \big)} \right] \leq \exp \left\{ \frac{\lambda^2 \alpha^2 \Vert \bm{\Gamma} \Vert^2 \Vert \bm{c} \Vert^2}{8} \right\}.
\end{equation}
Taking expectation over $(A_{i,j})_{n \times n}$ on both side yields
\begin{equation}
    \mathbb{E}_{S_{[n]}, (A_{i,j})_{n \times n}} \left[ e^{\lambda \alpha \big(\widetilde{\mathcal{E}}_2(w, S_{[n]}, (a_{i,j})_{n \times n}) - \mathbb{E}_{S'_{[n]}} [ \widetilde{\mathcal{E}}_2(w, s_{[n]}, (a_{i,j})_{n \times n}) ] \big)} \right] \leq \exp \left\{ \frac{\lambda^2 \alpha^2 \Vert \bm{\Gamma} \Vert^2 \Vert \bm{c} \Vert^2}{8} \right\}.
\end{equation}
Following the procedure of deriving Eq.~(\ref{main_pro_bound1}), for any $\delta \in (0,1)$, with probability at least $1-\delta/2$ over the randomness of $S_{[n]}$ and $(A_{i,j})_{n \times n}$:
\begin{equation}\label{b5}
\begin{aligned}
    & \mathbb{E}_{W \sim Q} \left[ \widetilde{\mathcal{E}}_2(W, S_{[n]}, (A_{i,j})_{n \times n}) \right] \\
    \leq & \mathbb{E}_{W \sim Q} \mathbb{E}_{S'_{[n]}} \left[ \widetilde{\mathcal{E}}_2(W, S'_{[n]}, (A_{i,j})_{n \times n}) \right] + \sqrt{\frac{(2 c_x c^2_w c^2_a L L^2_{\phi} + M \Vert \widetilde{\bm{\Gamma}} \Vert_{\infty})^2({\rm D}_{\alpha}(Q||P)+\log(2\sqrt{2n}/\delta) ) \Vert \bm{\Gamma} \Vert^2}{(2n-1)}} .
\end{aligned}
\end{equation}
Following the procedure of deriving Eq.~(\ref{exp_phi9}), we have
\begin{equation}\label{b6}
\begin{aligned}
    \mathbb{E}_{W \sim Q} \mathbb{E}_{S'_{[n]}} \left[ \widetilde{\mathcal{E}}_2(W, S'_{[n]}, (A_{i,j})_{n \times n}) \right] \leq \frac{M}{n} \sum_{i=1}^n \mathbb{E}_S \left[ {\rm D}_{\rm TV} \Big( P_{S_{[i+1,n]}|S_{[i-1]}} \otimes P_{S_i |S_{[i-1]}}, P_{S_{[i,n]} | S_{[i-1]}} \Big) \right].
\end{aligned}
\end{equation}
Combining Eq.~(\ref{b1}), (\ref{b2}), (\ref{b3}), (\ref{b5}), and (\ref{b6}), with probability at least $1-\delta/2$ over the randomness of $S_{[n]}$ and $(A_{i,j})_{n \times n}$:
\begin{equation}\label{b7}
\begin{aligned}
    & \mathbb{E}_{W \sim Q} \left[ \mathcal{E}_2(W, S_{[n]}, (A_{i,j})_{n \times n}) \right] \\
    \leq & \sqrt{\frac{(2 c_x c^2_w c^2_a L L^2_{\phi} + M \Vert \widetilde{\bm{\Gamma}} \Vert_{\infty})^2({\rm D}_{\alpha}(Q||P)+\log(2\sqrt{2n}/\delta) ) \Vert \bm{\Gamma} \Vert^2}{(2n-1)}}\\
    & + M {\rm D}_{\rm TV} \left( P_{A_{n+1,[n+1]} | (A_{i,j})_{n \times n}}, Q_{A_{n+1,[n+1]} | (A_{i,j})_{n \times n}} \right) + \frac{M}{n} \sum_{i=1}^n {\rm D}_{\rm TV} \left( P_{S_{n+1}|S_{[n]}}, P_{S_i |S_{[i-1]}} \right) \\
    & + \frac{M}{n} \sum_{i=1}^n \mathbb{E}_S \left[ {\rm D}_{\rm TV} \Big( P_{S_{[i+1,n]}|S_{[i-1]}} \otimes P_{S_i |S_{[i-1]}}, P_{S_{[i,n]} | S_{[i-1]}} \Big) \right] + \frac{c_x c^2_w L L^2_{\phi} \Vert (\hat{A}_{ij})_{n \times n} \Vert^2_F}{n}.
\end{aligned}
\end{equation}
The final result is now can be obtained by combining Eq.~(\ref{b0}) and Eq.~(\ref{b7}).

\bibliographystyle{plainnat}
\bibliography{main.bib}
\end{document}